\documentclass[runningheads]{llncs}

\usepackage[mobile]{eccv}

\usepackage{eccvabbrv}

\usepackage{graphicx}
\usepackage{booktabs}
\usepackage{wrapfig}

\usepackage[accsupp]{axessibility}  %

\usepackage[pagebackref,breaklinks,colorlinks]{hyperref}

\usepackage{orcidlink}
\usepackage{soul}
\usepackage{makecell}
\usepackage{colortbl}
\usepackage{xcolor}
\usepackage{multirow}

\usepackage[normalem]{ulem}  %
\usepackage{overpic}
\def\method{SWAG} %

\begin{document}

\title{SWAG: Splatting in the Wild images with Appearance-conditioned Gaussians} 

\titlerunning{SWAG}

\author{Hiba Dahmani \and
Moussab Bennehar  \and
Nathan Piasco \and 
Luis Rold$\Tilde{a}$o
and Dzmitry Tsishkou }

\authorrunning{H.~Dahmani et al.}

\institute{Noah’s Ark, Huawei Paris Research Center, France}

\maketitle

\begin{figure}[htb]

\centering
\includegraphics[width=.95\linewidth]{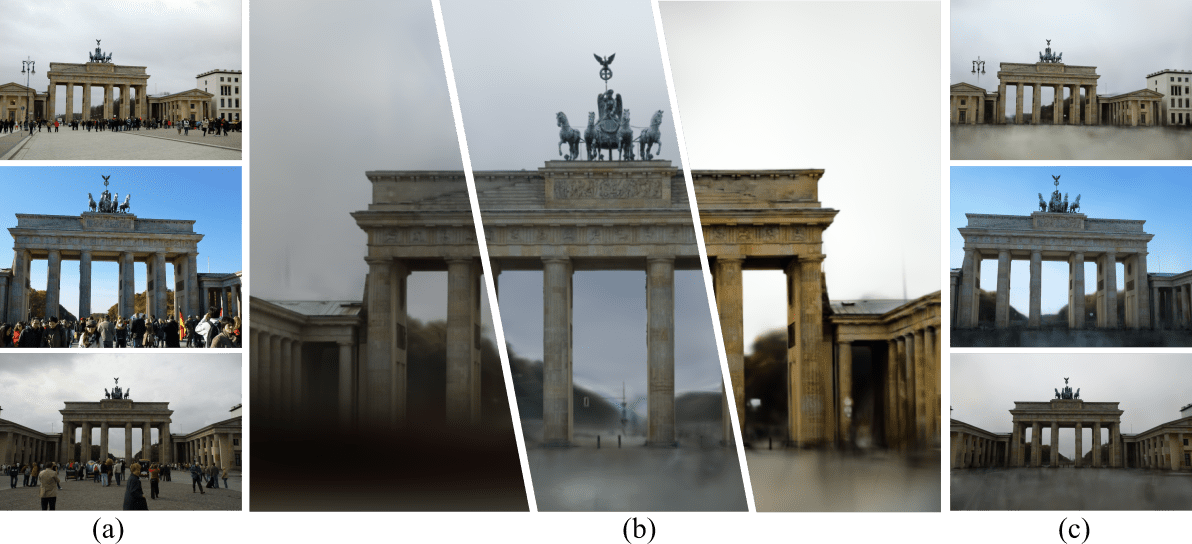}
  
  \caption{Given \textit{in-the-wild} captures (a), our model enables transient objects removal (b) and scene reconstruction with variable appearances (c).}
  \label{fig:teaser}
  \end{figure}

\begin{abstract}

Implicit neural representation methods have shown impressive advancements in learning 3D scenes from unstructured \textit{\textit{in-the-wild}} photo collections but are still limited by the large computational cost of volumetric rendering. More recently, 3D Gaussian Splatting emerged as a much faster alternative with superior rendering quality and training efficiency, especially for small-scale and object-centric scenarios. Nevertheless, this technique suffers from poor performance on unstructured \textit{\textit{in-the-wild}} data. To tackle this, we extend over 3D Gaussian Splatting to handle unstructured image collections. We achieve this by modeling appearance to seize photometric variations in the rendered images. Additionally, we introduce a new mechanism to train transient Gaussians to handle the presence of scene occluders in an unsupervised manner. Experiments on diverse photo collection scenes and multi-pass acquisition of outdoor landmarks show the effectiveness of our method over prior works achieving state-of-the-art results with improved efficiency.

\end{abstract}

\section{Introduction}
Novel View Synthesis (NVS) and 3D scene reconstruction are long-standing challenging tasks in the realms of computer vision and computer graphics. Over the past few years, Neural radiance Fields (NeRF)~\cite{mildenhall2020nerf} have shown groundbreaking results in rendering photorealistic views from novel viewpoints using an implicit neural representation of the radiance field and density in a scene. Although NeRFs are very effective in static scenes, their performance significantly degrades in dynamic scenarios (i.e. containing moving transient objects) or in the presence of changing conditions such as weather, exposure, and lighting (i.e. \textit{in-the-wild} datasets). To tackle these challenges, NeRF-W~\cite{martinbrualla2020nerfw} proposed an extension to NeRF enabling the reconstruction of outdoor landmarks from \textit{in-the-wild} images. They achieve this by learning the changing appearance of the images through per-image embeddings. Furthermore, to handle the presence of occluders, the scenes are decomposed into ``static'' and ``transient'' elements being modeled by separate radiance fields. These modifications improved NeRF's performance when confronted with appearance changes and transient occluders. More recent works~\cite{chen2022hallucinated,yang2023crossray} improve transient object handling by leveraging 2D visibility maps. Nonetheless, these methods still suffer from the inherent cost of volumetric rendering. 

Recently, 3D Gaussian Splatting (3DGS)~\cite{kerbl3Dgaussians} attracted considerable interest thanks to its explicit representation resulting in faster training and rendering by leveraging GPU-based rasterization. 3DGS achieves fast training and matches visual quality obtained by the current SOTA NeRF methods~\cite{DBLP:journals/corr/abs-2201-05989}. This work represents the scene as a large number of anisotropic 3D Gaussians with color features and opacities. The Gaussians' set is then used in a differentiable splatting process to project the 3D Gaussians into a 2D image plane and blend them to get the rendered image. Similar to the original NeRF~\cite{mildenhall2020nerf}, 3DGS suffers from the same limitations regarding varying lighting conditions and the presence of occluders on some of the training images.

In this work, we present SWAG, the first \textit{in-the-wild} extension for 3DGS. To achieve this, we propose to expand the capabilities of 3DGS and improve its robustness in these scenarios. %
To this aim, we capture the appearance of each image using a learned embedding space that modulates the color of the Gaussians with a multilayer perceptron (MLP). Second, we learn an image-dependent opacity variation to each Gaussian that enables a better handling of transient objects and a finer scene reconstruction as illustrated in~\Cref{fig:teaser}.

We conduct a variety of experiments on the Phototourism dataset~\cite{Jin_2020} and NeRF-OSR~\cite{rudnev2022nerfosr} benchmark and show that not only do we improve 3DGS performance in these scenarios, but we also achieve state-of-the-art rendering quality with significantly faster training and rendering speed compared to previous works~\cite{martinbrualla2020nerfw,kassab2024refinedfields,chen2022hallucinated,yang2023crossray}. To summarize, our main contributions are as follows:

\begin{itemize}
  \item we introduce a new method for reconstructing scenes from unconstrained photo collections with varying appearances,
  \item we design a specific training scheme that captures occluders and enables transient object removal from a trained scene,
  \item we show that our method achieves SoTA performance in NVS \textit{in-the-wild} scenarios,
  \item we demonstrate the ability of SWAG to generate new images with smooth visual transitions from learned appearances and to remove in an unsupervised manner transient objects from a captured scene.
\end{itemize}

\section{Related Work}

\subsection{Neural Rendering \textit{in-the-wild}}
The abundance of \textit{in-the-wild} unconstrained image data has encouraged adapting neural radiance fields to such cases. Early works like NeRF-W~\cite{martinbrualla2020nerfw} have proposed a disentanglement between static and transient occluders using two appearance and transient per-image embeddings in addition to two radiance fields for each component of the scene (i.e. static and transient). Ha-NeRF~\cite{chen2022hallucinated} on the other hand uses a 2D image-dependent visibility map to eliminate occluders instead of using a decoupled radiance field since transient phenomena are only observed in individual 2D images. This simplification reduces the blurry artifacts that NeRF-W~\cite{martinbrualla2020nerfw} suffered from in the attempt to reconstruct transient phenomena with a 3D transient field. Building upon previous methods, CR-NeRF~\cite{yang2023crossray} enhances their performance by exploiting the interaction information from multiple rays and fusing it into global information. Using a light-weight segmentation network, this method learns a visibility map without the supervision of ground truth segmentation masks to eliminate transient parts in 2D images. Another recent work, RefinedFields~\cite{kassab2024refinedfields}, leverages K-Planes and generative priors for \textit{in-the-wild} scenarios. The learning is done through the alternation of two stages: a scene fitting to optimize the K-Planes~\cite{fridovichkeil2023kplanes} representation and a scene enriching phase that finetunes a pre-trained generative prior and infers a new  K-Planes representation. As presented, Implicit-field representations have found diverse adaptations for \textit{in-the-wild} scenarios. However, their time-consuming training and inference pose a challenge to achieving real-time rendering. This constrains their application in practical scenarios where fast rendering speed is crucial, especially in diverse interactive 3D applications. %

\subsection{Point-Based Rendering}

Although NeRF demonstrated exceptional capabilities in generating photorealistic images, the demand for faster and more efficient rendering methods became increasingly evident. Point-based rendering has emerged as a prominent approach in computer graphics. It mainly revolves around the representation and rendering of the scene using discrete primitives. Starting from the most elementary form, Point Sample Rendering~\cite{article} authors proposed to sample a fixed-size formless set of points from a surface. Despite the developed strategies to adequately sample objects or surfaces, point sample rendering still suffers from holes and discontinuity. To overcome the issue of 3D points sparsity, the following works used different geometric primitives like discs~\cite{1500313} and surfels~\cite{10.1145/383259.383300} that have a larger extent compared to pixels.
Differentiable point-based rendering is the groundwork of these methods since it enables end-to-end training of 3D primitives from images. Pulsar~\cite{DBLP:journals/corr/abs-2004-07484} introduced an efficient sphere-based differentiable renderer where the scene is represented by a set of spheres with learned positions, feature vectors, opacities, and radii. These parameters are optimized to minimize the photometric reconstruction loss using an efficient differentiable rendering. The rasterization is done using either a convolutional U-Net or a per-pixel one-by-one convolutional network applied to a 2D screen space feature map. The feature map is a blending function that combines each sphere channel information using weighting based on position, radius, and opacity. Recently 3DGS~\cite{kerbl3Dgaussians} used 3D Gaussians as primitives and introduced a fully differentiable fast tile-based rasterizer for Gaussian splats allowing real-time rendering while outperforming state-of-the-art visual quality. Despite the demonstrated capability of 3DGS, there is still room for refining this method to enhance the quality of rendered images. One observed issue with 3DGS~\cite{kerbl3Dgaussians} is the appearance of strong artifacts when changing the sampling rate. Additionally, the rendering quality declines significantly in low resolutions or far away from camera positions due to the aliasing caused by the pixel size compared to the screen size and the Nyquist frequency.

\subsection{3DGS Rendering Improvement}

To solve the aforementioned rendering issue, a multi-scale adaptation of 3DGS~\cite{yan2023multiscale} has been introduced. This work explains the aliasing phenomena by the splitting of large amounts of Gaussians with a smaller extent than pixels in 3D regions featuring high-frequency details. To render at a certain scale, authors filter out too large or too small Gaussians. Additionally, small Gaussians are aggregated to form larger ones selected on bigger scales during training to prevent missing areas and low-frequency details. Motivated by Mip-NeRF's~\cite{barron2021mipnerf} rendering quality, Mip-Splatting~\cite{Yu2023MipSplatting} replaces the 2D dilation filter introduced in Surface splatting~\cite{10.1145/383259.383300} using 2D Mip filters that replicate box filters to solve the aliasing issue and uses 3D smoothing filters with a maximal sampling rate obtained from training images. One other issue with 3DGS is its performance deterioration in scenes that may present variation in appearance. VastGaussian~\cite{lin2024vastgaussian} introduced a decoupled appearance modeling using a pixel-wise appearance embedding and a CNN to predict a pixel-wise transformation for the rendered images. Hence these predicted transformations (multiplication, addition, and Gamma correction) work well for large scenes but they are insufficient to model more varying appearances that appear \textit{in-the-wild} scenarios. Our work aims to extend \textit{in-the-wild} 3DGS scene representation beyond closed-world setups. %

\section{Background}
The scene in 3DGS~\cite{kerbl3Dgaussians} is represented by a set of 3D anisotropic Gaussians. Each Gaussian is parameterized by its centroid $\mathbf{x} \in \mathbb{R}^3$, scale $\mathbf{S} \in \mathbb{R}^3$, rotation matrix  $\mathbf{R} \in {\mathbb{R}^{3 \times 3}}$, opacity $\mathbf{\alpha} \in \mathbb{R}$ and color $\mathbf{c} \in \mathbb{R}^3$ encoded in spherical harmonic ($\mathbf{SH}$) coefficients. The 3D covariance matrix 
$\mathbf{\Sigma}$ of the 3D Gaussian is obtained using its rotation matrix $\mathbf{R}$ and scale $\mathbf{S}$:
\begin{equation}
  \mathbf{\Sigma} = \mathbf{R}\mathbf{S}\mathbf{S}^T\mathbf{R}^T.
  \label{eq:covariance}
\end{equation} 
The 3D Gaussians are defined in world space following:
\begin{equation}
  \mathbf{G}(\mathbf{y}) = e^{-\frac{1}{2}(\mathbf{y-x})^T\mathbf{\Sigma}^{-1}(\mathbf{y-x})}.
  \label{eq:worldspaceG}
\end{equation} 
Given $\mathbf{J}$, the Jacobian of the affine projective transformation and $\mathbf{W}$, a viewing transformation, the covariance matrix $\mathbf{\Sigma'}$ in camera coordinates is as follows:
\begin{equation}
  \mathbf{\Sigma'}= \mathbf{J}\mathbf{W}\mathbf{\Sigma} \mathbf{W}^T\mathbf{J}^T.
  \label{eq:cameraCov}
\end{equation}
These 3D Gaussians are projected to 2D splats and blended during a fast differentiable $\alpha$-blending process to get 2D rendered images. Each pixel $\mathbf{y}$ color value $\mathbf{C}$ is calculated using N-ordered 2D splats using the formula:

\begin{equation}
 \mathbf{C}=\sum_{i \in \mathbf{N}} \mathbf{\alpha}_i' \mathbf{c}_i \prod_{j=1}^{i-1}(1-\mathbf{\alpha}_j'), \; \; \; \;
  \label{eq:alphablend}
\end{equation} 
$\mathbf{\alpha}_i'$ is the final opacity of the Gaussian obtained by: 

\begin{equation}
 \mathbf{\alpha}_i'= \mathbf{\alpha}_i *  e^{-\frac{1}{2}(\mathbf{y'}-\mathbf{x'})^T\mathbf{\Sigma'}^{-1}(\mathbf{y'}-\mathbf{x'})}. \; \; 
  \label{eq:opa}
\end{equation} 
where $\mathbf{x'}$ and $\mathbf{y'}$ are coordinates in the projected space. 

Although 3D Gaussians can be arbitrarily initialized, a good prior is often required for optimal results. Typically, the 3D Gaussians centers are initialized using the sparse Structure from Motion (SfM) point cloud obtained from the set of images. To avoid leaving huge holes in the scene, their corresponding covariances are initialized to have isotropic Gaussians with initial radii equal to the mean distance to neighboring points.
During training, the 3D Gaussian features are optimized such that the photometric difference between the rendered image $I_r$ and the ground truth one $I_{gt}$ is minimized with intervening controlling steps. To control the Gaussians' density and to address under- and over-distribution issues, authors use an adaptive densification step where Gaussians with large spatial gradients are split while transparent Gaussians that have very low opacity are pruned.

Although 3DGS~\cite{kerbl3Dgaussians} works well on object-centric and small scenes, it struggles with scenes presenting varying appearances and transient objects.
\section{Method}

In this section, we introduce \method, a novel 3DGS-based method for 3D scene reconstruction from \textit{in-the-wild} photo collections.
We propose to adapt the 3D Gaussians parameters to handle variable visual appearances and the presence of occluders typically found in such unconstrained image collections.

\begin{figure}[tb]
  \centering
  \includegraphics[width=.95\textwidth]{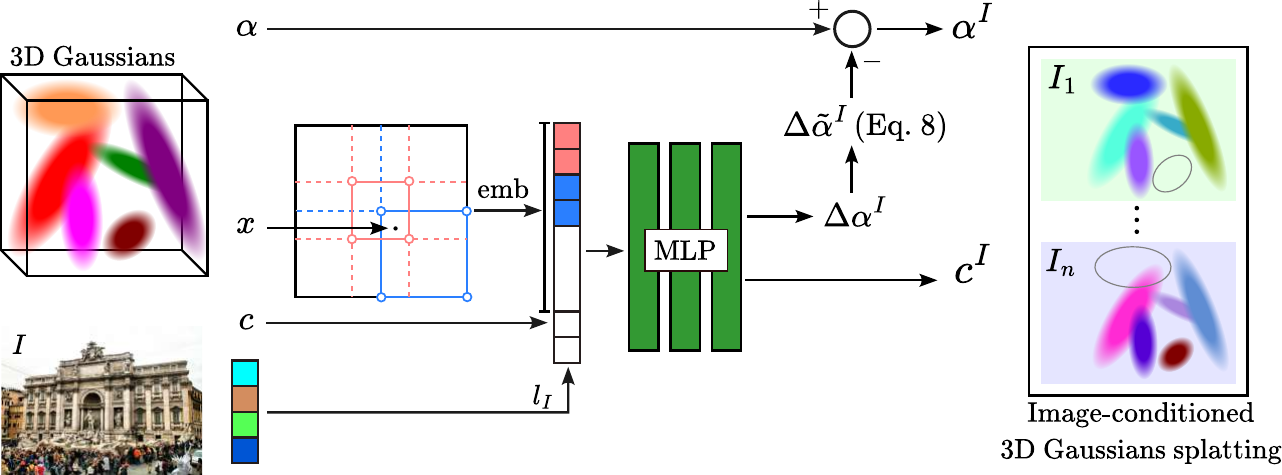}
  \caption{\textbf{SWAG model architecture --}
  In addition to the typical Gaussians' features, we also optimize a Hash Grid encoding their centers $\mathbf{emb(x)}$, a per-image embedding vector $l_I$ and an MLP. This MLP takes as inputs the Gaussians' colors $c$, the associated image embedding $l_I$, and their encoded centers $\mathbf{emb(x)}$ and outputs an image-dependent color $\mathbf{c^{I}}$ as well as an image-dependent opacity variation parameter $\Delta \alpha^I$. This parameter is set as the location variable of a concrete distribution which we sample to get the opacity variation $\Delta 
 \mathbf{\Tilde{\alpha}^{I}}$.
 Leveraging this opacity variation across diverse training images enables identifying and excluding transient Gaussians within the scene, as demonstrated by the grey Gaussians.}

  \label{fig:model}
\end{figure}

In~\Cref{sec:appearance}, we explain how \method~handles appearance variations using image-dependent embeddings injected into the Gaussian's colors. In~\Cref{sec:transient}, we explain how our proposal handles transient occluders by learning image-dependent Gaussians' opacities variations.
Since our model's knowledge about the presence of occluders is embedded in the Gaussians opacities variation, it easily enables a clear disentanglement between static and transient elements of the scene.
The overall architecture of our method is illustrated in~\Cref{fig:model}.  

\subsection{Appearance Variation Modeling}
\label{sec:appearance}

To adapt 3DGS to photometric variations, we associate, similarly to previous methods~\cite{martinbrualla2020nerfw, Turki_2022_CVPR}, a trainable embedding vector ${l_I}$ for each image. To inject the appearance information, one naive solution would be to calculate a global transformation per image to the 3D Gaussian colors or SH features. This strategy, however, might be insufficient since it cannot model photometric features at the local level in an image. To model local appearance variances in an image, one possible way would be to predict a local transformation or directly a specific color for each Gaussian on a per-image basis. However, such solution is intractable as the number of training parameters will grow according to the number of Gaussians times the number of images.

In \method, we propose a more effective solution where we use an MLP that takes as input a concatenation of the image embeddings and a positional encoding of the Gaussians' centers. By using a positional encoding conditioned on the center of the Gaussians, we are able to model local appearance variations that occur in the image while maintaining a tractable number of parameters to optimize. Initially, we considered estimating an affine color transformation similar to Urban Radiance Fields~\cite{rematas2021urban} to be applied to each Gaussian color. However, our initial results showed that affine color transformations cannot model all appearance changes.
To solve this problem, we directly predict the Gaussian color from the MLP. We noticed that spatial embeddings coupled with an MLP produce a smoothing effect on the predicted colors. Hence, to render view-dependent specularities and high-frequency details and counterbalance this smoothing effect, we also add the Gaussian color obtained from the $\mathbf{SH}$ coefficients as input to the MLP.

Our final solution employs an MLP $\mathcal{F}_{\mathbf{\theta}}$ that, for a given view, takes as input the color of the 3D Gaussian $\mathbf{c}$, the embedding vector of its center $\mathbf{emb(x)}$ and the image embedding $l_I$, and outputs the image-conditioned color $\mathbf{c^{I}}$. Accordingly, for a 3D Gaussian, the image-dependent color $\mathbf{c^I}$ is computed as follows:
\begin{equation}
  \mathbf{c^{I}}=\mathcal{F}_{\mathbf{\theta}}(\mathbf{c},\mathbf{emb(x)}, l_I).
  \label{eq:colorMLP}
\end{equation} 

\subsection{Transient Gaussians Modeling}
\label{sec:transient}
To address the challenge of handling transient objects, we introduce a learnable image-dependent opacity variation 
 term $\Delta 
 \mathbf{\Tilde{\alpha}^{I}}$ to each Gaussian.
On the one hand, this opacity variation parameter allows Gaussians to reconstruct occluders present in some images. On the other hand, it enables those same Gaussians to be transparent (i.e. not involved in the image rendering process in other images) where these occluders are absent.
During training, we want to encourage this variation parameter to either fully mask (i.e. in the case of occluders) or fully maintain (i.e. in the case of the static background) the 3D Gaussians' opacities.
For this purpose, we sample $\Delta 
 \mathbf{\Tilde{\alpha}^{I}}$ using a Binary Concrete random variable~\cite{maddison2017the}.
This distribution is a continuous approximation of a Bernoulli distribution concentrating most of the mass on the boundaries of the interval $\left[0, 1\right]$.
We return an additional output, $\Delta 
 \mathbf{{\alpha}^{I}}$, of the previously introduced MLP $\mathcal{F}_{\mathbf{\theta}}$ as the location parameter to the concrete function:
\begin{equation}
(\mathbf{c^{I}},\Delta\mathbf{{\alpha}^{I}})=\mathcal{F}_{\mathbf{\theta}}(\mathbf{c},\mathbf{emb(x)},l_I),
  \label{eq:fullMLPoutput}
\end{equation} 
\begin{equation}
 \Delta 
 \mathbf{\Tilde{\alpha}^{I}}=\text{sigmoid} \left[ \frac{1}{\mathbf{T}} \left ( log \left ( | \Delta 
 \mathbf{{\alpha}^{I}}| \right ) +log(\mathbf{U})-log(1-\mathbf{U}) \right ) \right ],
  \label{eq:concreteFunction}
\end{equation}

\begin{equation}
  \mathbf{U} \sim \text{Uniform}(0,1),
  \label{eq:uniform}
\end{equation} 

\noindent where $\mathbf{T}$ is the temperature hyper-parameter. We do not use any annealing on the temperature parameter, rather we set $\mathbf{T}=0.1$ to force from the very beginning of the training the opacity variation to match a binary distribution. During the evaluation, we fix $\mathbf{U}$ at $0.5$. 
The final image-dependent 3D Gaussian opacity is formulated as:
\begin{equation}
  \mathbf{{\alpha}^I} =\max \left ( \mathbf{{\alpha}} - \Delta  \mathbf{\Tilde{\alpha}^{I}}, 0 \right ).
  \label{eq:finalOpacity}
\end{equation} 
 
\begin{figure}[tb] 
\centering

  \includegraphics[width=.90\textwidth]{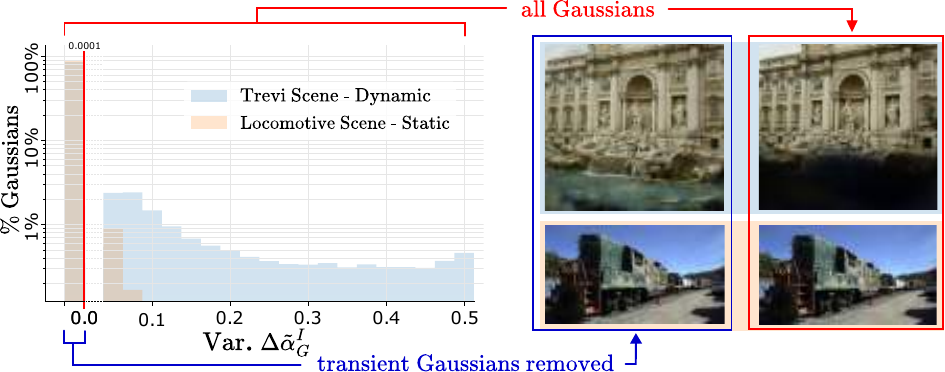}

\caption{ Variance histogram analysis of the Gaussians' opacity variation $\Delta 
\mathbf{\Tilde{\alpha}^{I}}$  w.r.t training images for a dynamic (i.e. containing transient objects) scene (top, Trevi Fountain from Phototourism \cite{Jin_2020}); and a static scene (bottom,  Locomotive from Tanks\&Temples~\cite{Knapitsch2017TanksAT}).
The left column views are rendered using only static Gaussians (i.e having $\mathbf{Var} \left [ \Delta\mathbf{\Tilde{\alpha}^{I}} \right ] = 0$) whereas right column views are rendered using all Gaussians.} 
\label{fig:variance}

\end{figure}

With our formulation of the opacity, we can naturally disentangle between the transient and the static parts of the scene by defining transient Gaussians as the ones having non-zero variance of their associated parameter $\Delta  \mathbf{\Tilde{\alpha}^{I}}$ over training images: $\mathbf{Var} \left [ \Delta\mathbf{\Tilde{\alpha}^{I}} \right ] \neq 0$.
\Cref{fig:variance} shows the histogram of $\mathbf{Var} \left [ \Delta\mathbf{\Tilde{\alpha}^{I}} \right ]$ on two scenes: one without transient content (\textit{Locomotive}) and one with occluders (\textit{Trevi}). We also show two rendered images of each scene: the right image using all the Gaussians and the left one using only Gaussians not detected as transient. The histogram shows that for a static scene, we detected less than $4\%$ of the Gaussians as transient and around $96\%$ of the Gaussians having $\mathbf{Var} \left [ \Delta\mathbf{\Tilde{\alpha}^{I}} \right ] =0$. Removing the $4\%$ of transient Gaussians does not affect the rendered image quality. For the \textit{Trevi Fountain} scene, our model was able to represent transient objects using less than $20\%$ of the Gaussians and to maintain the majority of the Gaussians to reconstruct static parts of the scene. Comparing the two rendered images shows that the transient Gaussians correspond to the mass of tourists present near the fountain and removing the transient Gaussians guaranteed an occluders-free rendering.

\section{Experiments}
\subsection{Implementation details}
Similarly to 3DGS, we minimize the $\mathcal{L}_1$ loss combined with D-SSIM term to optimize the 3D Gaussians parameters alongside with the MLP $\mathcal{F}_{\theta}$ weights, the Hash Encoder parameters and the per-image embedding vectors. We fix the size of the latent embedding vectors to $n=24$. For the Hash encoding parameters, we set the hash table size $T=2^{19}$, the finest resolution $N_{max}=2048$, the number of levels $L=12$ and the coarsest resolution $N_{min}=16$. The MLP we employ has 3 layers of 64 hidden units. See the supplement for additional details on hyperparameters.
\subsection{Datasets}

Similar to SoTA \textit{in-the-wild} reconstruction methods, we evaluate \method's NVS capabilities using the Phototourism dataset~\cite{Jin_2020} and report our results on three main touristic monuments: \textit{Brandenburg Gate}, \textit{Sacre Coeur}, and \textit{Trevi Fountain}.
We also evaluate our method on the benchmark introduced in NeRF-OSR~\cite{rudnev2022nerfosr} for outdoor scene relighting tasks.
We report our performance using NeRF-MS's~\cite{Li_2023_ICCV} data split on 4 sites: \textit{stjohann (St. Johann, Saarbrücken)}, \textit{lwp (Landwehrplatz, Saarbrücken)}, \textit{st (Staatstheater, Saarbrücken)}, and \textit{europa (Galerie Europa, Saarbrücken)}  and the original NeRF-OSR~\cite{rudnev2022nerfosr}'s data split on 3 sites: \textit{stjohann}, \textit{lwp} and \textit{lk2 (Ludwigskirche, Saarbrücken)}.
More details about the datasets can be found in supplementary materials.

\begin{table}[bt]
    \caption{\textbf{Quantitative results --} Results on three real-world scenes from Phototourism~\cite{Jin_2020} and efficiency comparison among SoTA methods for NVS \textit{in-the-wild}. Given the significant gap between previous baselines~\cite{chen2022hallucinated,martinbrualla2020nerfw,yang2023crossray,kassab2024refinedfields} and 3DGS~\cite{kerbl3Dgaussians}, we marginalize on the different GPU configurations. %
    }
    \label{tab:quantitative1}
    \centering
    \resizebox{1.0\columnwidth}{!}
    {
        \renewcommand{\arraystretch}{0.5}
        \setlength{\tabcolsep}{0.0035\linewidth}
        \begin{tabular}{lccccccccccccc}
            \cmidrule[\heavyrulewidth]{2-10}
            \cmidrule[\heavyrulewidth]{12-13}
             & \multicolumn{3}{c}{Bradenburg Gate}  & \multicolumn{3}{c}{Sacre Coeur} & \multicolumn{3}{c}{Trevi Fountain} & ~~~  
             & \multicolumn{2}{c}{Mean Efficiency}  &\\
            \cmidrule(rl){2-4}
            \cmidrule(rl){5-7}
            \cmidrule(rl){8-10}
            \cmidrule(rl){12-13}
            
             & PSNR~$\uparrow$  & SSIM$~\uparrow$  & LPIPS~$\downarrow$ & PSNR~$\uparrow$  & SSIM~$\uparrow$  & LPIPS~$\downarrow$ & PSNR~$\uparrow$  & SSIM~$\uparrow$  & LPIPS~$\downarrow$  & ~~~
              & Train time~(h) & FPS\\
                    
            \cmidrule{0-9}
            \cmidrule{12-13}
    
            NeRF~\cite{mildenhall2020nerf}  
            &18.90 & 0.815 & 0.231 &  
            15.60 &   0.715 &0.291&
            16.14 & 0.600 & 0.366 & ~~~ & - & -\\

            NeRF-W~\cite{martinbrualla2020nerfw} 
            &  24.17 & \cellcolor{yellow!25}{0.890}& 0.167&
            19.20 & 0.807& 0.191 &
             18.97 &0.698&0.265 & ~~~ & 400$^{\dag}$ & <1$^{\dag}$  \\

            Ha-NeRF~\cite{chen2022hallucinated}&  
            24.04 & 0.877 &\cellcolor{orange!25}{0.139} &
            20.02 & 0.801 & \cellcolor{orange!25}{0.171} &
            20.18 & 0.690 & \cellcolor{yellow!25}{0.222} & ~~~ & 452$^{\dag}$ & 0.20$^{\dag}$  \\

            CR-NeRF~\cite{yang2023crossray} 
            &  \cellcolor{orange!25}{26.53}& \cellcolor{orange!25}{0.900} & \cellcolor{red!25}{0.106} & 
            \cellcolor{orange!25}22.07 & \cellcolor{yellow!25}{0.823} & \cellcolor{red!25}{0.152} &
            \cellcolor{yellow!25}21.48&0.711 &\cellcolor{red!25}{0.206} & ~~~ & 420$^{\dag}$ & 0.25$^{\dag}$ \\

            RefinedFields~\cite{kassab2024refinedfields}~~  
            &  \cellcolor{red!25}{26.64} &0.886 & - & 
             \cellcolor{red!25}{22.26} & 0.817  & -
            & \cellcolor{red!25}{23.42} & \cellcolor{yellow!25}{0.737}& - & ~~~ & \cellcolor{yellow!25}{150$^{\dag}$} & \cellcolor{yellow!25}{<1$^{\dag}$}\\
    
            3DGS~\cite{kerbl3Dgaussians} 
            & 19.99  & 0.889 &0.180  
            & 17.57 &\cellcolor{orange!25}{0.831} &0.219 
            &18.47  &\cellcolor{orange!25}{0.761} &0.234  & ~~~ & \cellcolor{red!25}{0.50$^{*}$} & \cellcolor{red!25}{181$^{*}$}  \\

            \cmidrule{0-9}
            \cmidrule{12-13}

            SWAG~(ours) 
                &\cellcolor{yellow!25}{26.33}   & \cellcolor{red!25}{0.929} & \cellcolor{orange!25}{0.139}
           
            & \cellcolor{yellow!25}{21.16} & \cellcolor{red!25}{0.860}&\cellcolor{yellow!25}{0.185}
            & \cellcolor{orange!25}{23.10}  & \cellcolor{red!25}{0.815} & \cellcolor{orange!25}{0.208} & ~~~ & \cellcolor{orange!25}{0.83$^{*}$} & \cellcolor{orange!25}{15.29$^{*}$} \\   
            
            \cmidrule[\heavyrulewidth]{0-9}
            \cmidrule[\heavyrulewidth]{12-13}
        \end{tabular}
    }
    
    {\scriptsize $^{\dag}$ Results reported on \cite{kassab2024refinedfields}.  $^{*}$ Results computed using a high-tier GPU.} \\
\end{table}

\begin{table}[tb]
    \caption{\textbf{Quantitative results --} Results on five sites from NeRF-OSR \cite{rudnev2022nerfosr} outdoor scenes benchmark. %
    }
    \label{tab:quantitatives}
    \resizebox{1.0\columnwidth}{!}{%
    \renewcommand{\arraystretch}{0.5}
    \setlength{\tabcolsep}{0.0035\linewidth}
    \centering
    \begin{tabular}{lccccccccccccc}
        \cmidrule[\heavyrulewidth]{2-13}
         & \multicolumn{3}{c}{stjohann}  & \multicolumn{3}{c}{lwp} & \multicolumn{3}{c}{st} 
         & \multicolumn{3}{c}{europa} \\
        \cmidrule(rl){2-4}
        \cmidrule(rl){5-7}
        \cmidrule(rl){8-10}
        \cmidrule(rl){11-13}
        
             & PSNR~$\uparrow$  & SSIM~$\uparrow$  & LPIPS~$\downarrow$ & PSNR~$\uparrow$  & SSIM~$\uparrow$  & LPIPS~$\downarrow$ & PSNR~$\uparrow$  & SSIM~$\uparrow$  & LPIPS~$\downarrow$ & PSNR~$\uparrow$  & SSIM$~\uparrow$  & LPIPS~$\downarrow$ \\
                
        \midrule
        NeRF~\cite{mildenhall2020nerf}

        &14.89 &0.432 &0.639 &
        11.51& 0.468 &0.574 &
        17.20& 0.514 &0.502 &
        17.49 &0.551 &0.503\\
        
        NeRF-W~\cite{martinbrualla2020nerfw}

        &\cellcolor{yellow!25}{21.23} &0.667 &0.426 &
        19.61& 0.616 &0.445&
        \cellcolor{yellow!25}{20.31} &0.607& 0.438& 
        20.00 &0.699& 0.340\\

        Ha-NeRF~\cite{chen2022hallucinated}
        &17.19 &0.686& 0.331&
        \cellcolor{yellow!25}{20.03}&\cellcolor{yellow!25}{0.685}& \cellcolor{yellow!25}{0.365} &
        17.30& 0.538& 0.483& 
        17.79 &0.632& 0.421\\

        NeRF-MS~\cite{Li_2023_ICCV} ~
        &\cellcolor{orange!25}{22.84} &\cellcolor{orange!25}{0.793} &\cellcolor{orange!25}{0.235}&
        \cellcolor{orange!25}{21.90} &\cellcolor{orange!25}{0.719}& \cellcolor{orange!25}{0.336}&
        \cellcolor{orange!25}{20.68} &\cellcolor{orange!25}{0.630}& \cellcolor{orange!25}{0.402} &
        \cellcolor{orange!25}{21.03} &\cellcolor{yellow!25}{0.721} &\cellcolor{yellow!25}{0.294}\\

        3DGS~\cite{kerbl3Dgaussians} 
        &16.77 &\cellcolor{yellow!25}{0.741} &\cellcolor{yellow!25}{0.268}&
        11.76 &0.609& 0.414&
        17.16 &\cellcolor{yellow!25}{0.629}& \cellcolor{yellow!25}{0.406} &
        \cellcolor{yellow!25}{20.18} &\cellcolor{orange!25}{0.782}& \cellcolor{orange!25}{0.252}&\\
        
        \midrule

        SWAG~(ours) 
           &\cellcolor{red!25}{23.91} &\cellcolor{red!25}{0.864} &\cellcolor{red!25}{0.172}&
        \cellcolor{red!25}{22.07} &\cellcolor{red!25}{0.783}& \cellcolor{red!25}{0.303}&
         \cellcolor{red!25}{22.29} & \cellcolor{red!25}{0.713}&  \cellcolor{red!25}{0.364} &
        \cellcolor{red!25}{23.74} &\cellcolor{red!25}{0.845} &\cellcolor{red!25}{0.242}\\
        \bottomrule        \\
        \multicolumn{6}{@{}l}{\em(a) Reported on NeRF-MS split~\cite{Li_2023_ICCV}}\\
    \end{tabular}
    }

    \vspace{0.2cm}
    \centering
    \resizebox{0.58\columnwidth}{!}{%
    \renewcommand{\arraystretch}{0.5}
    \setlength{\tabcolsep}{0.0035\linewidth}
    
    \begin{tabular}{lccccccc}
        \cmidrule[\heavyrulewidth]{2-7}
         & \multicolumn{2}{c}{lk2}  & \multicolumn{2}{c}{st} & \multicolumn{2}{c}{lwp}   \\
        \cmidrule(rl){2-3}
        \cmidrule(rl){4-5}
        \cmidrule(rl){6-7}
        
             & PSNR $\uparrow$  & SSIM $\uparrow$   & PSNR $\uparrow$  & SSIM $\uparrow$  & PSNR $\uparrow$  & SSIM $\uparrow$   \\

        \midrule
        NeRF-OSR~\cite{rudnev2022nerfosr} 
        &\cellcolor{red!25}19.86 & \cellcolor{yellow!25}0.626 &  
                    \cellcolor{orange!25}15.83&   \cellcolor{yellow!25}0.556&
                    \cellcolor{orange!25}17.38  &\cellcolor{yellow!25}0.576 \\
        3DGS~\cite{kerbl3Dgaussians}   
        & \cellcolor{yellow!25}15.91  &\cellcolor{orange!25}0.687 
        & \cellcolor{yellow!25}12.53 &\cellcolor{orange!25}0.585  
        &\cellcolor{yellow!25}13.72& \cellcolor{orange!25}0.659  \\
        \midrule
        SWAG~(ours) 
        & \cellcolor{orange!25}19.59  &\cellcolor{red!25}0.756  
        & \cellcolor{red!25}18.73  &\cellcolor{red!25}0.721  
        & \cellcolor{red!25}18.61  &\cellcolor{red!25}0.753 \\
        
        \bottomrule
        \\
        \multicolumn{7}{@{}l}{\em(b) Reported on original benchmark split from NeRF-OSR~\cite{rudnev2022nerfosr}}\\
    \end{tabular}
    }

\end{table}

\subsection{Evaluation}
We provide visual comparisons with rendered images and report quantitative results based on common rendering metrics from the literature: Peak Signal-to-Noise Ratio (PSNR), Structural Similarity Index Measure (SSIM), and Learned Perceptual Image Patch Similarity (LPIPS).
Since only the embeddings associated with the training images are optimized during training, we follow NeRF-W's~\cite{martinbrualla2020nerfw} evaluation approach: we optimize an embedding on the left half of each test image and report metrics on the right half.

\subsection{Results}
\Cref{tab:quantitative1} shows quantitative results on three Phototourism scenes. Unsurprisingly, optimizing 3DGS on \textit{in-the-wild} photo collections showed poor results due to the lack of image appearance dependency. The 3D Gaussians colors converge to a mean color which makes the method unable to model visual changes caused by weather, lightning conditions and different camera specifications. The presence of transient objects also resulted in the persistence of artifacts. \method~improves 3DGS performance and outperforms the baselines on SSIM across all datasets. In particular, our proposal improves 3DGS's PSNR by an average margin of 5.01~dB. SWAG outperforms NeRF-W across all datasets in terms of LPIPS, PSNR, and SSIM and shows competitive performance with CR-NeRF and RefinedFields despite the priors these two methods use and requiring significantly longer times to train.

\Cref{tab:quantitatives} shows quantitative results on NeRF-OSR~\cite{rudnev2022nerfosr} benchmark. \method~improves 3DGS performance by an average PSNR margin of 5~dB and outperformed all the other baselines regardless of the train/test split choices.

\begin{figure*}[!h] 
	\centering
	\scriptsize
	\setlength{\tabcolsep}{0.002\linewidth}
	\renewcommand{\arraystretch}{0.8}
	\begin{tabular}{cccc}

        \multirow{2}{*}[8mm]{\rotatebox[origin=c]{90}{Brandenburg Gate}}  &
        \includegraphics[clip=true, trim={0 70 0 100},width=0.265\textwidth]{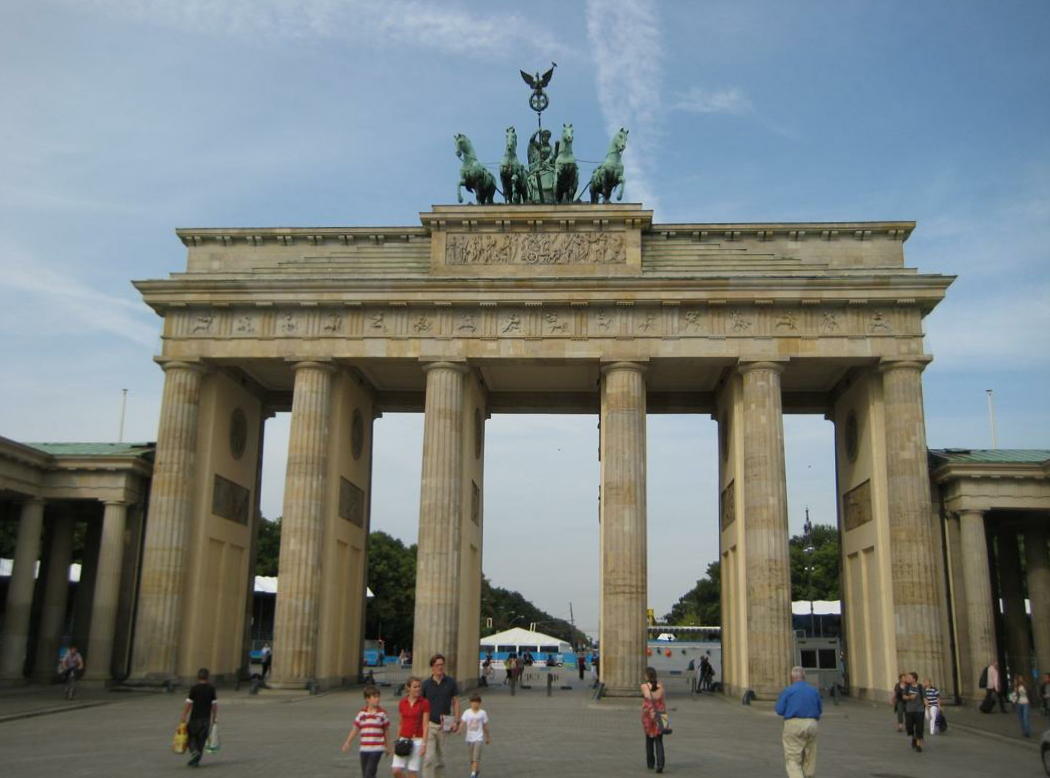} & 
        \includegraphics[clip=true, trim={0 70 0 100},width=0.265\textwidth]{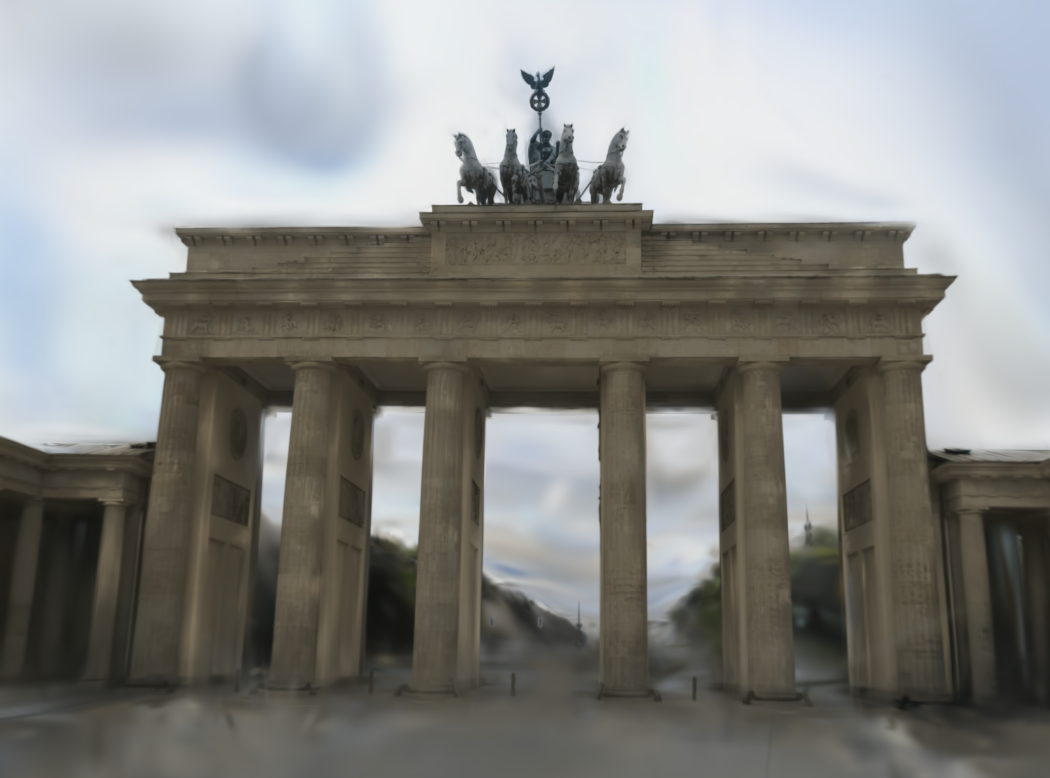} & 
        \includegraphics[clip=true, trim={0 70 0 100},width=0.265\textwidth]{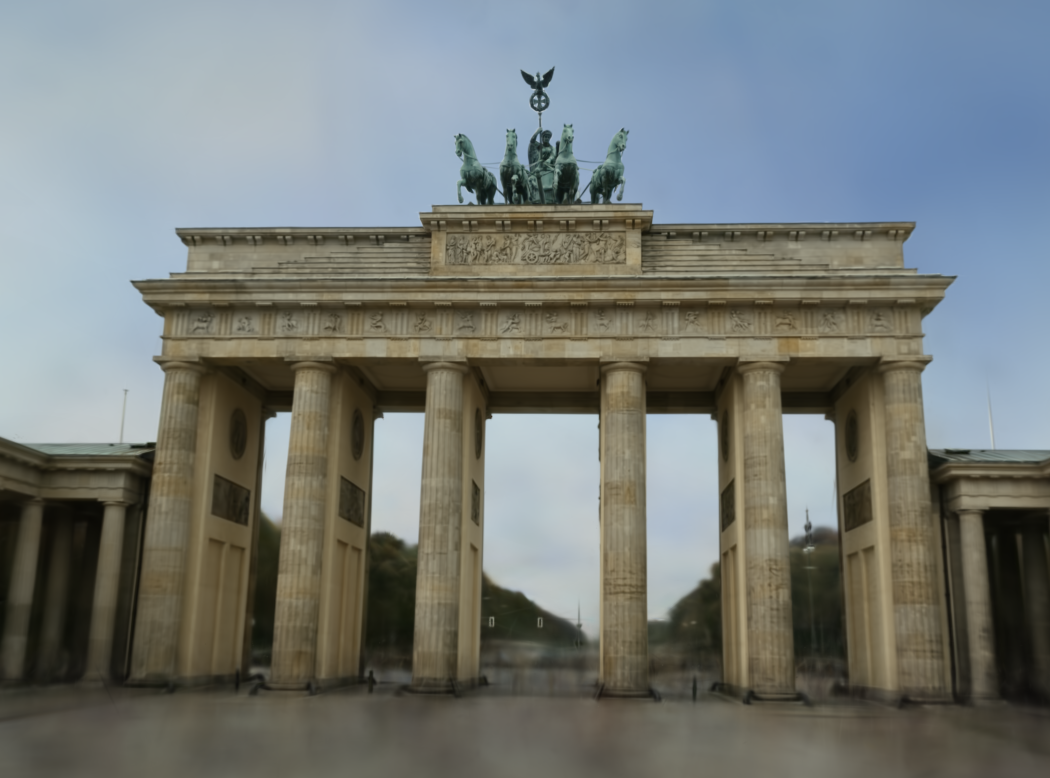}
        \\
        &
        \includegraphics[clip=true, trim={0 35 0 88},width=0.265\textwidth]{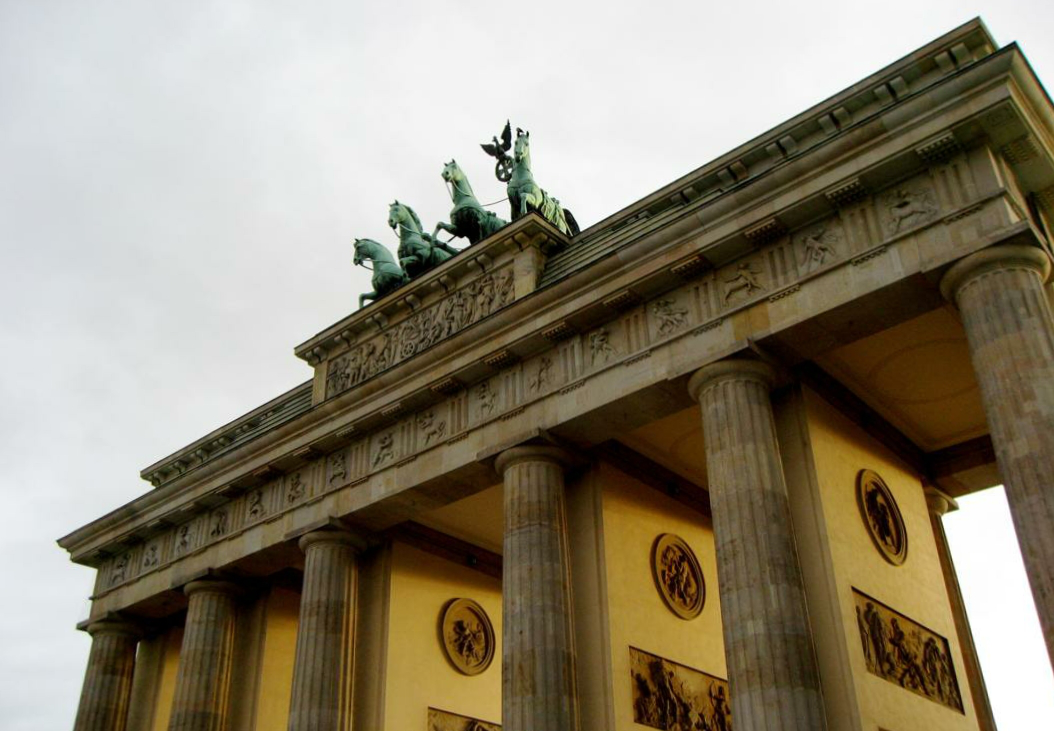} & 
        \includegraphics[clip=true, trim={0 35 0 88},width=0.265\textwidth]{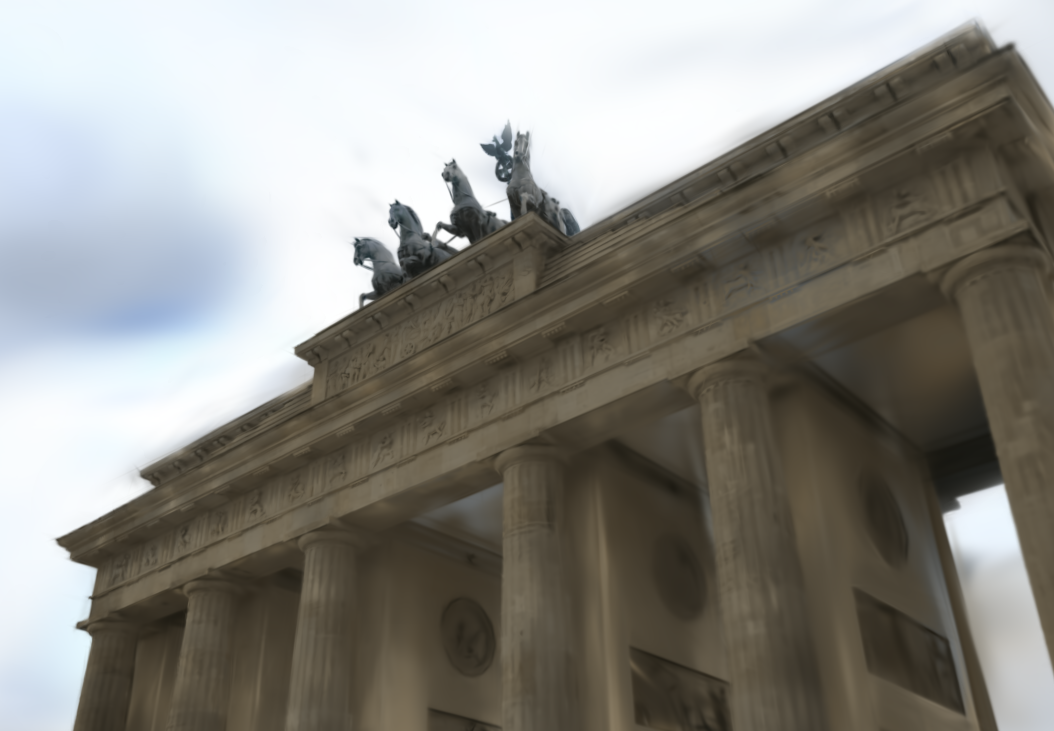} & 
        \includegraphics[clip=true, trim={0 35 0 88},width=0.265\textwidth]{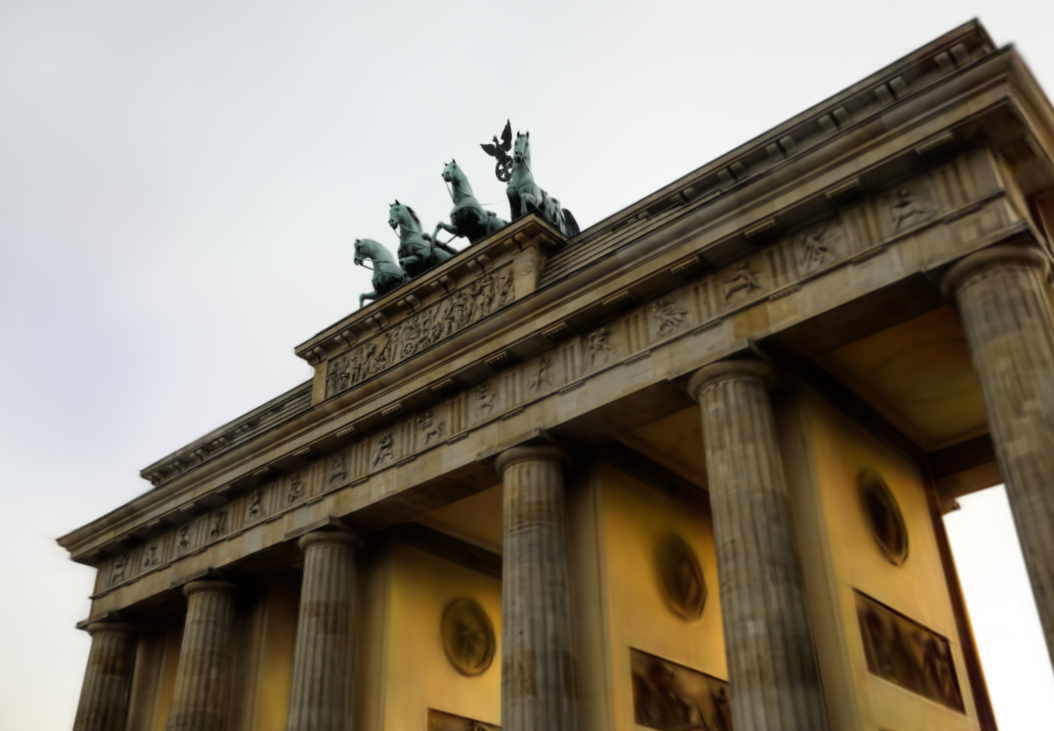}
        \\

        \cmidrule(rl){2-4}
         \multirow{2}{*}[6mm]{\rotatebox[origin=c]{90}{Sacre Coeur}}  & 
         \includegraphics[clip=true, trim={0 90 0 80},width=0.265\textwidth]{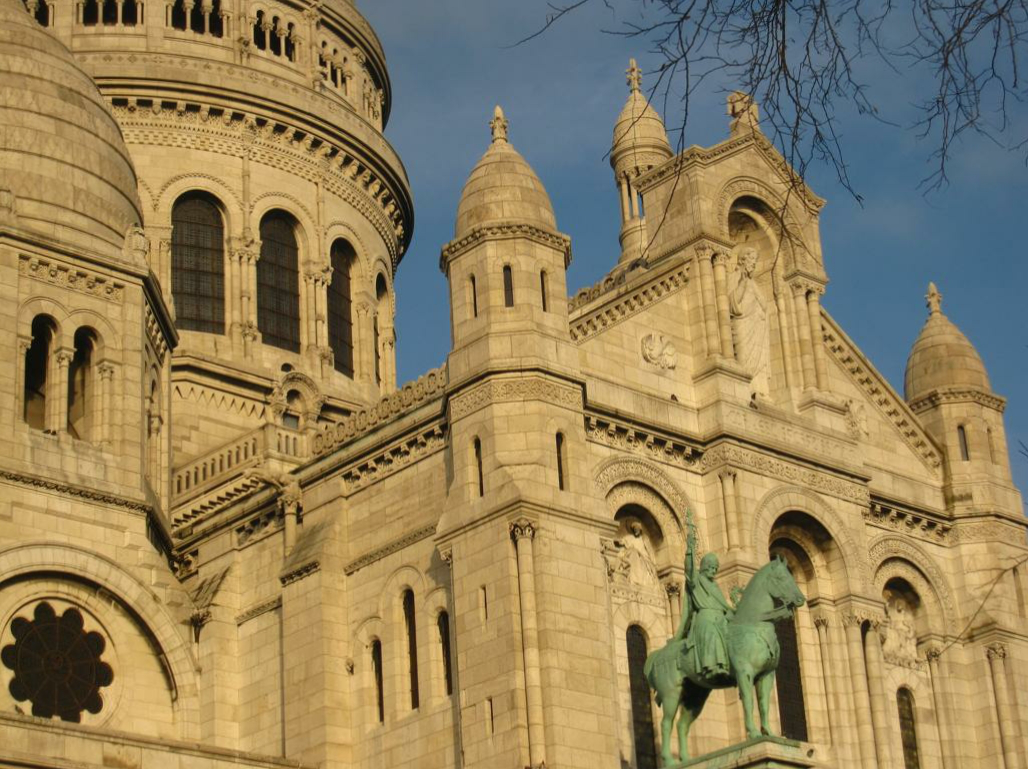} & 
         \includegraphics[clip=true, trim={0 90 0 80},width=0.265\textwidth]{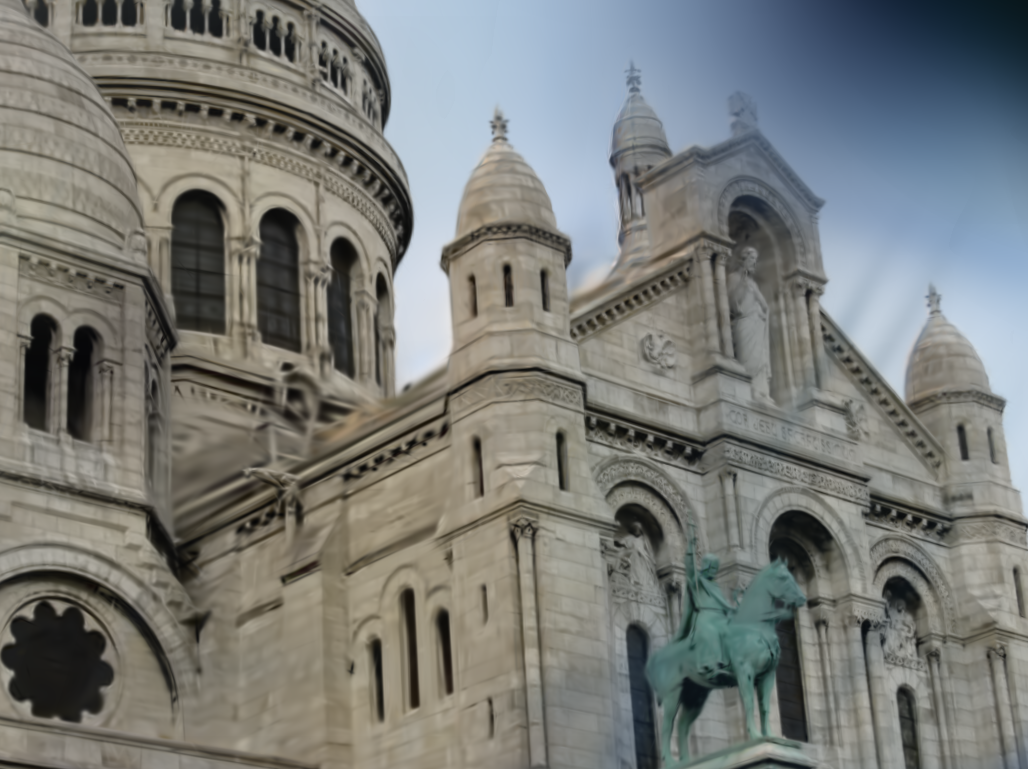} & 
         \includegraphics[clip=true, trim={0 90 0 80},width=0.265\textwidth]{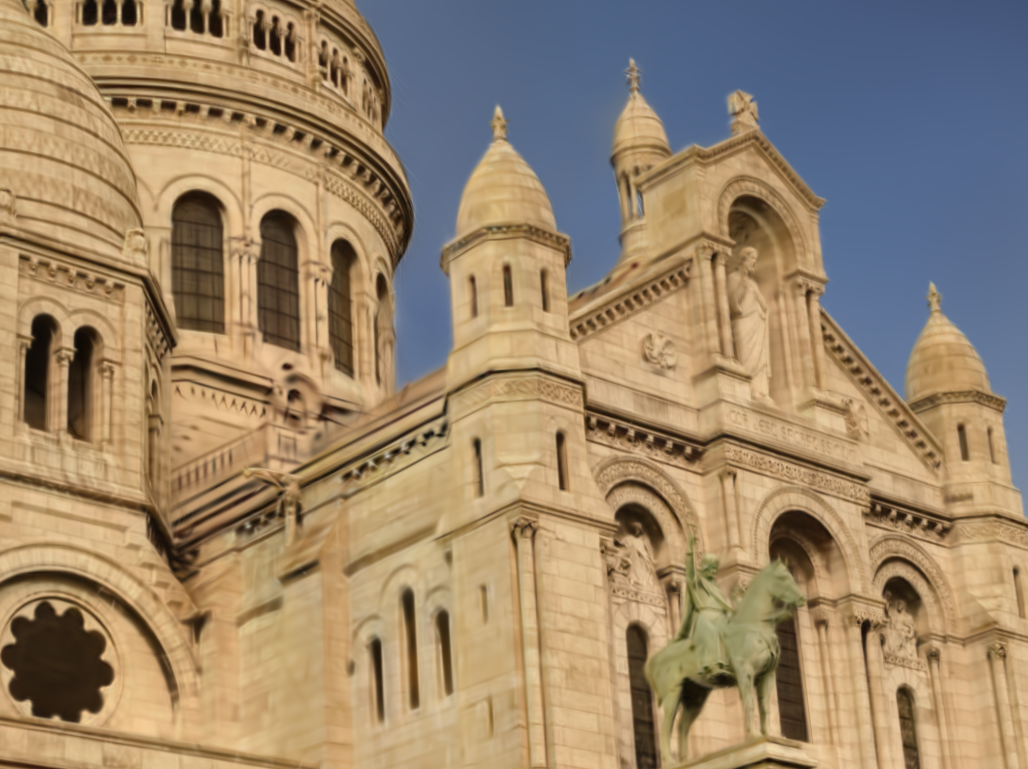}
         \\
         &
         \includegraphics[clip=true, trim={0 0 0 209},width=0.265\textwidth]{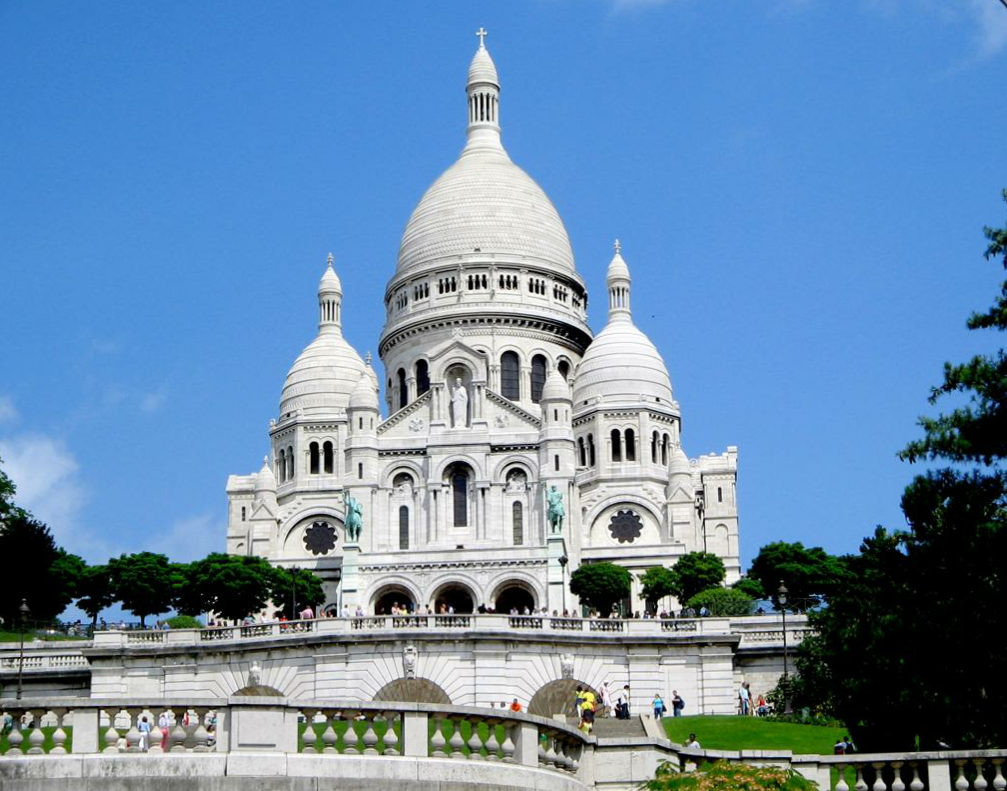} 
         & 
          \includegraphics[clip=true, trim={0 0 0 209},width=0.265\textwidth]{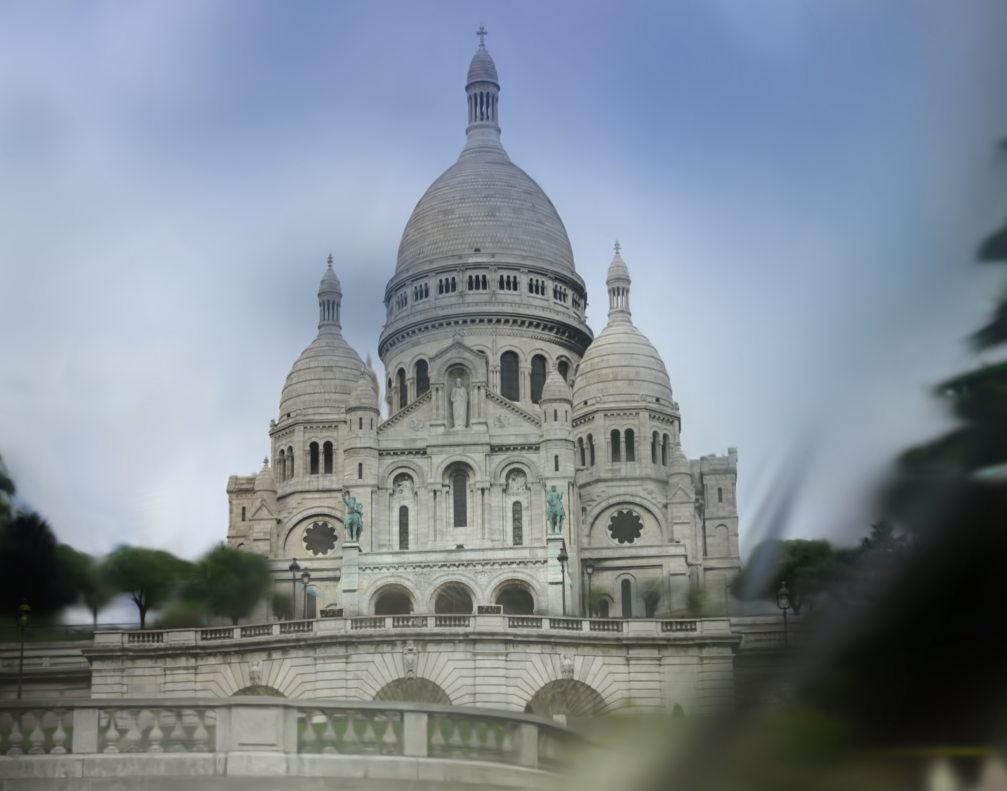} &
        
        \includegraphics[clip=true, trim={0 0 0 209},width=0.265\textwidth]{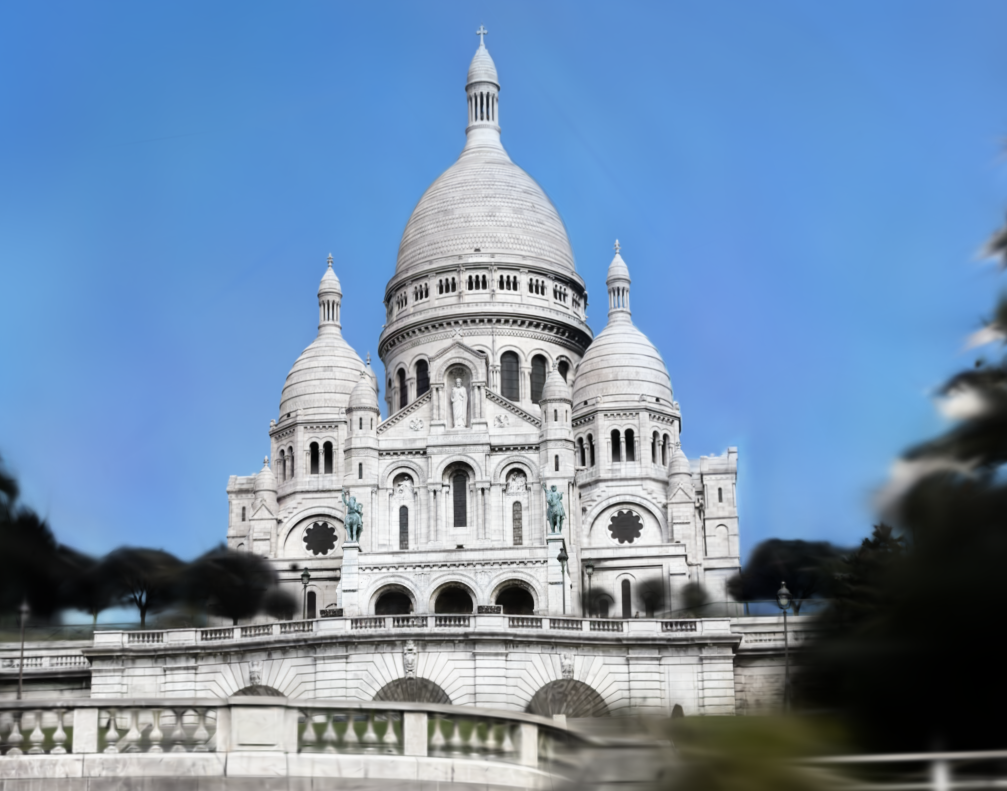}
         \\
        \cmidrule(rl){2-4}
        \multirow{2}{*}[6mm]{\rotatebox[origin=c]{90}{Trevi Fountain}} & \includegraphics[clip=true, trim={0 290 0 305},width=0.265\textwidth]{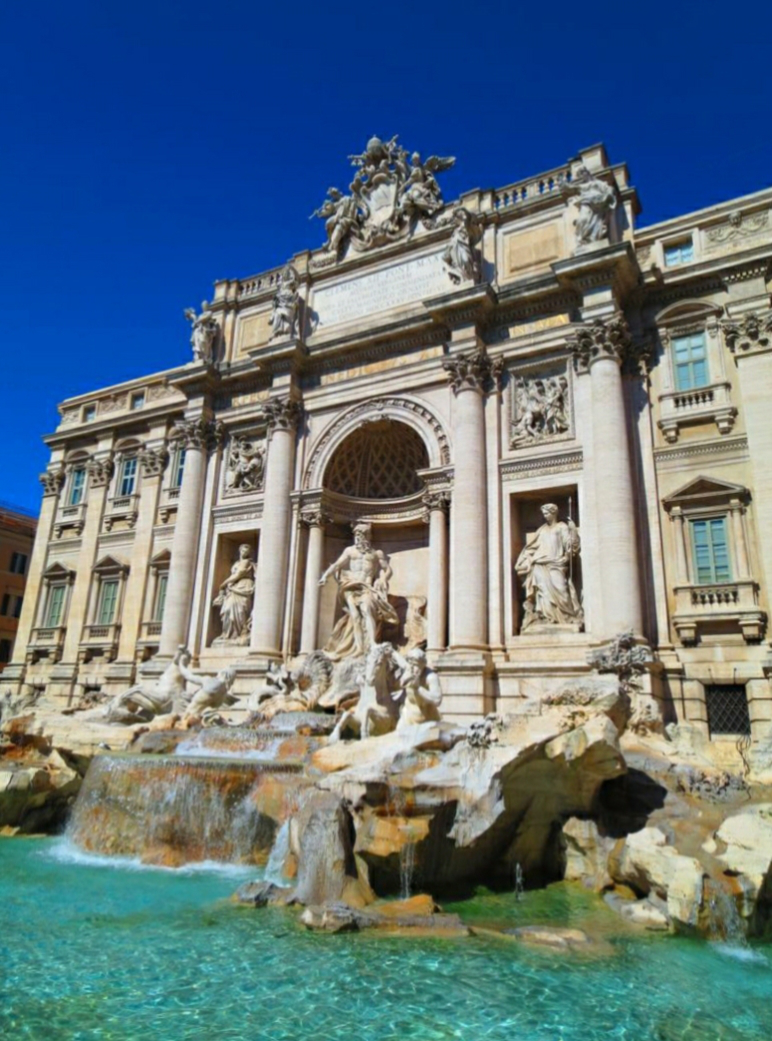} & 
        \includegraphics[clip=true, trim={0 290 0 305},width=0.265\textwidth]{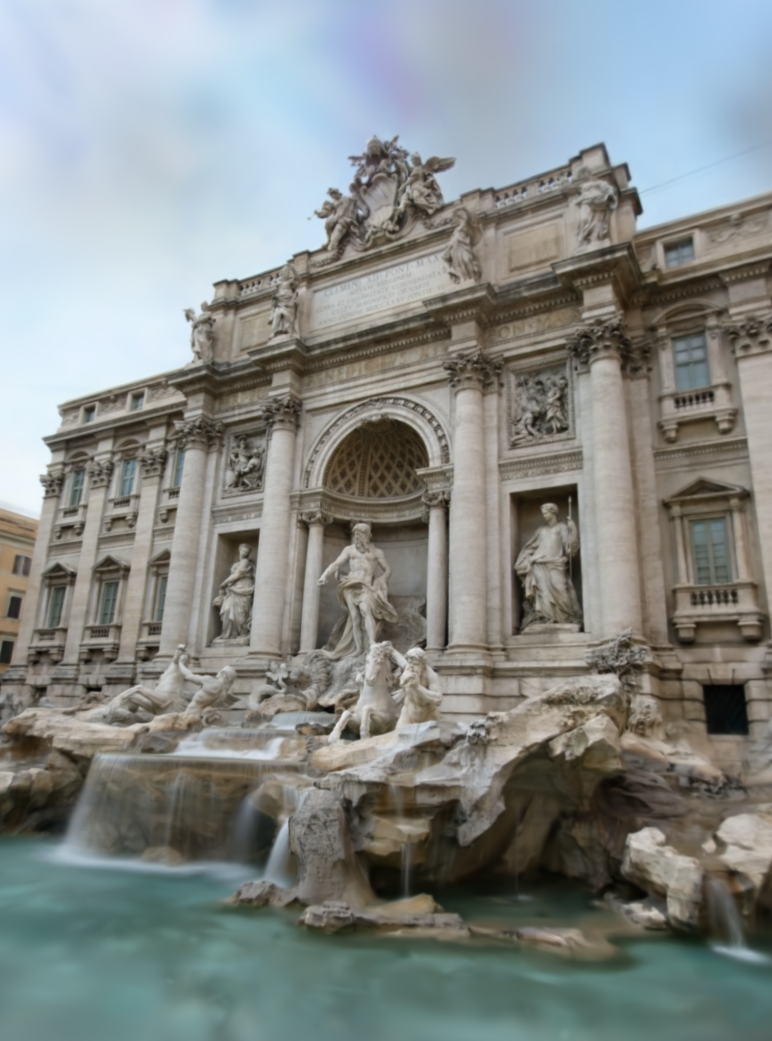} & 
        \includegraphics[clip=true, trim={0 290 0 305}  ,width=0.265\textwidth]{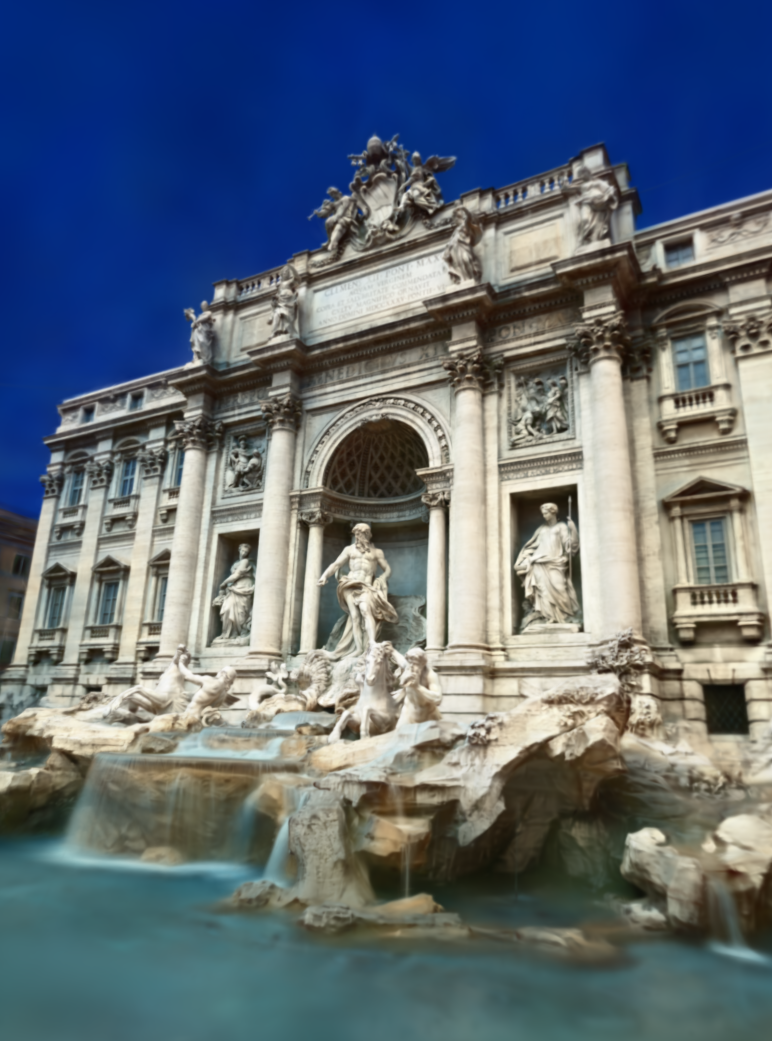}
        \\
         & \includegraphics[clip=true, trim={0 220 0 380},width=0.265\textwidth]{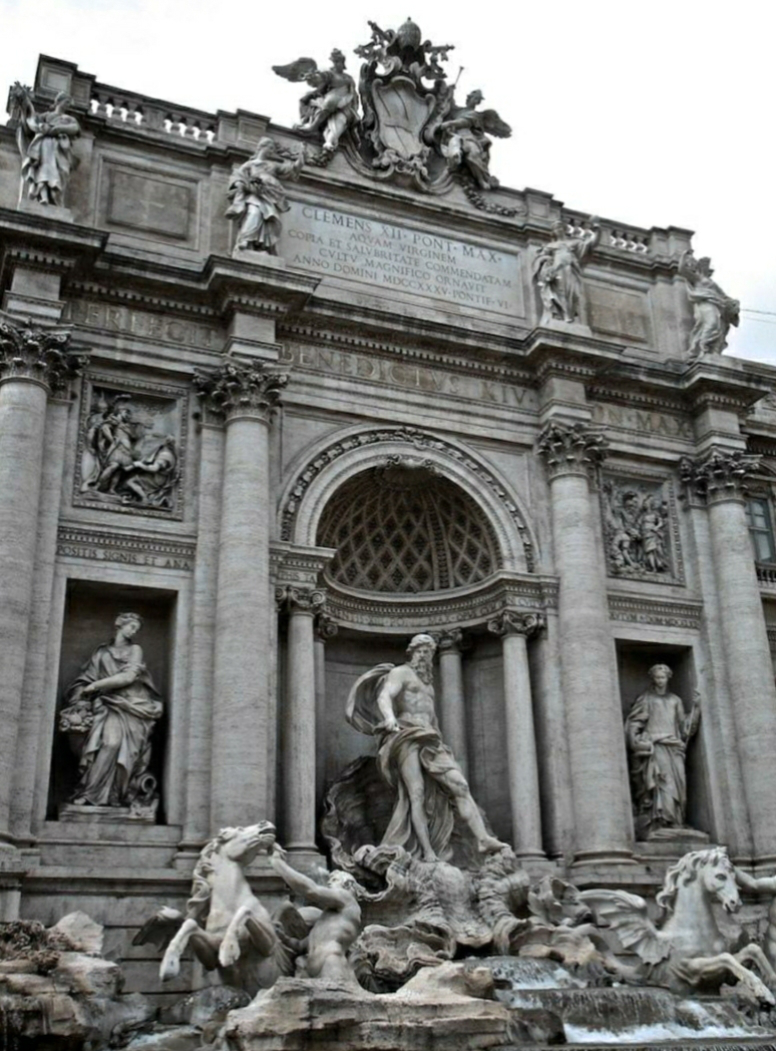} & 
        \includegraphics[clip=true, trim={0 220 0 380},width=0.265\textwidth]{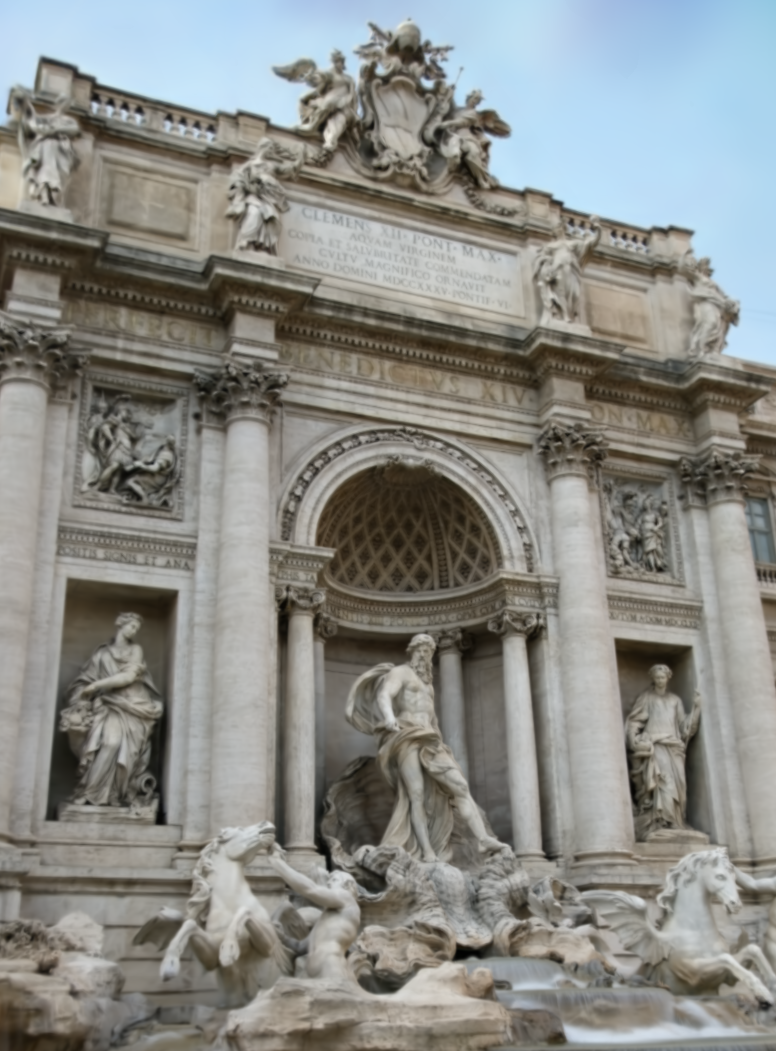} & 
        \includegraphics[clip=true, trim={0 220 0 380}  ,width=0.265\textwidth]{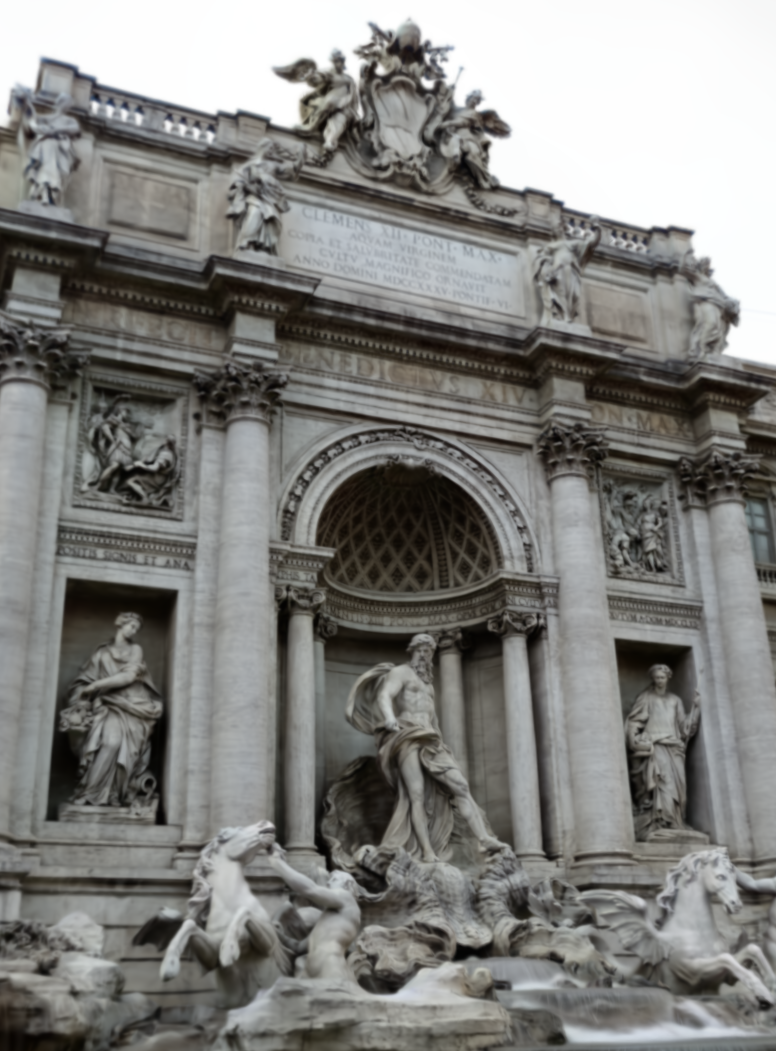} 
        \\
            
        & Ground-truth & 3DGS~\cite{kerbl3Dgaussians} & SWAG~(ours) \\

	\end{tabular}
	\caption{Qualitative experimental results on three real-world scenes from Phototourism~\cite{Jin_2020}.
    }
	\label{fig:fig9}
\end{figure*}

\begin{figure*}[!h] 
	\centering
	\scriptsize
	\setlength{\tabcolsep}{0.002\linewidth}
	\renewcommand{\arraystretch}{0.8}
	\begin{tabular}{ccccc}

        \multirow{3}{*}[7mm]{\rotatebox[origin=c]{90}{Phototourism Dataset~\cite{Jin_2020}}}
        & \includegraphics[clip=true,trim={0 0 0 226},width=0.23\textwidth]{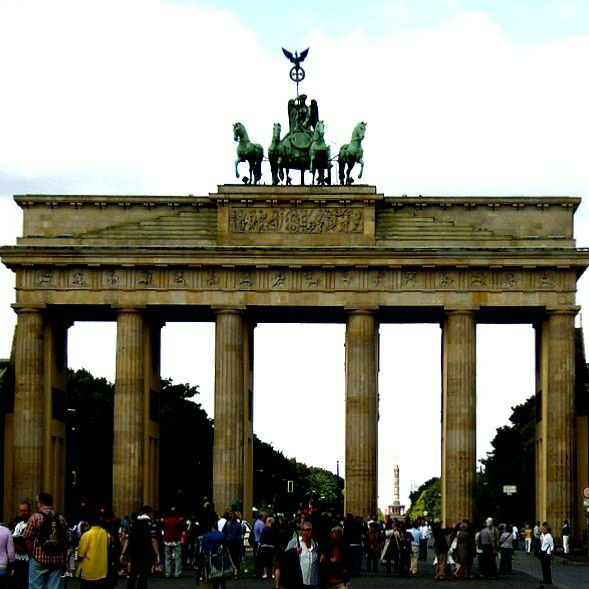} &
        \includegraphics[clip=true,trim={0 0 0 226},width=0.23\textwidth]{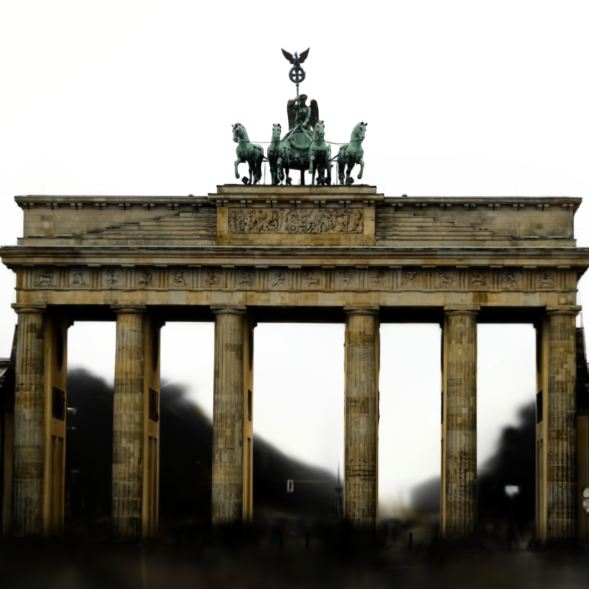} &
        \includegraphics[clip=true,trim={0 0 0 226},width=0.23\textwidth]{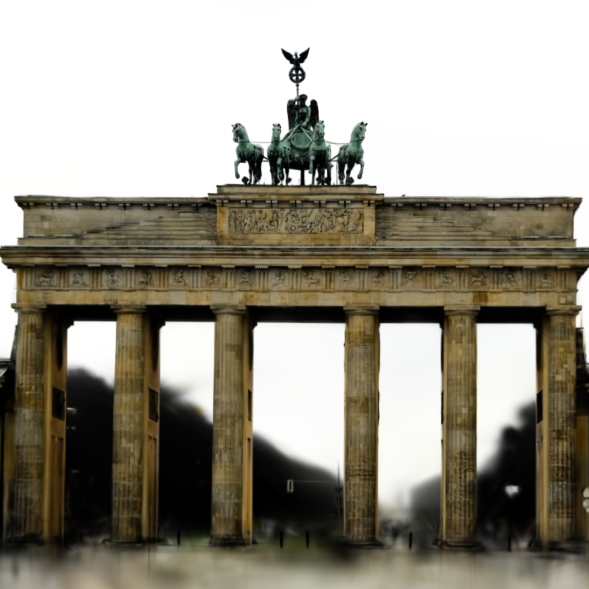} &
        \includegraphics[clip=true,trim={0 0 0 226},width=0.23\textwidth]{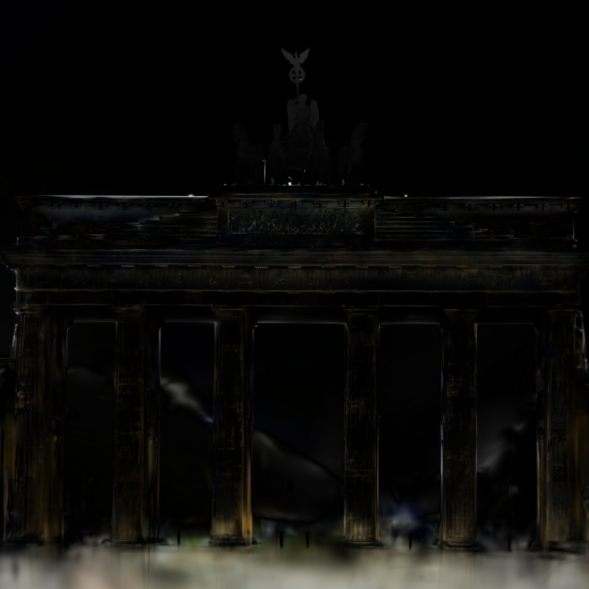}
        \\
        & \includegraphics[clip=true,trim={35 0 55 0},width=0.23\textwidth]{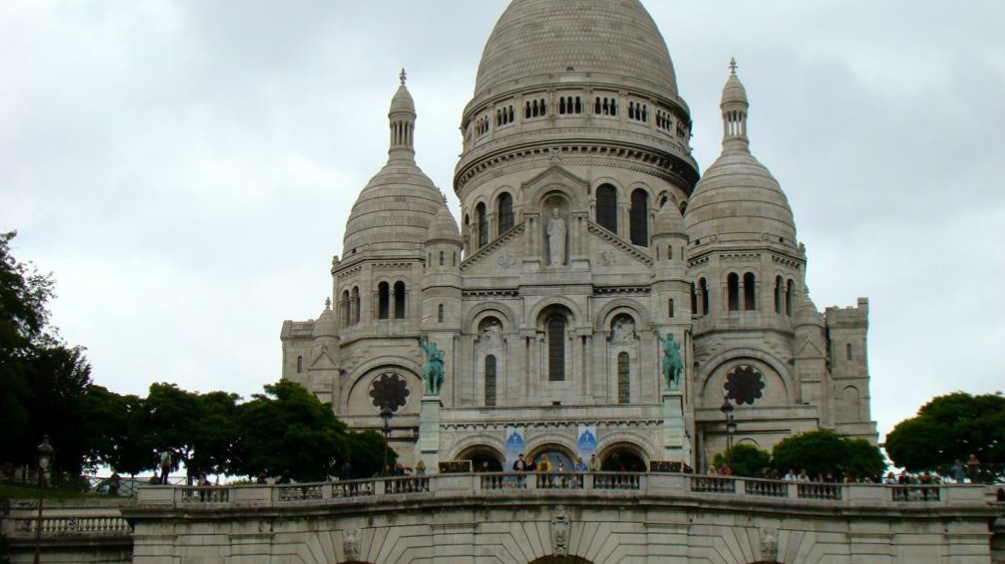} &
        \includegraphics[clip=true,trim={35 0 55 0},width=0.23\textwidth]{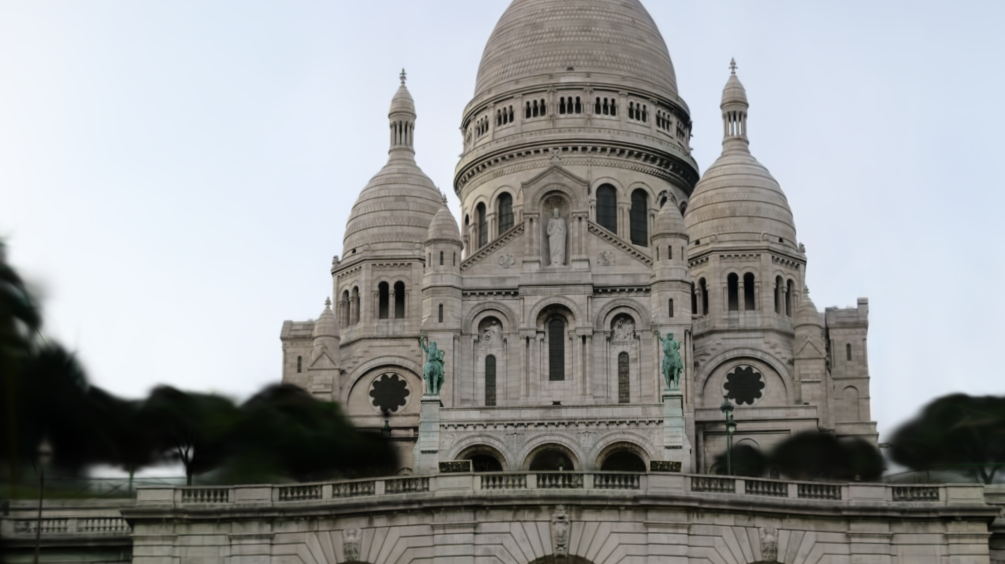} &
        \includegraphics[clip=true,trim={35 0 55 0},width=0.23\textwidth]{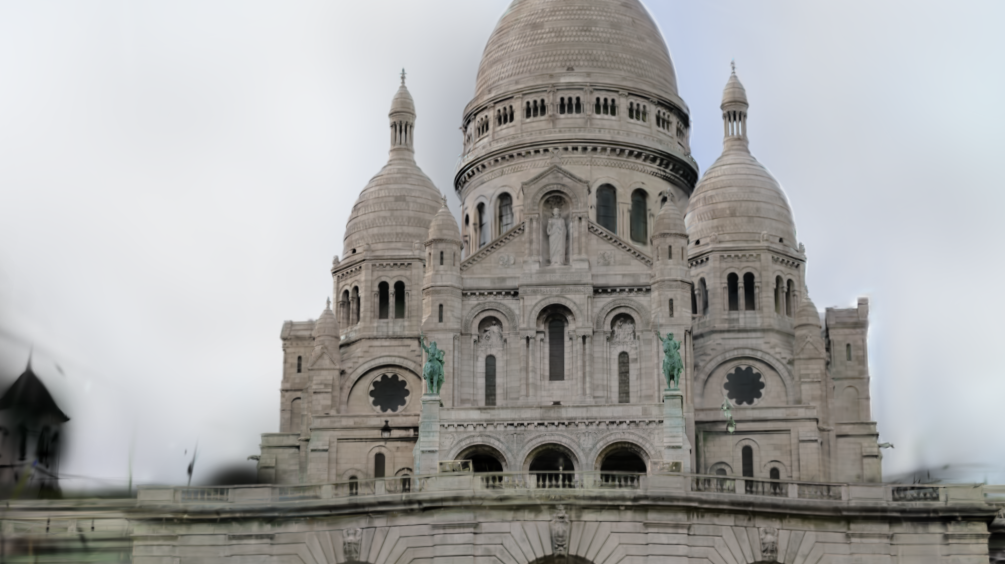} &
        \includegraphics[clip=true,trim={35 0 55 0},width=0.23\textwidth]{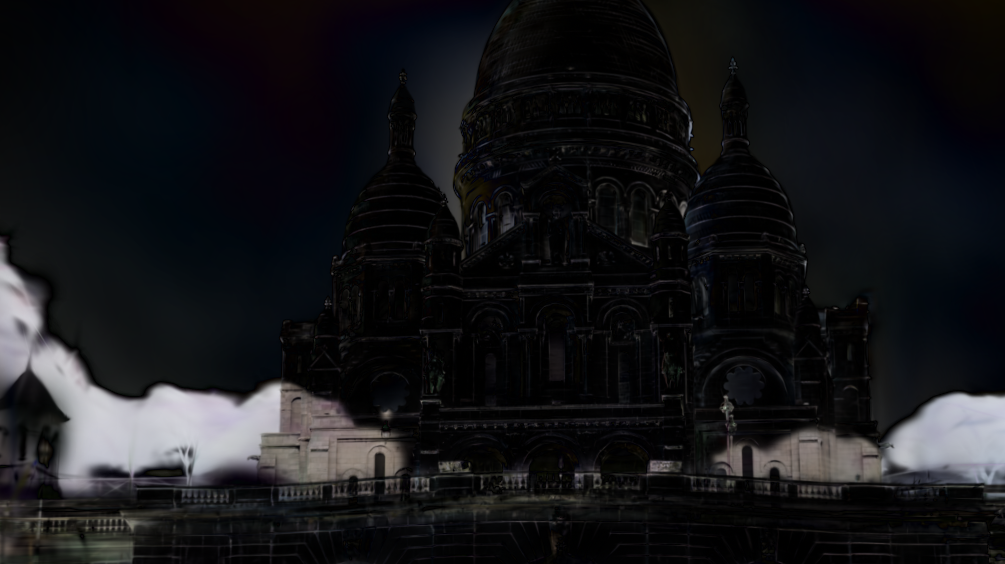}
        \\

        & \includegraphics[clip=true,trim={0 0 0 130},width=0.23\textwidth]{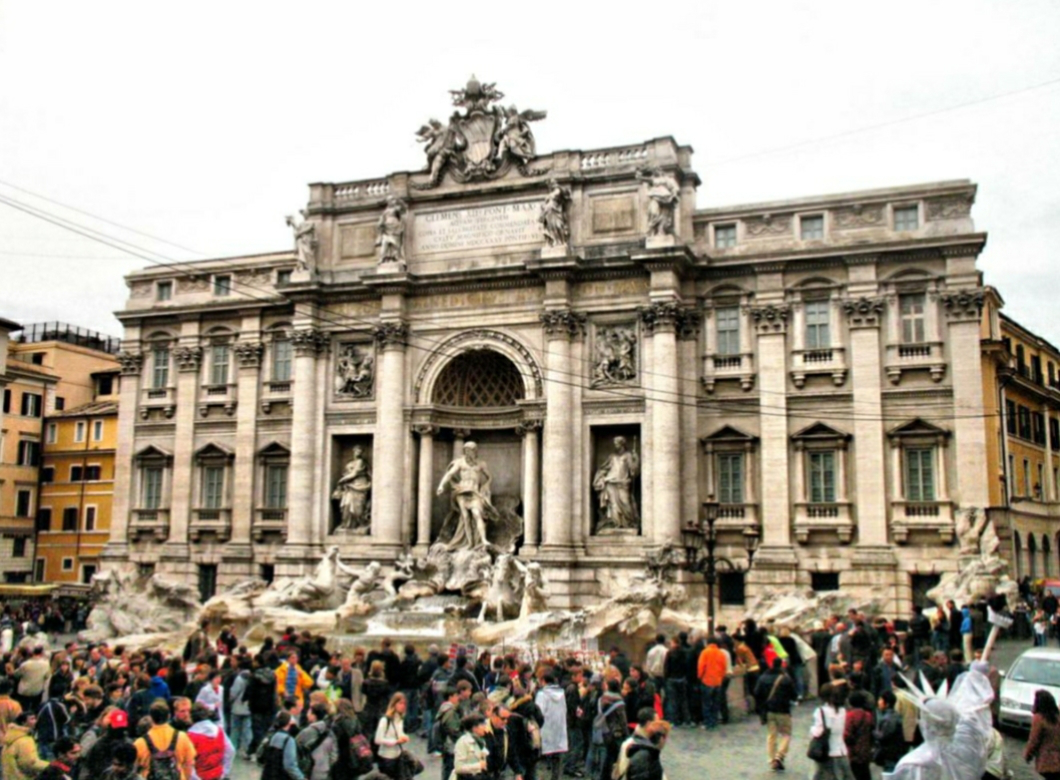} &
        \includegraphics[clip=true,trim={0 0 0 130},width=0.23\textwidth]{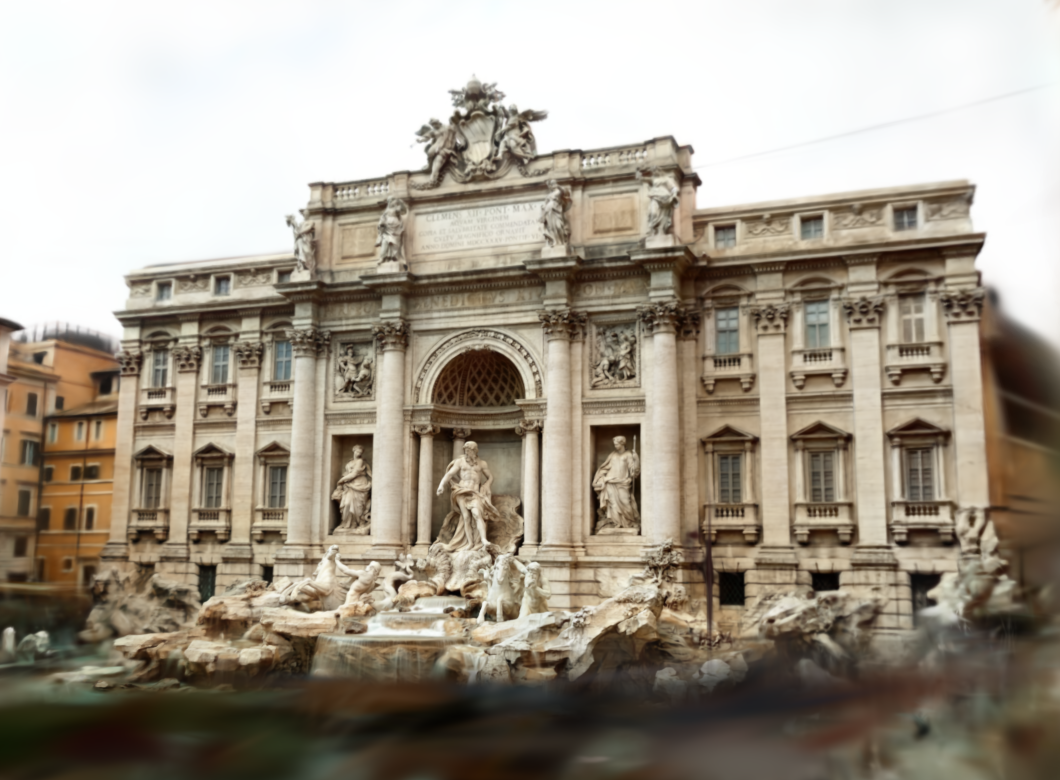} &
        \includegraphics[clip=true,trim={0 0 0 130},width=0.23\textwidth]{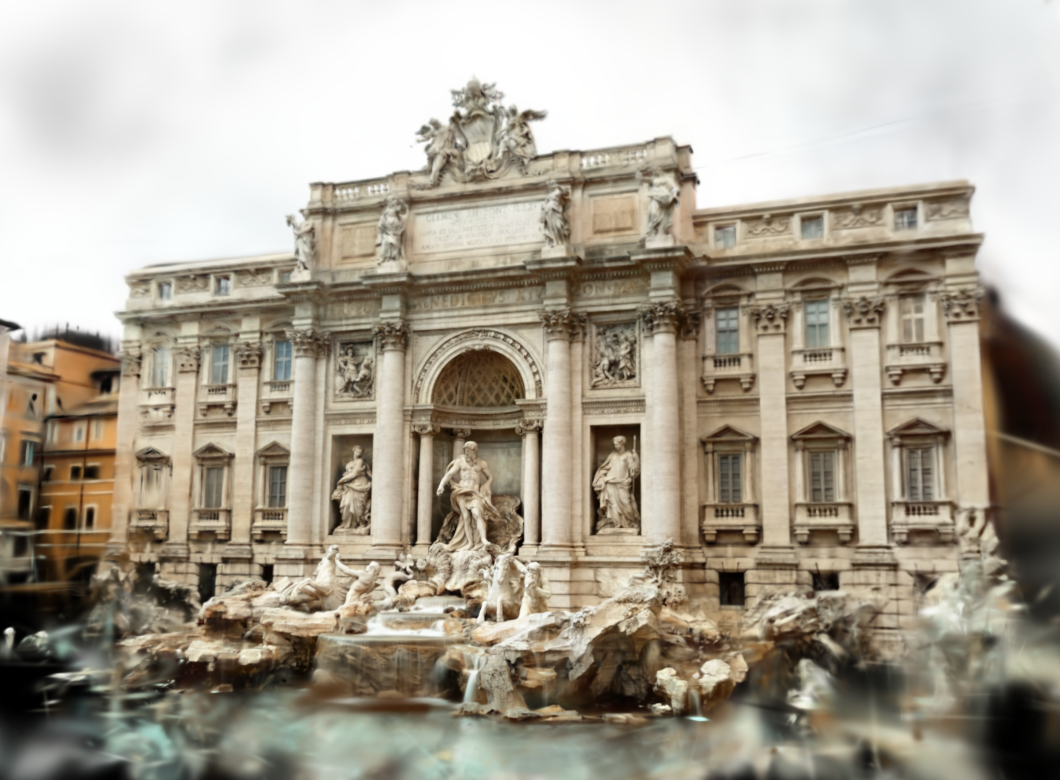} & 
        \includegraphics[clip=true,trim={0 0 0 130},width=0.23\textwidth]{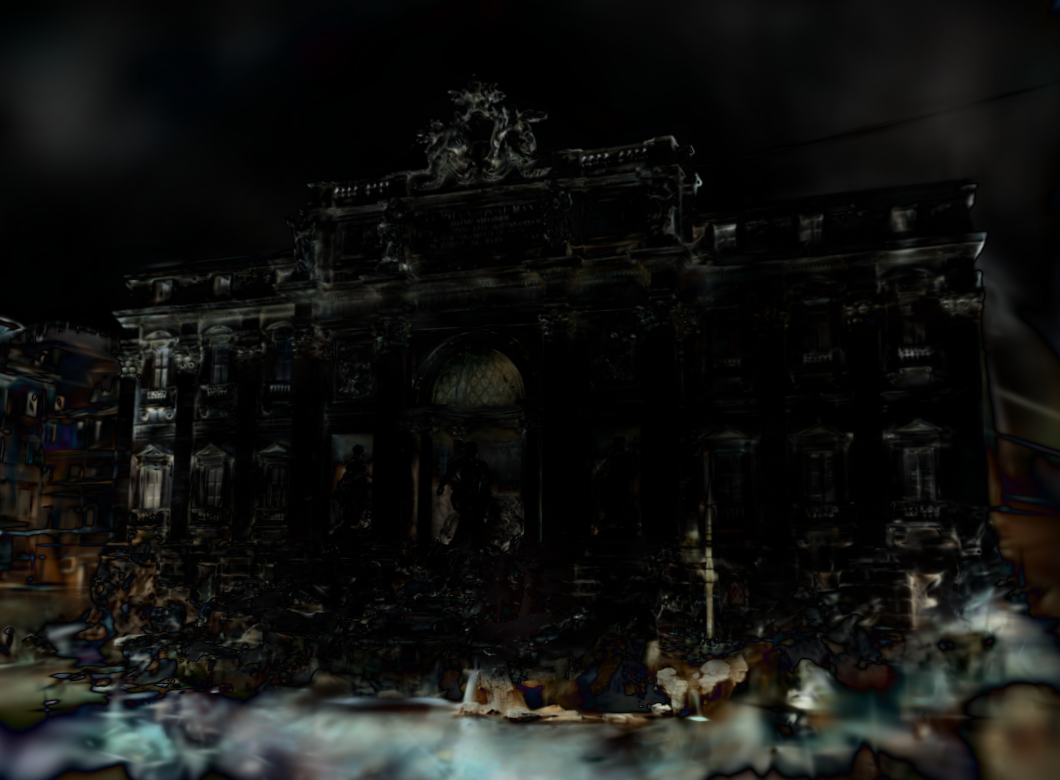}
        \\

        \cmidrule(rl){2-5}

        \multirow{4}{*}[0mm]{\rotatebox[origin=c]{90}{NeRF-OSR Dataset~\cite{rudnev2022nerfosr}}}
        & \includegraphics[clip=true,trim={0 0 0 29},width=0.23\textwidth]{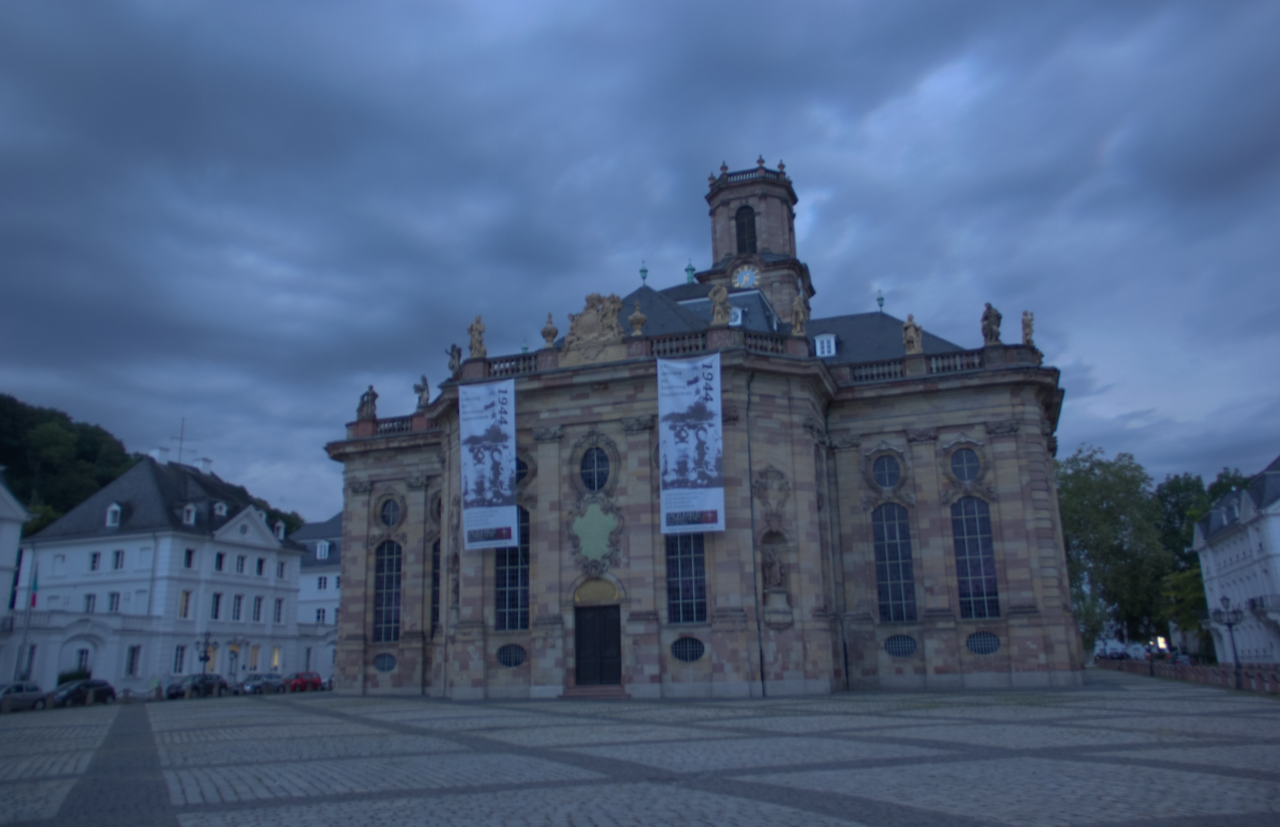} &
        \includegraphics[clip=true,trim={0 0 0 29},width=0.23\textwidth]{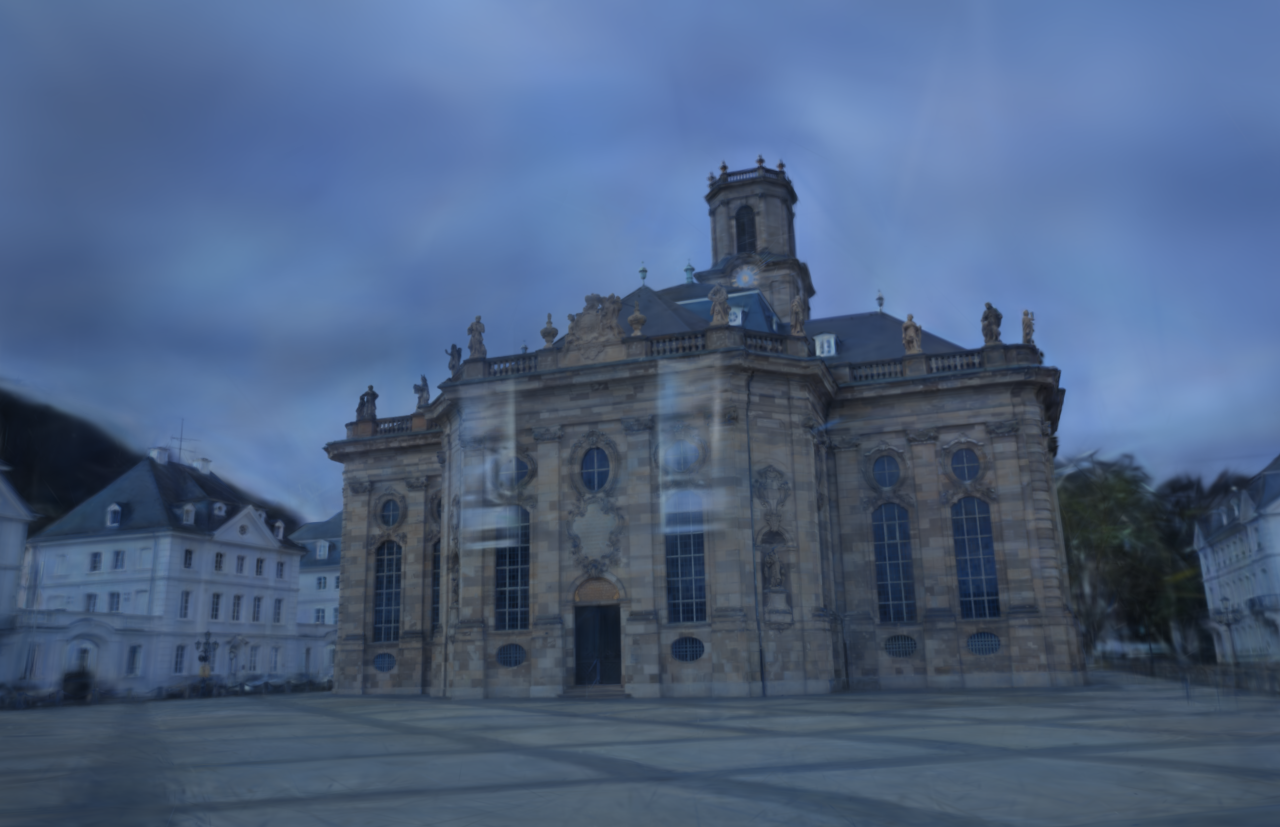} &
        \includegraphics[clip=true,trim={0 0 0 29},width=0.23\textwidth]{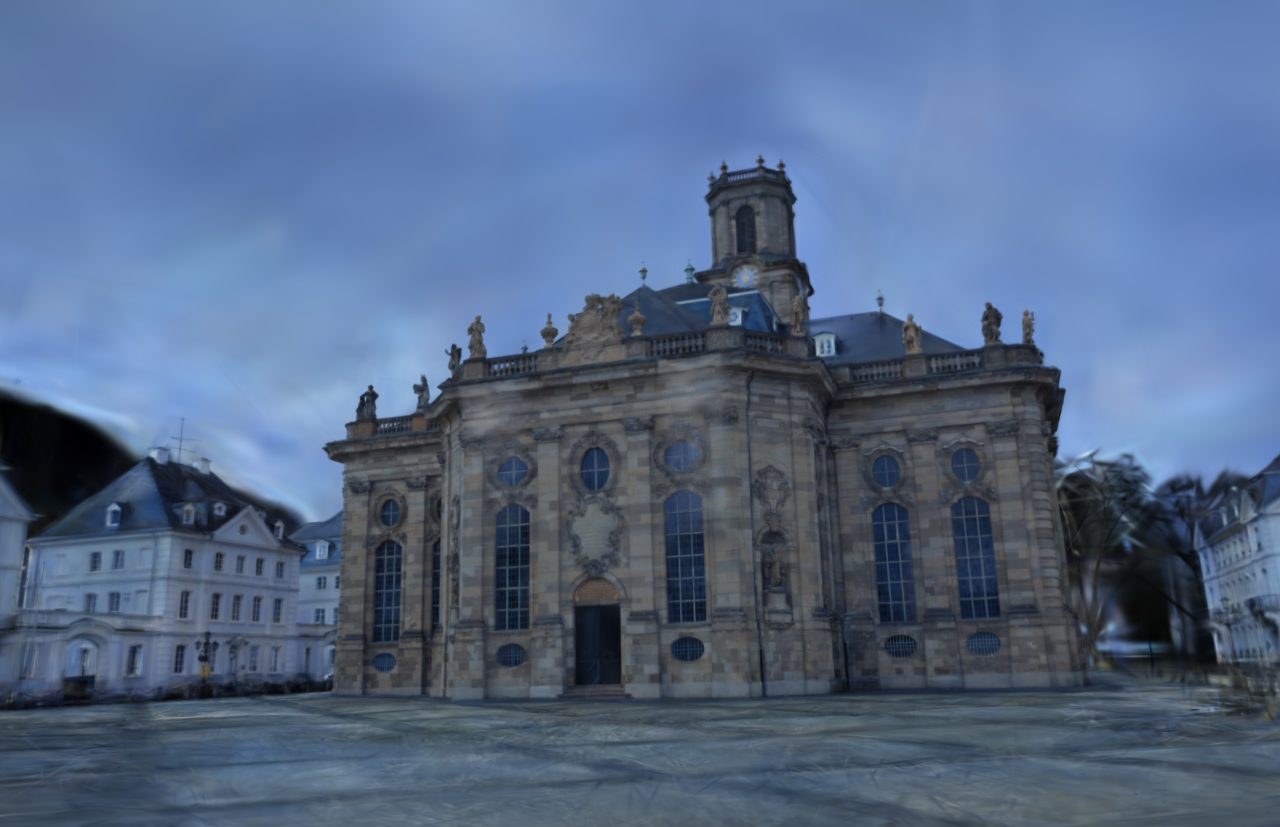} &
        \includegraphics[clip=true,trim={0 0 0 29},width=0.23\textwidth]{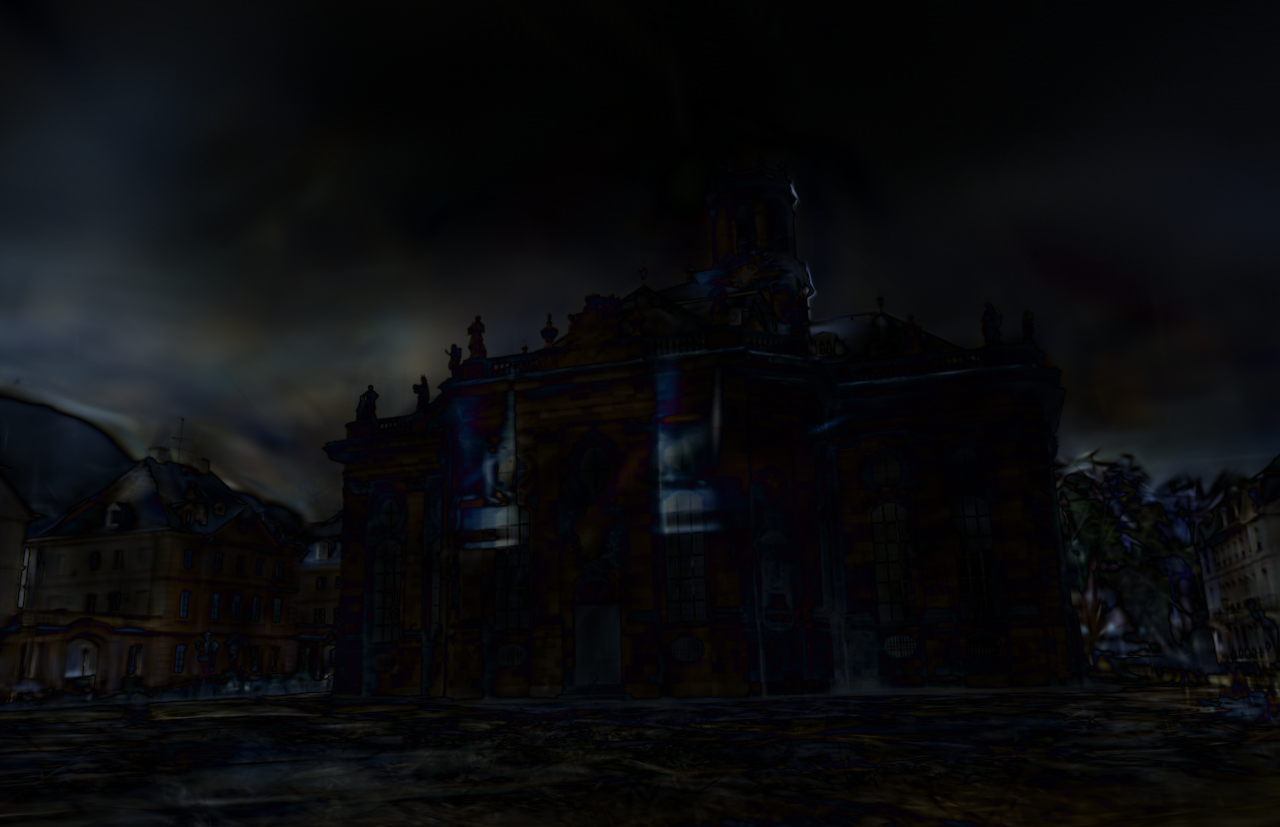}
        \\
    
        & \includegraphics[clip=true,trim={0 0 0 44},width=0.23\textwidth]{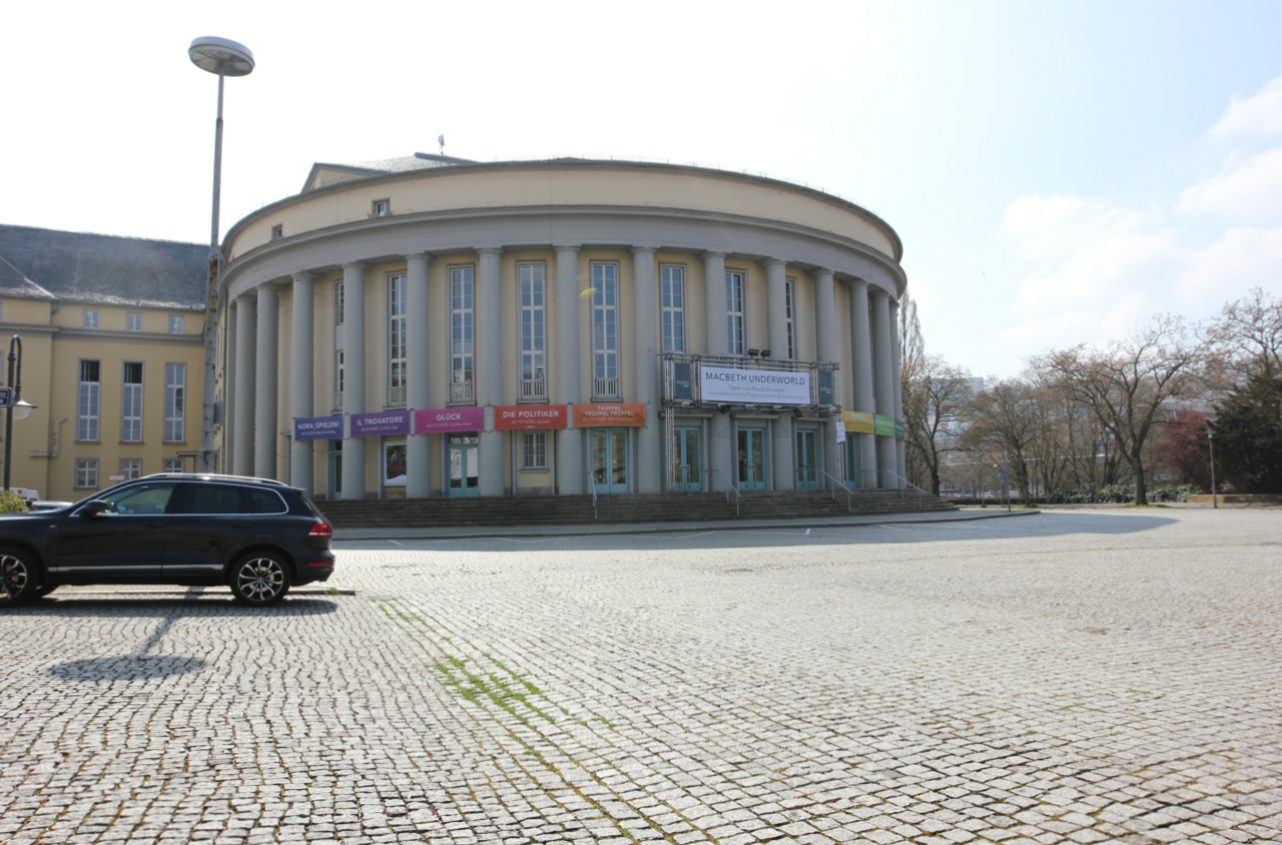} &
        \includegraphics[clip=true,trim={0 0 0 44},width=0.23\textwidth]{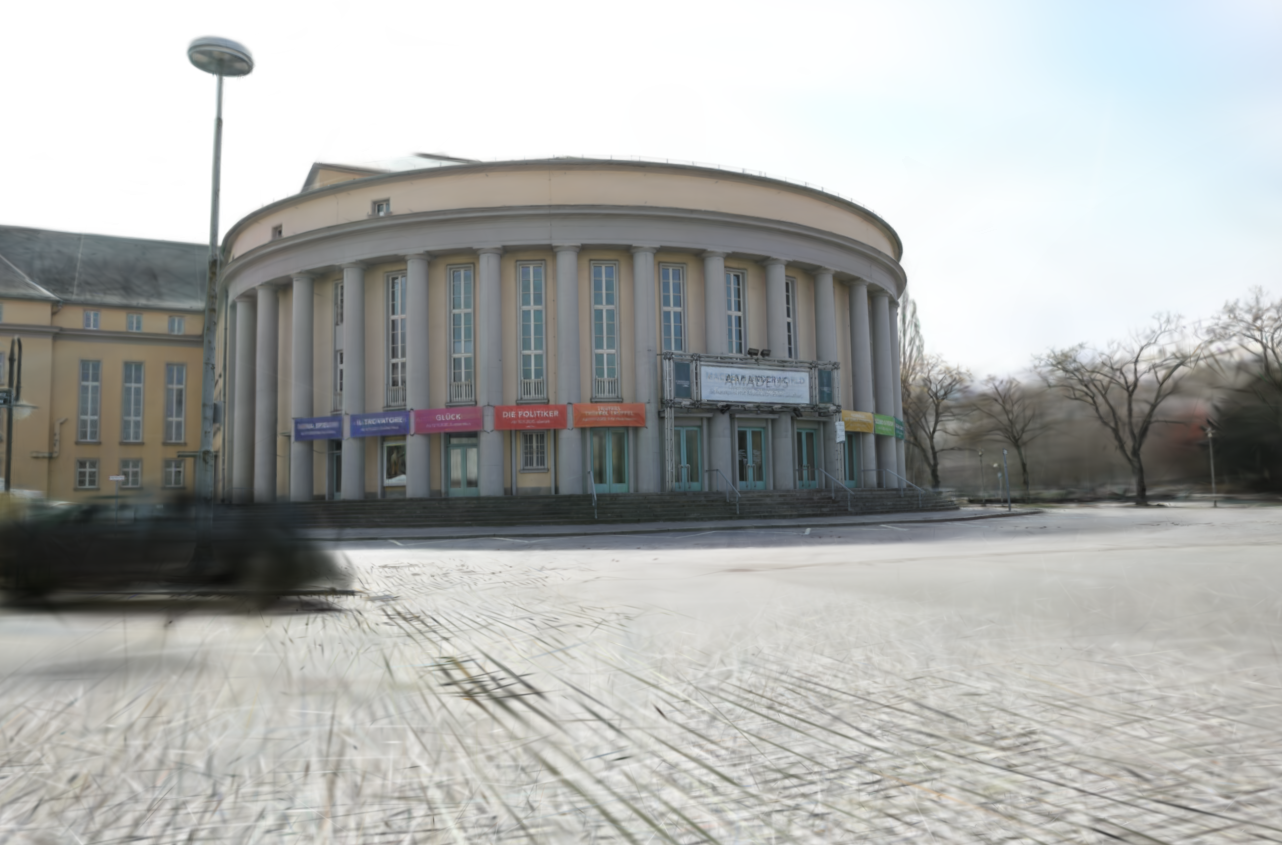} &
        \includegraphics[clip=true,trim={0 0 0 44},width=0.23\textwidth]{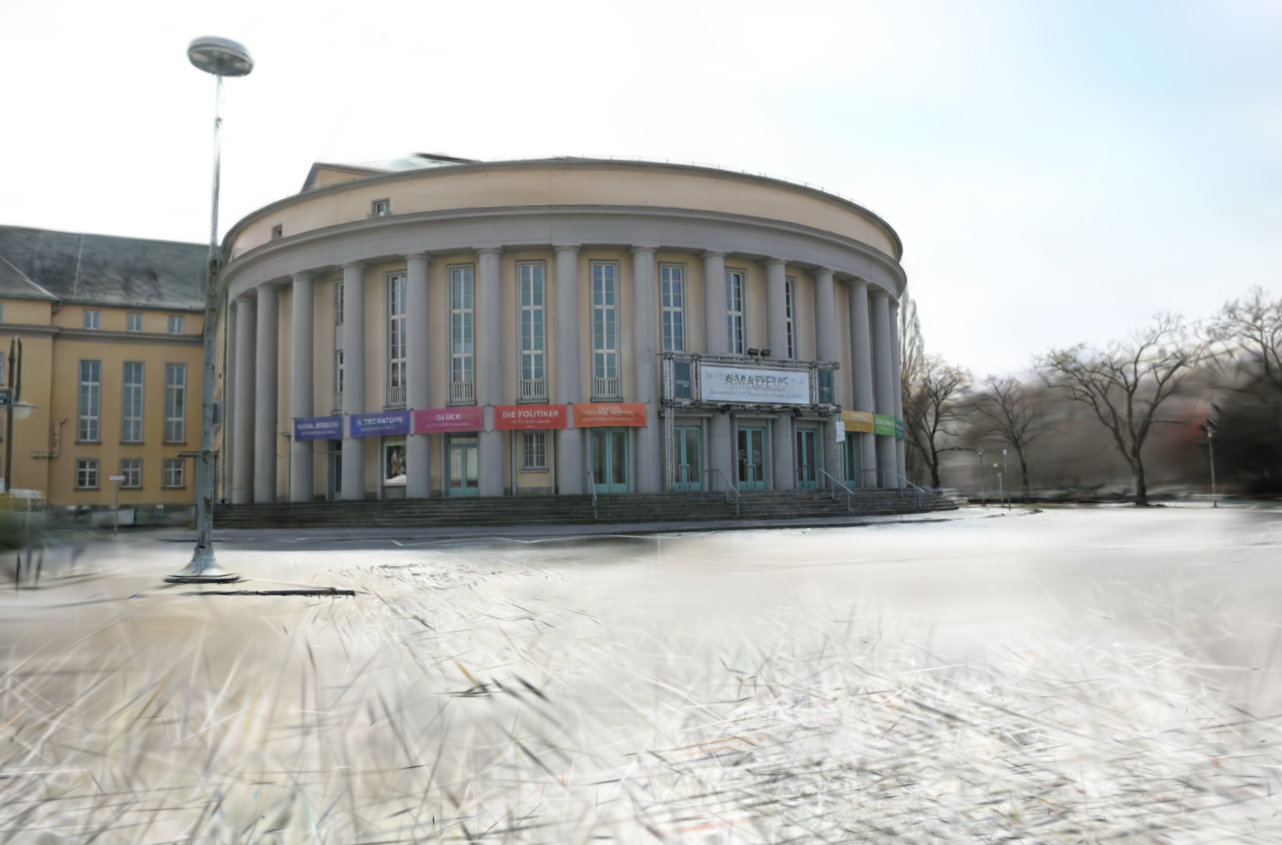} & 
        \includegraphics[clip=true,trim={0 0 0 44},width=0.23\textwidth]{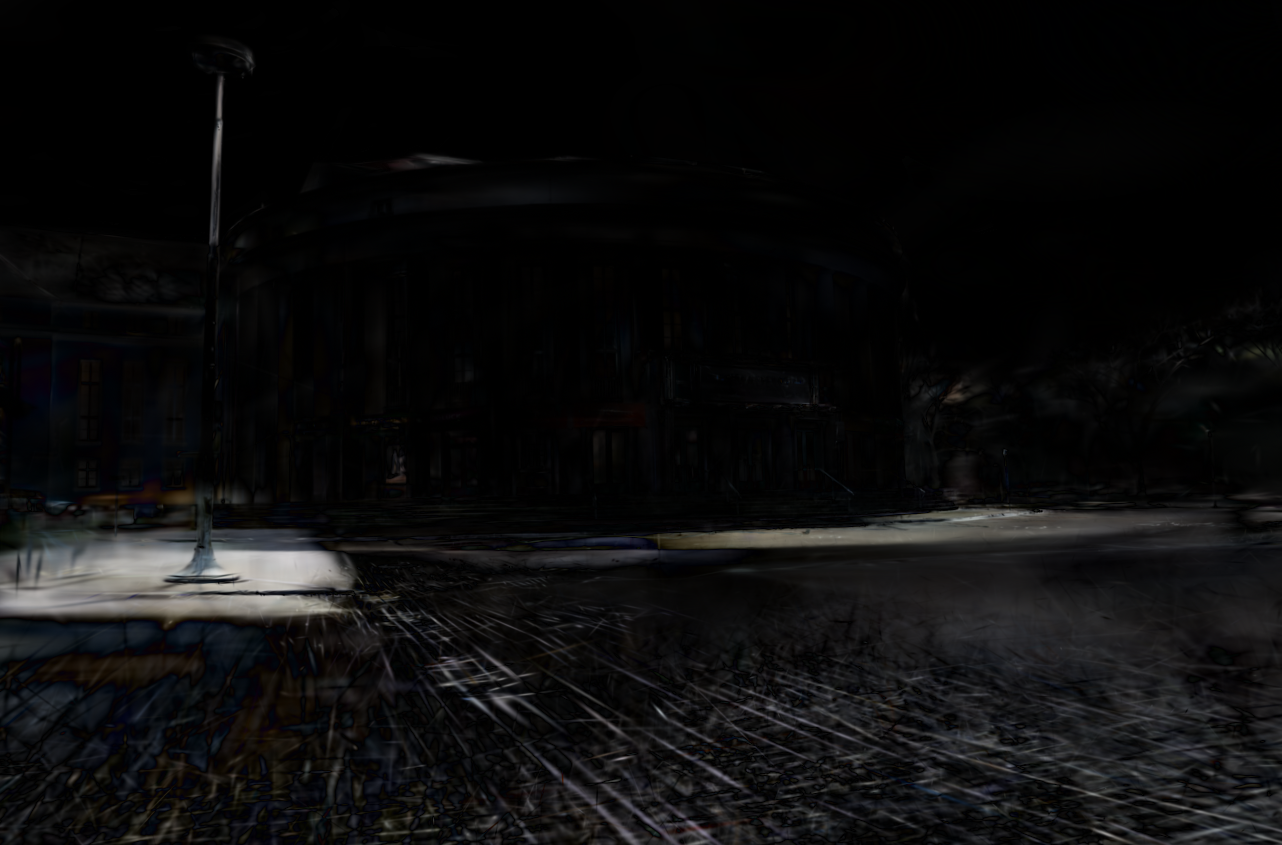}
        \\

        & \includegraphics[clip=true,trim={0 0 0 55},width=0.23\textwidth]{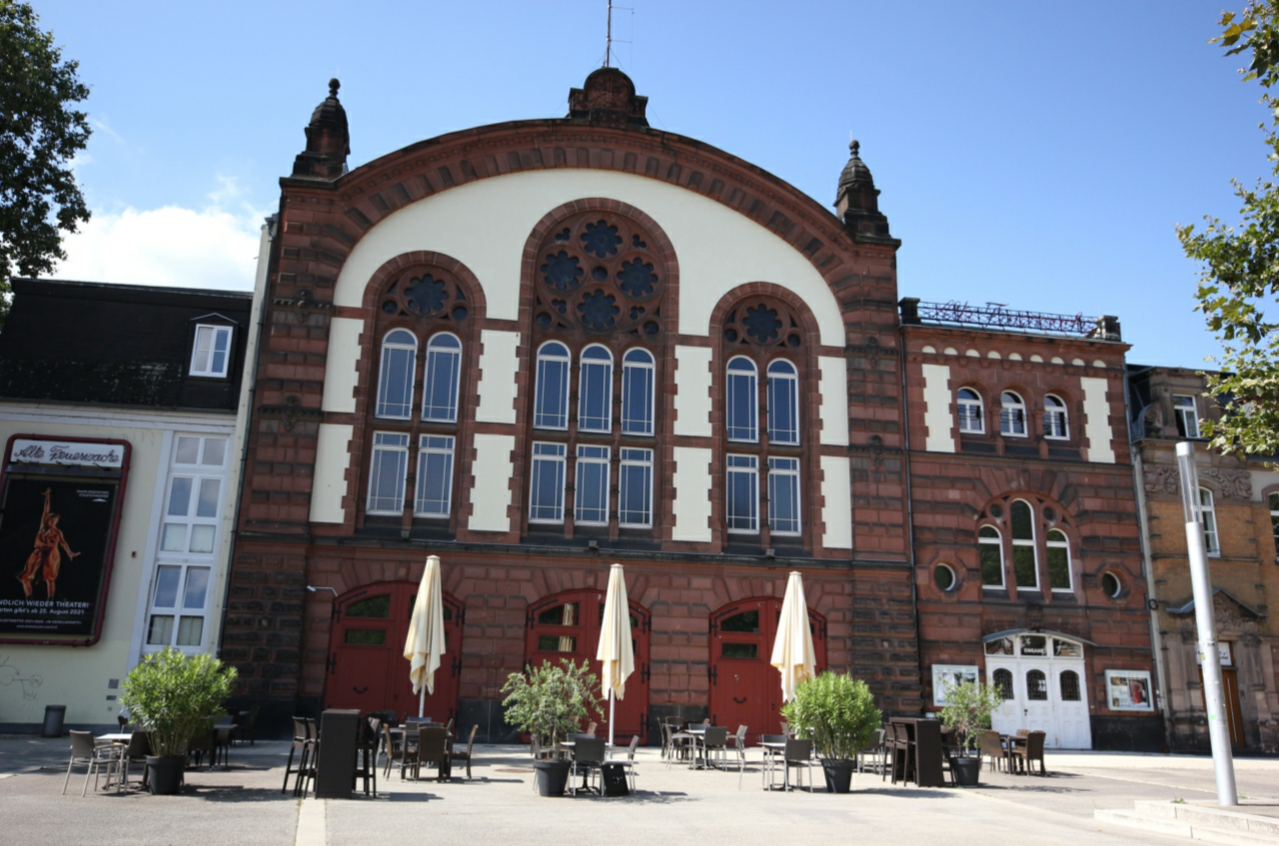} & 
        \includegraphics[clip=true,trim={0 0 0 55},width=0.23\textwidth]{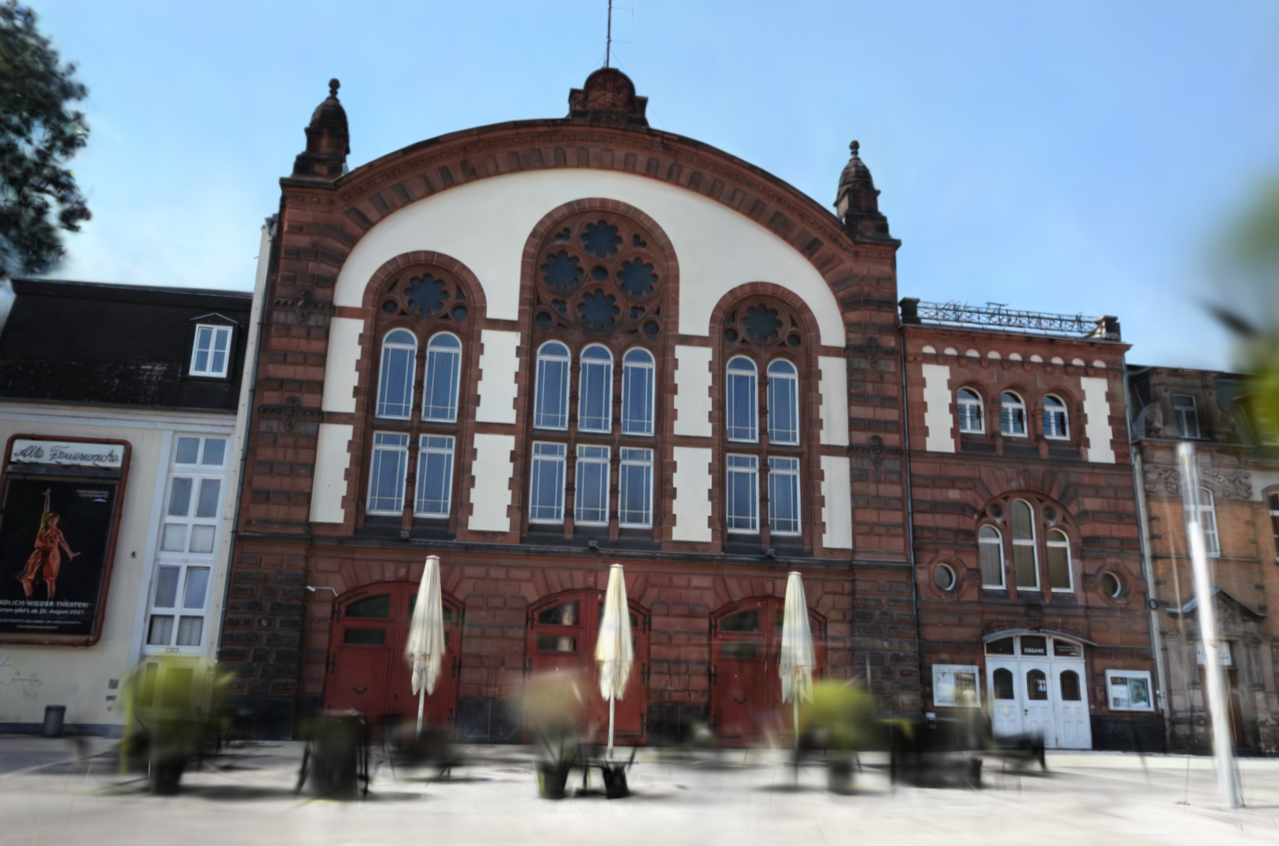} & 
        \includegraphics[clip=true,trim={0 0 0 55},width=0.23\textwidth]{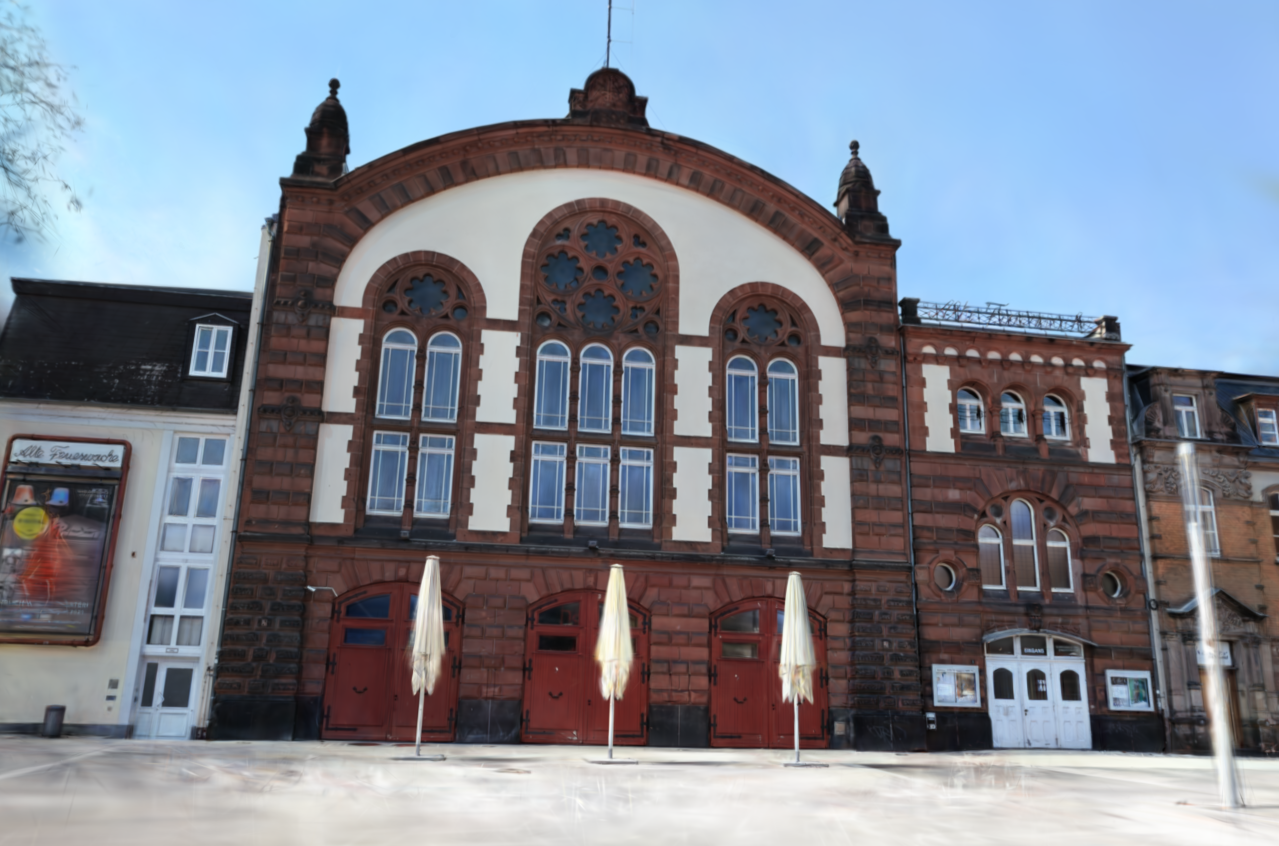} & 
        \includegraphics[clip=true,trim={0 0 0 55},width=0.23\textwidth]{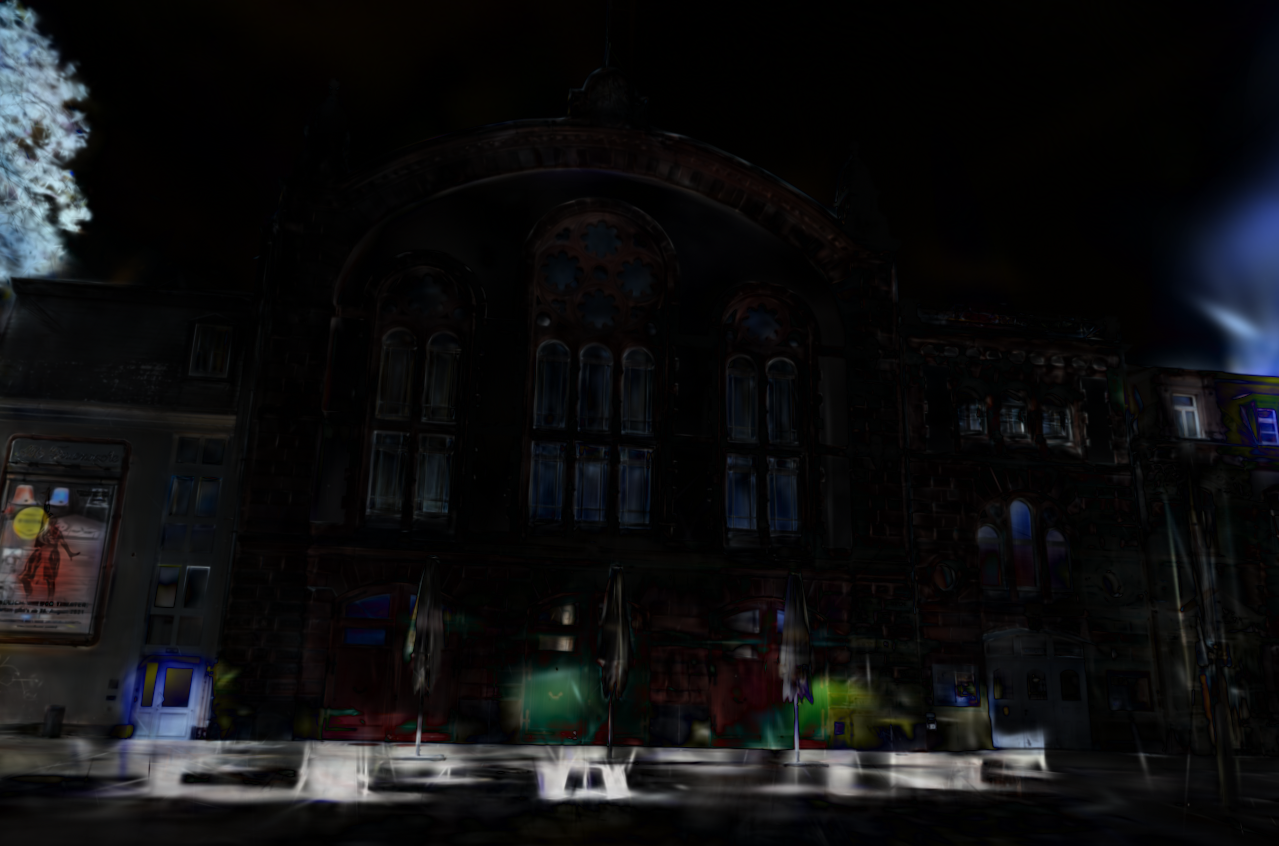} 
        \\

        & \includegraphics[clip=true,trim={0 0 0 52},width=0.23\textwidth]{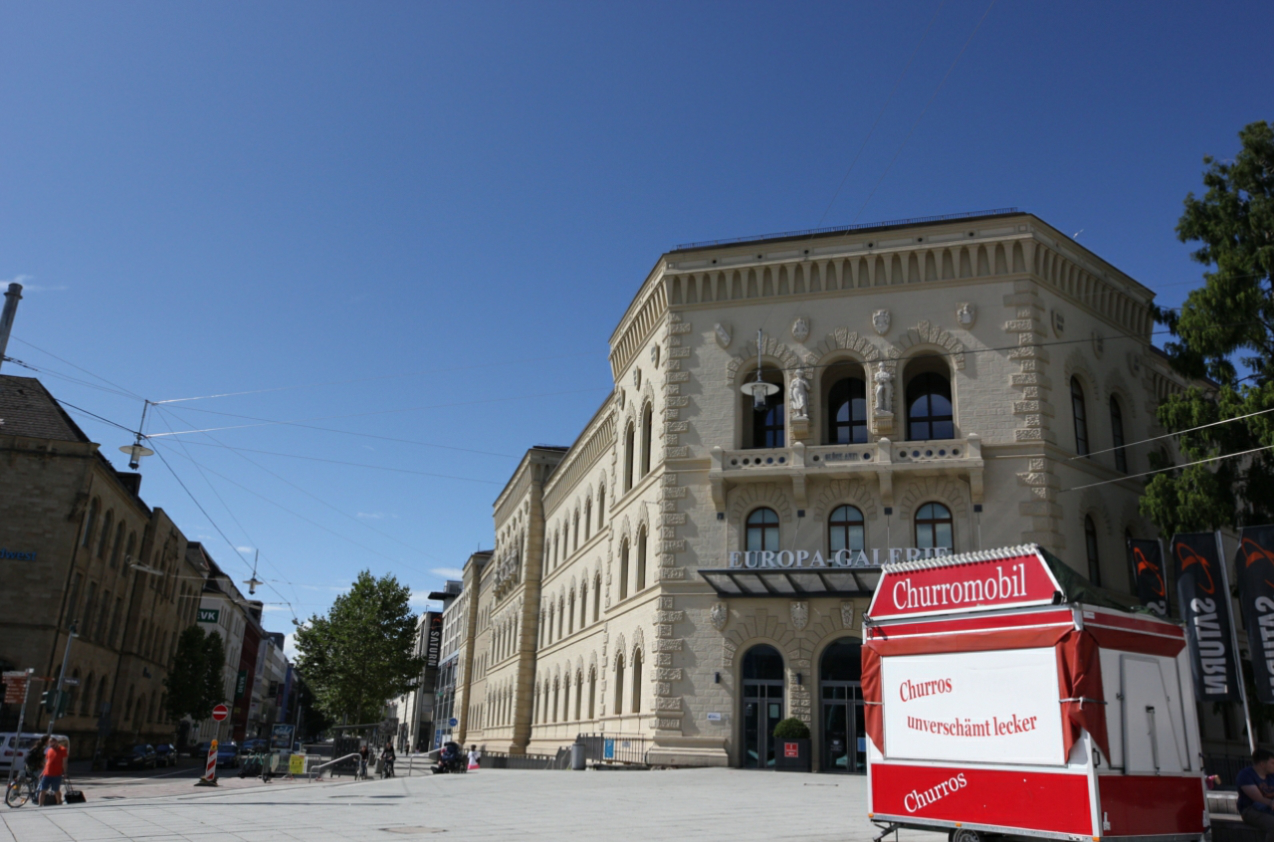}  &
        \includegraphics[clip=true,trim={0 0 0 52},width=0.23\textwidth]{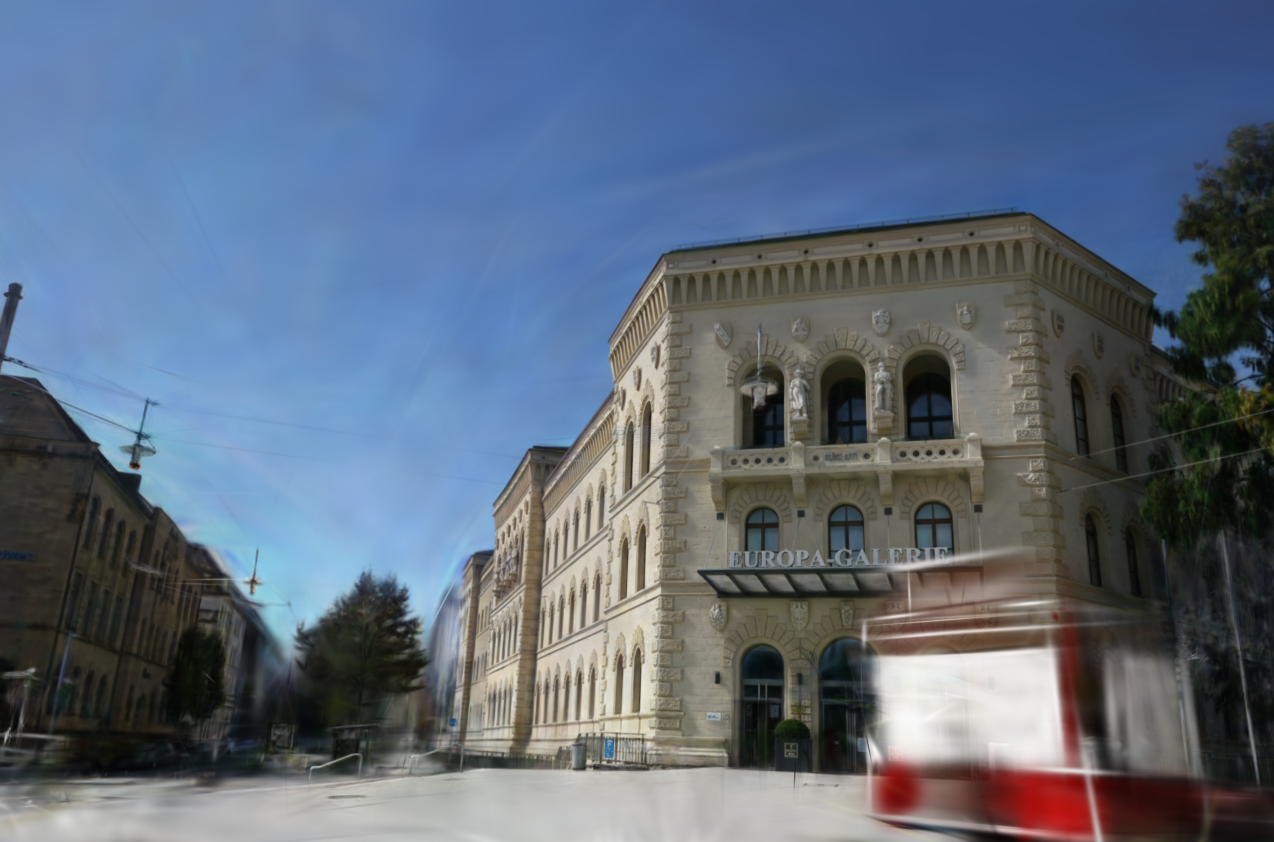}  &
        \includegraphics[clip=true,trim={0 0 0 52},width=0.23\textwidth]{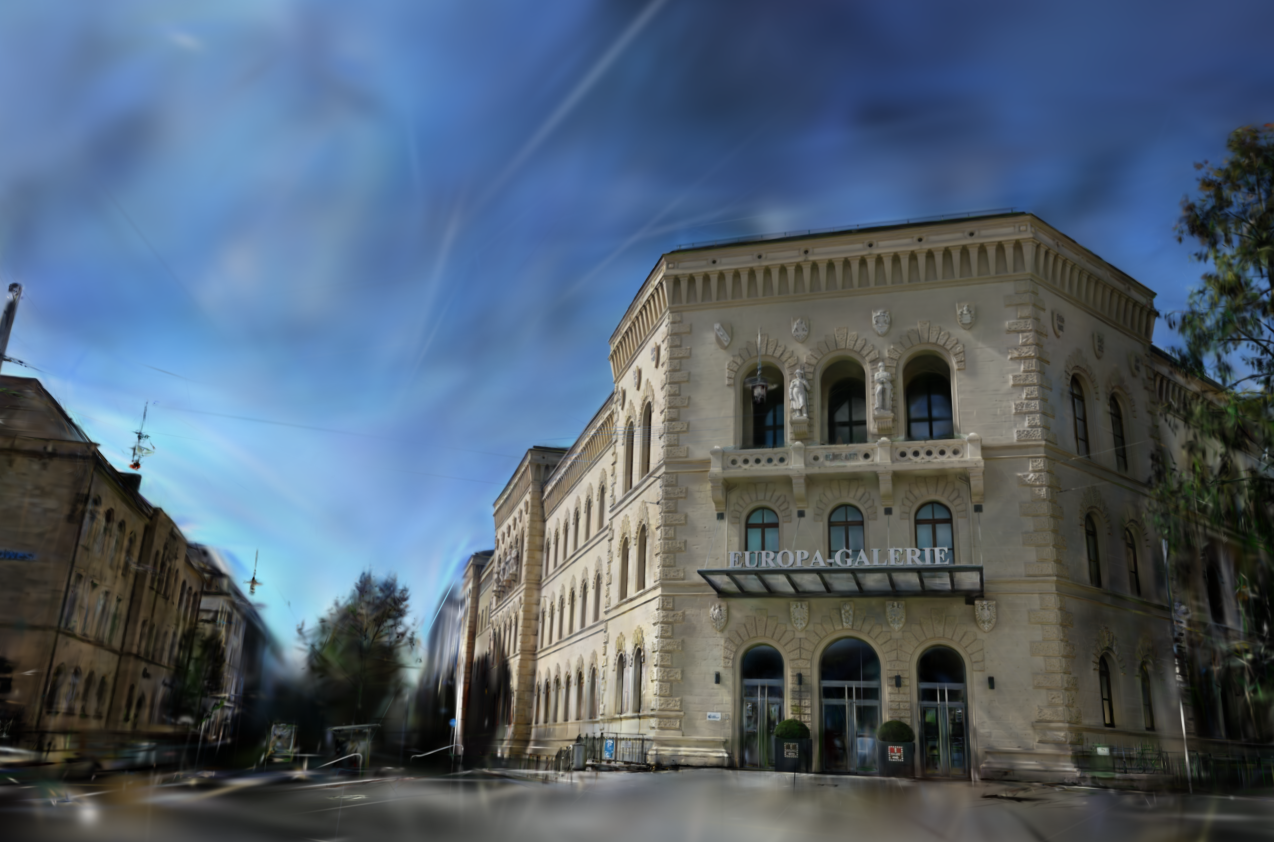}  &
        \includegraphics[clip=true,trim={0 0 0 52},width=0.23\textwidth]{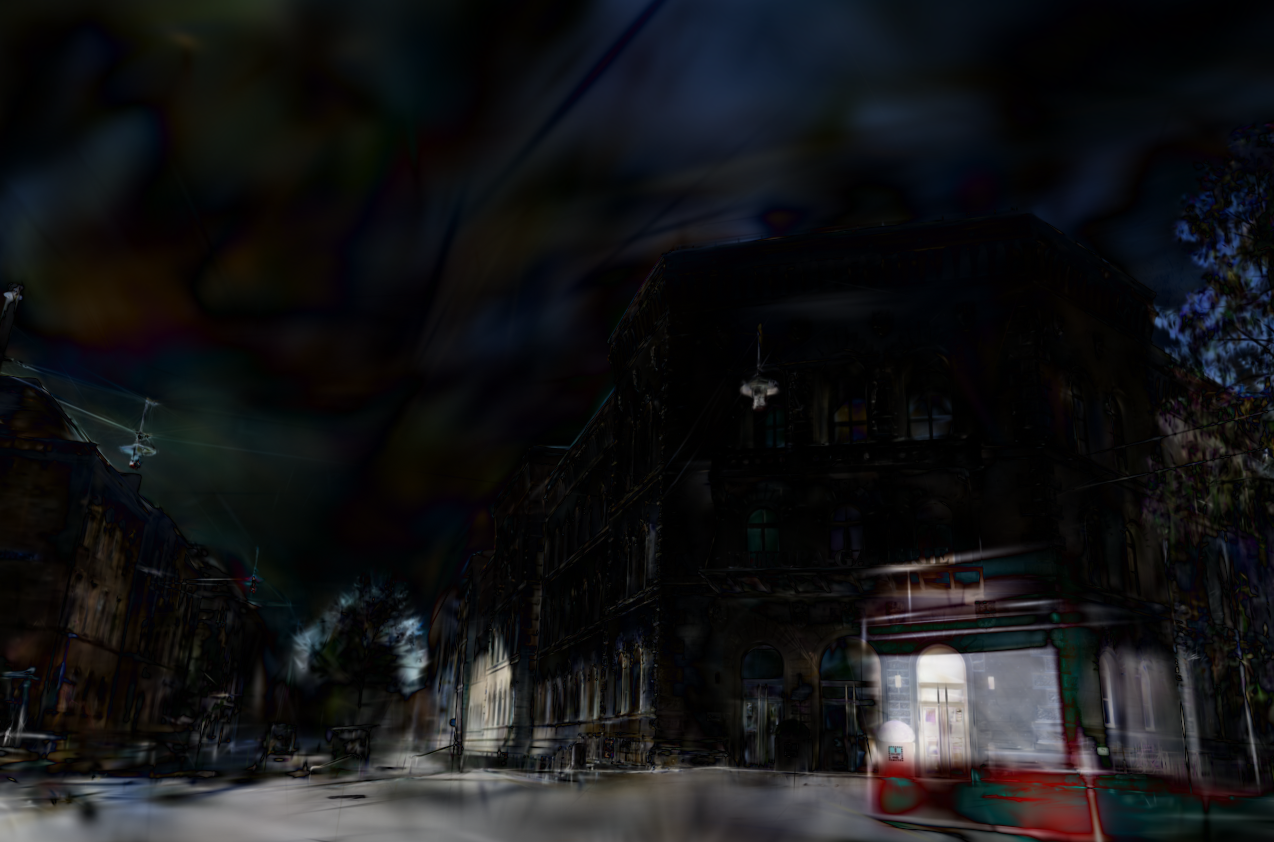}
        \\

        & Ground-truth & SWAG & SWAG Static & SWAG Transient  \\

	\end{tabular}
	\caption{Qualitative experimental results on three real-world scenes from Phototourism~\cite{Jin_2020} and four NeRF-OSR~\cite{rudnev2022nerfosr} scenes. We demonstrate the capability of SWAG to disentangle between static and transient parts of the scene.
    }
	\label{fig:decom}
\end{figure*}

\Cref{fig:fig9} and \Cref{fig:decom} show qualitative results of 3DGS and \method~on Phototourism scenes and four of NeRF-OSR sites. In some parts of the scene, 3DGS presents artifacts and floaters either in under-observed part of the scene or in areas occluded by transient objects. Adding to that, the rendered color is shifted from the ground truth color as shown in \Cref{fig:fig9}. \method~produces visual appearances close to the ground truth thanks to our per-image conditioning. Compared to previous implicit methods~\cite{martinbrualla2020nerfw,chen2022hallucinated,yang2023crossray,kassab2024refinedfields}, our proposal achieves one order of magnitude faster training and guarantees real-time rendering as shown in~\Cref{tab:quantitative1}.

\subsection{Controllable appearance}

\begin{figure}[tb] 
\centering

    \includegraphics[width=.95\textwidth]{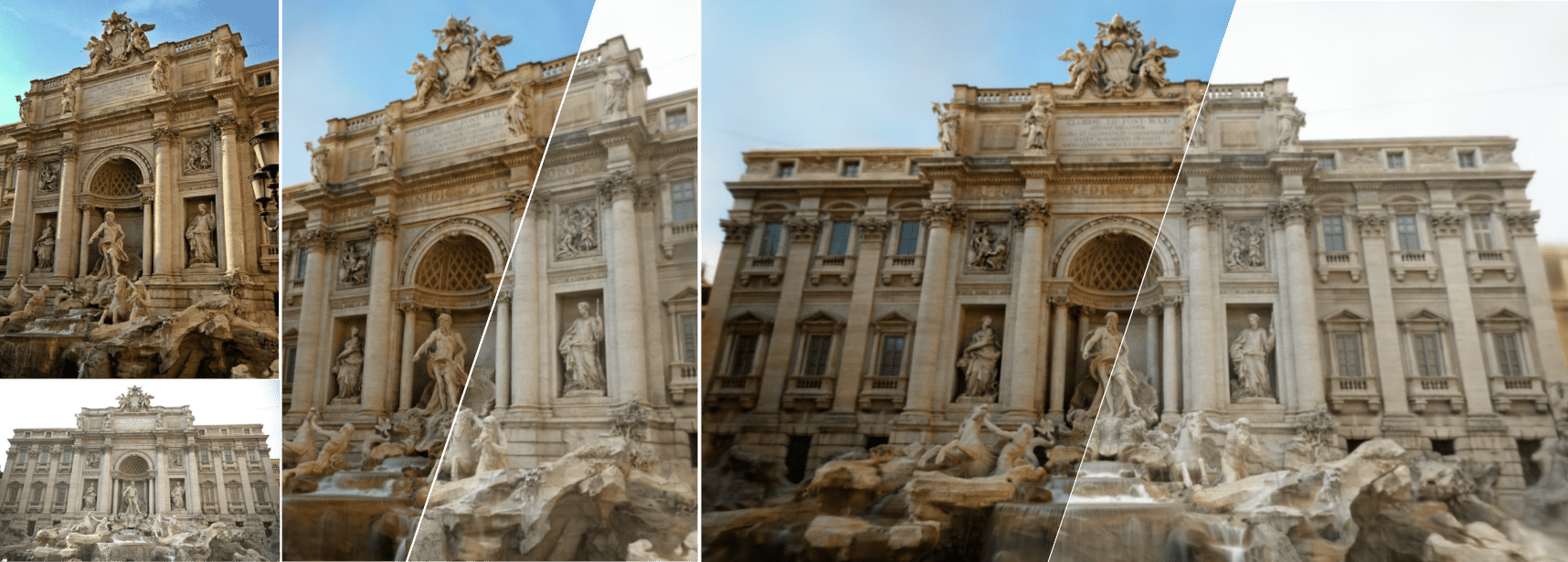}
   
    \caption{\textbf{Appearance transfer --} SWAG is able to render a scene at any viewpoint with the visual appearance of any training image thanks to the learned appearance embeddings. %
    } 
    \label{fig:learnt}
\end{figure}

\begin{figure*}[tb] 
	\centering
	\scriptsize
	\setlength{\tabcolsep}{0.002\linewidth}
	\renewcommand{\arraystretch}{0.8}
	\begin{tabular}{cccccccc}

    \includegraphics[width=0.16\textwidth]{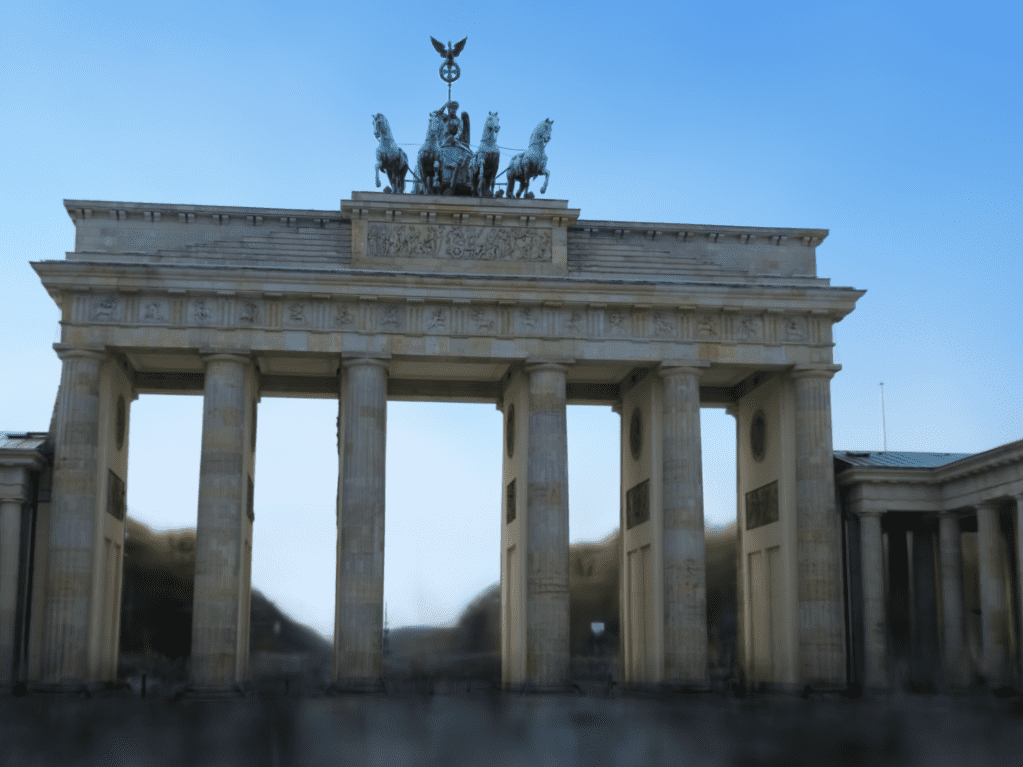} & ~ &
    \includegraphics[width=0.16\textwidth]{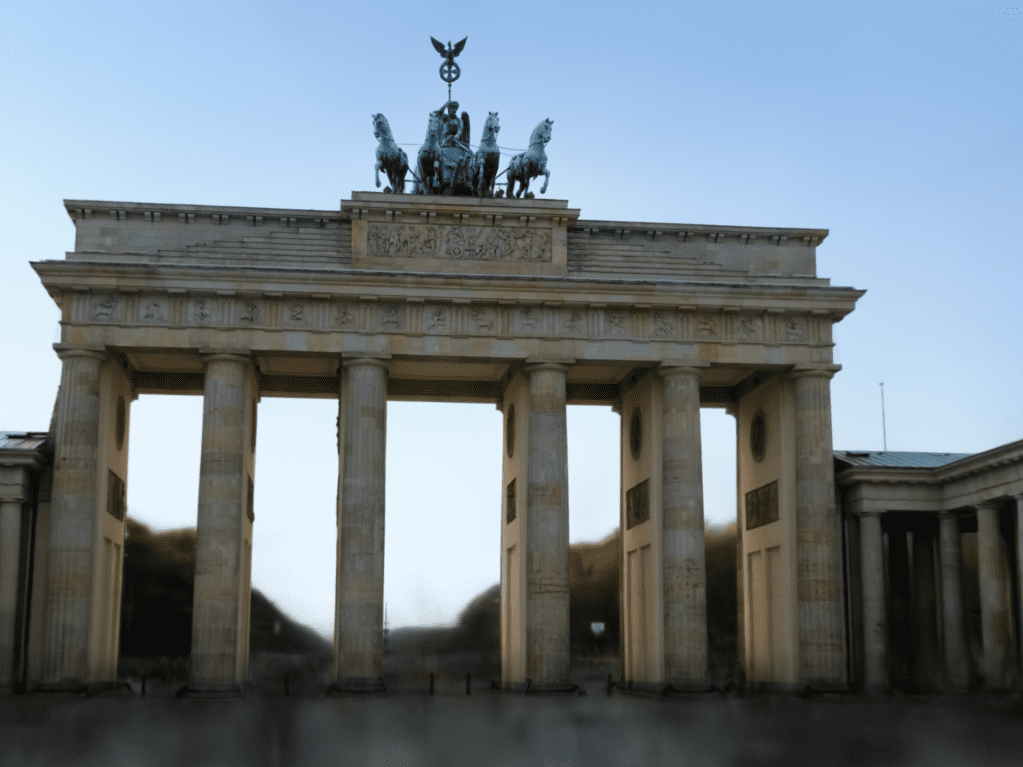}  &
    \includegraphics[width=0.16\textwidth]{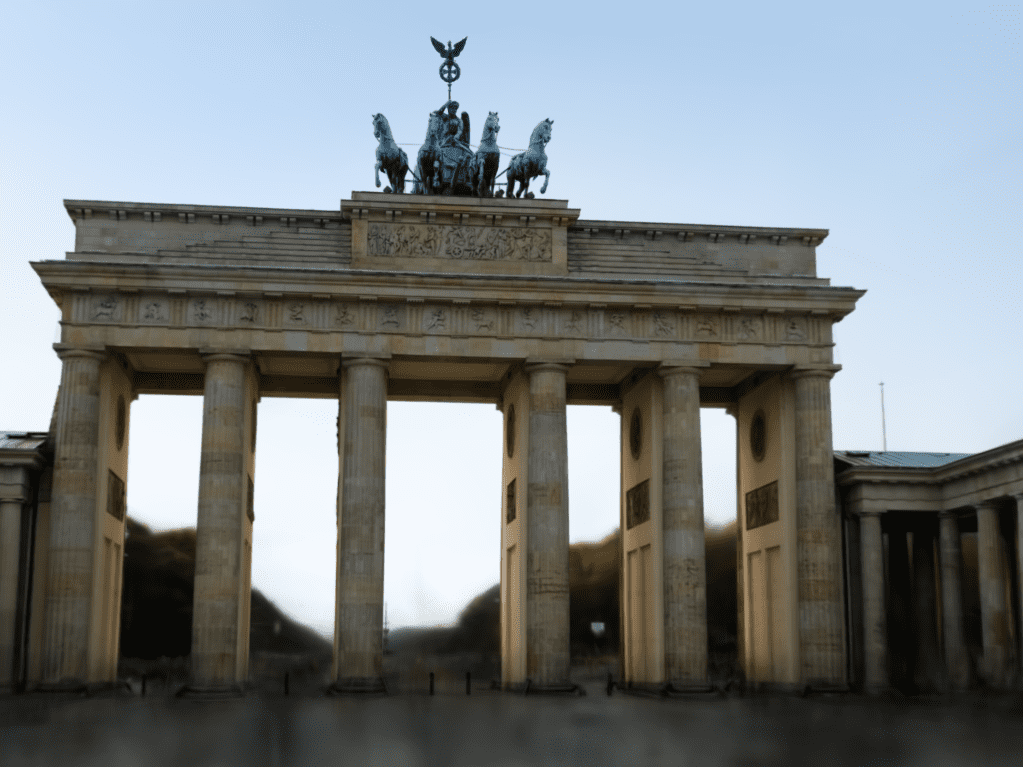} &
    \includegraphics[width=0.16\textwidth]{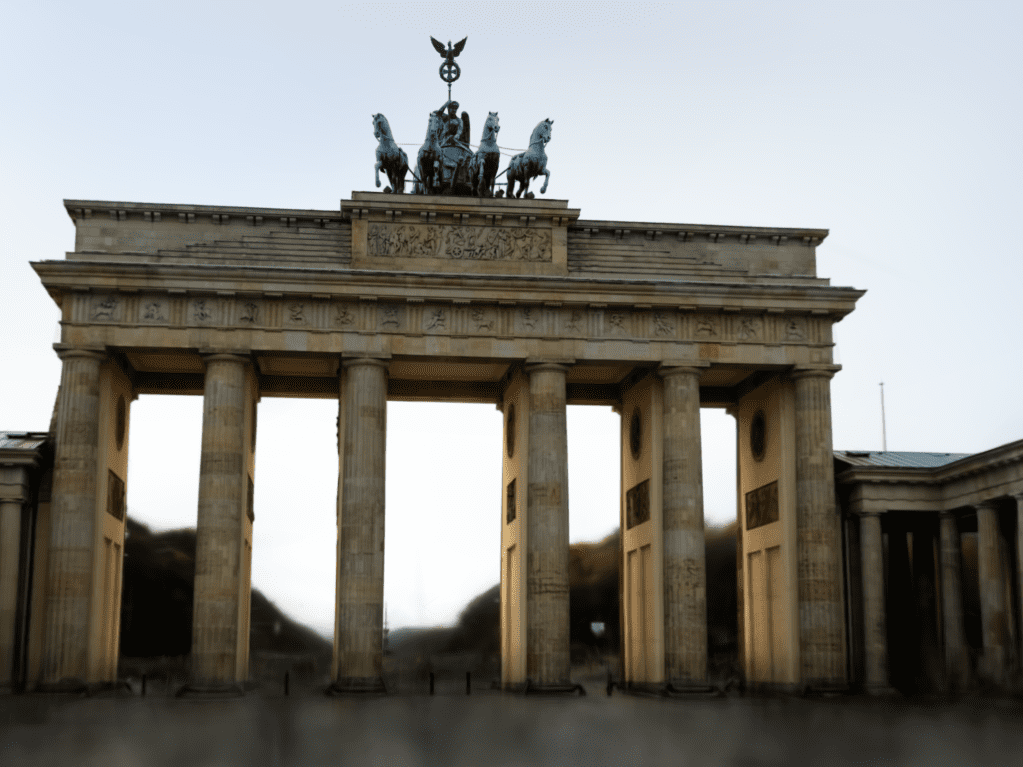} &
    \includegraphics[width=0.16\textwidth]{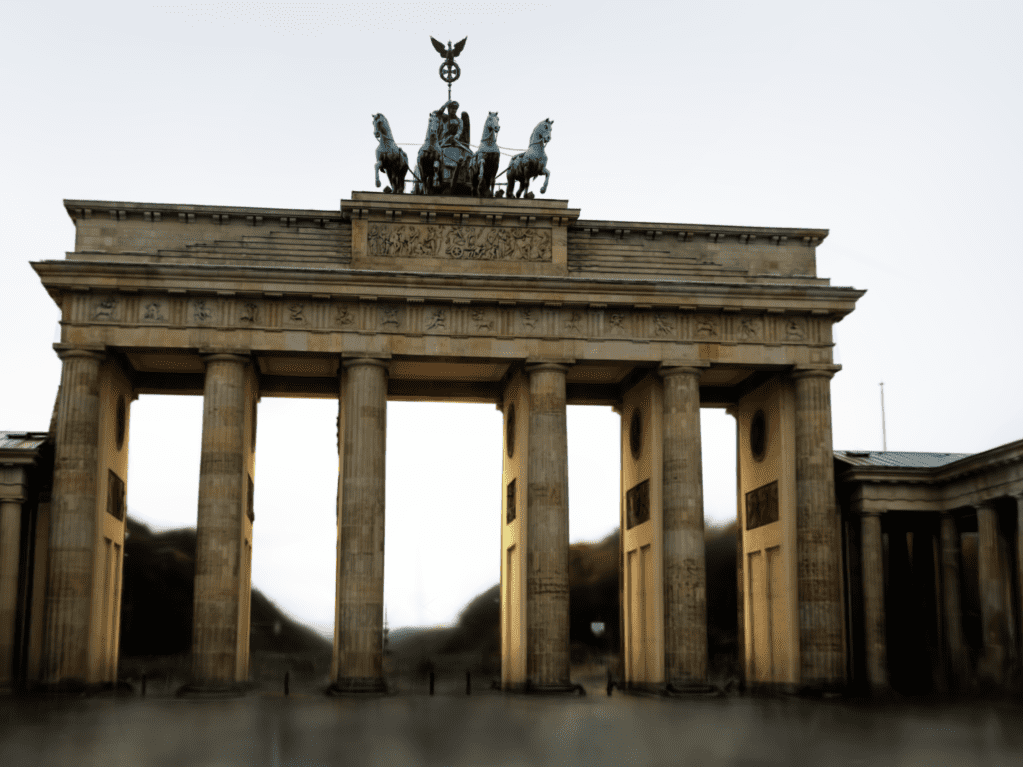} & ~ &
    \includegraphics[width=0.16\textwidth]{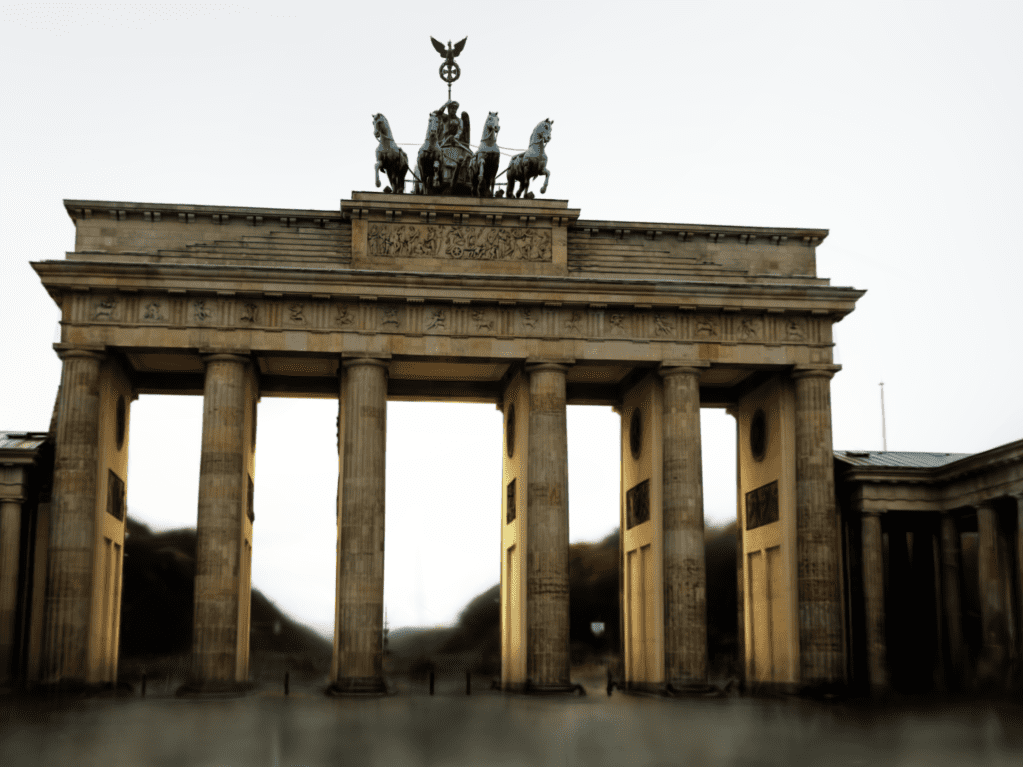} \\

    \includegraphics[width=0.16\textwidth]{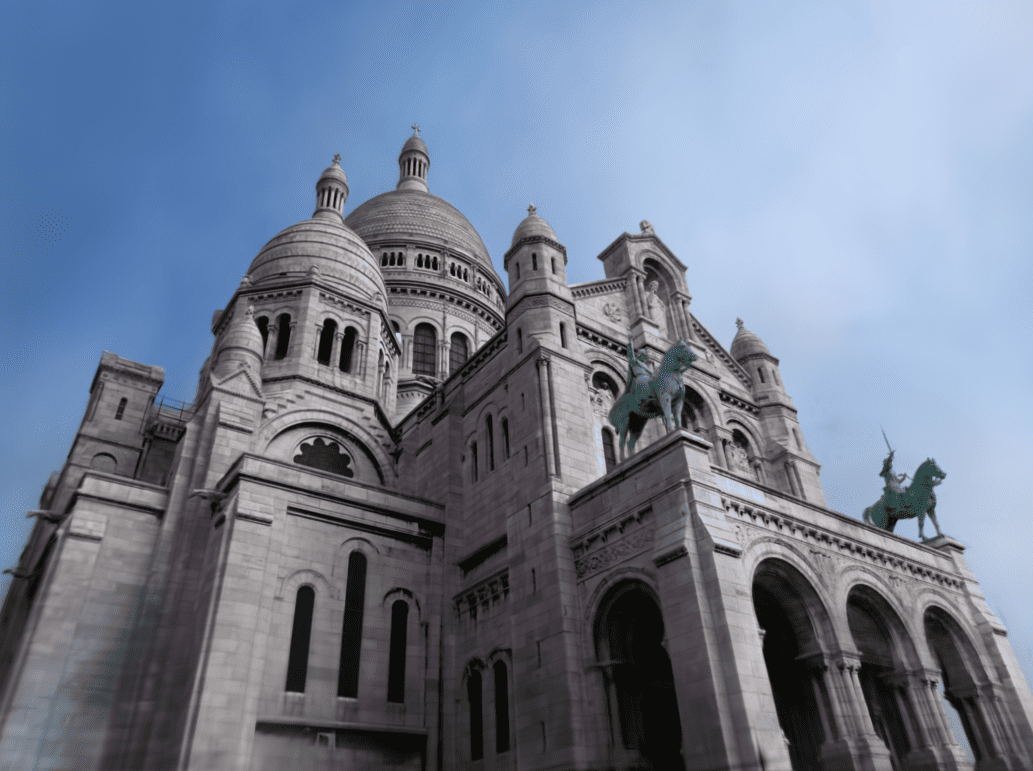} & ~ &
    \includegraphics[width=0.16\textwidth]{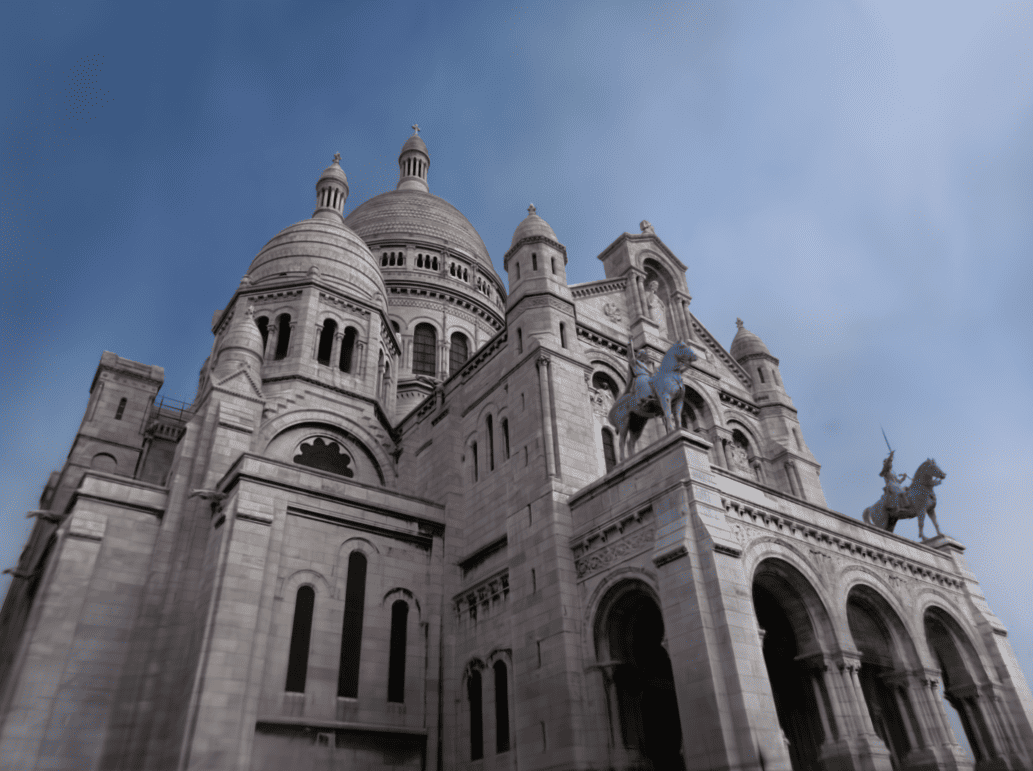}  &
    \includegraphics[width=0.16\textwidth]{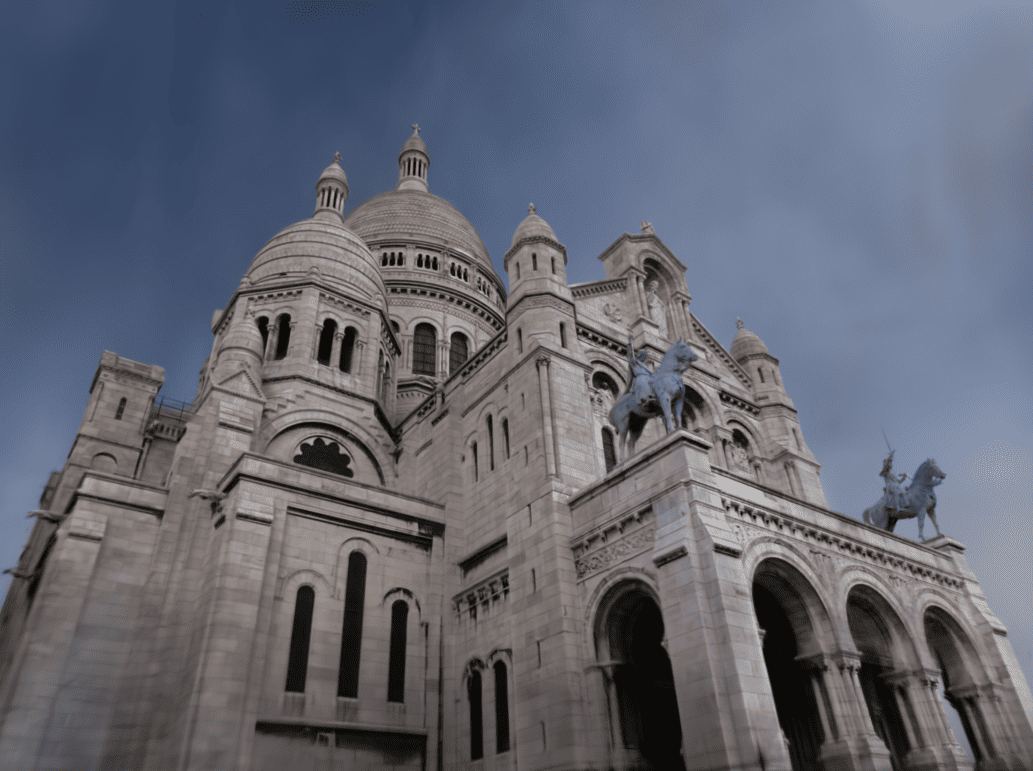}  &
    \includegraphics[width=0.16\textwidth]{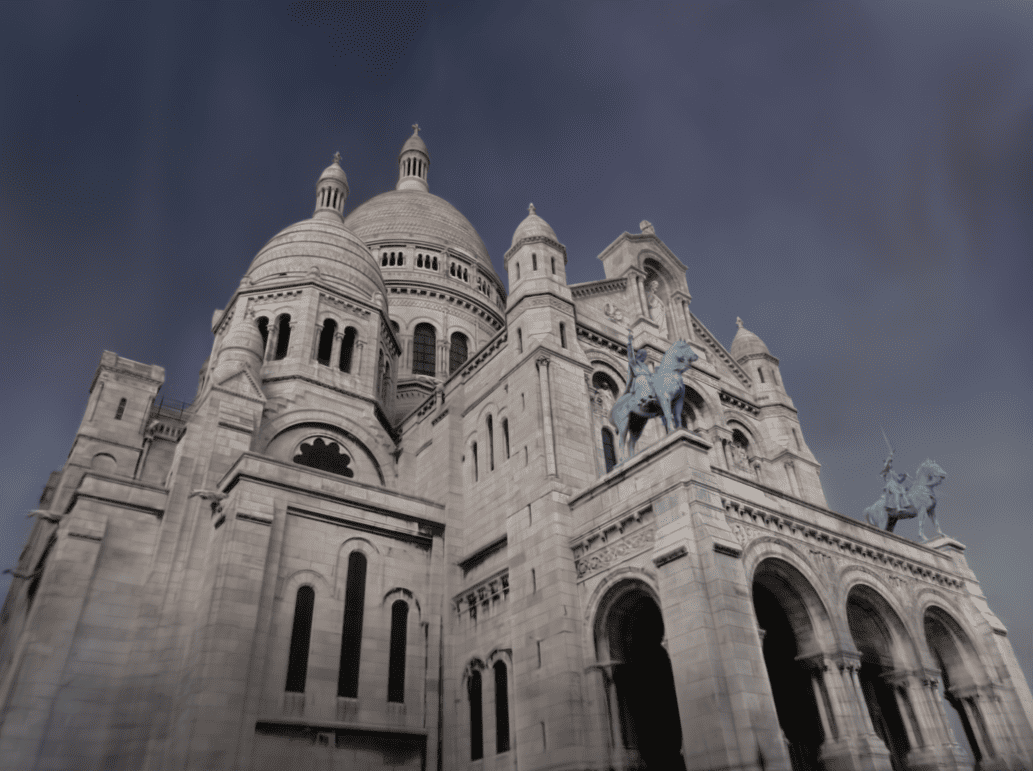}  &
    \includegraphics[width=0.16\textwidth]{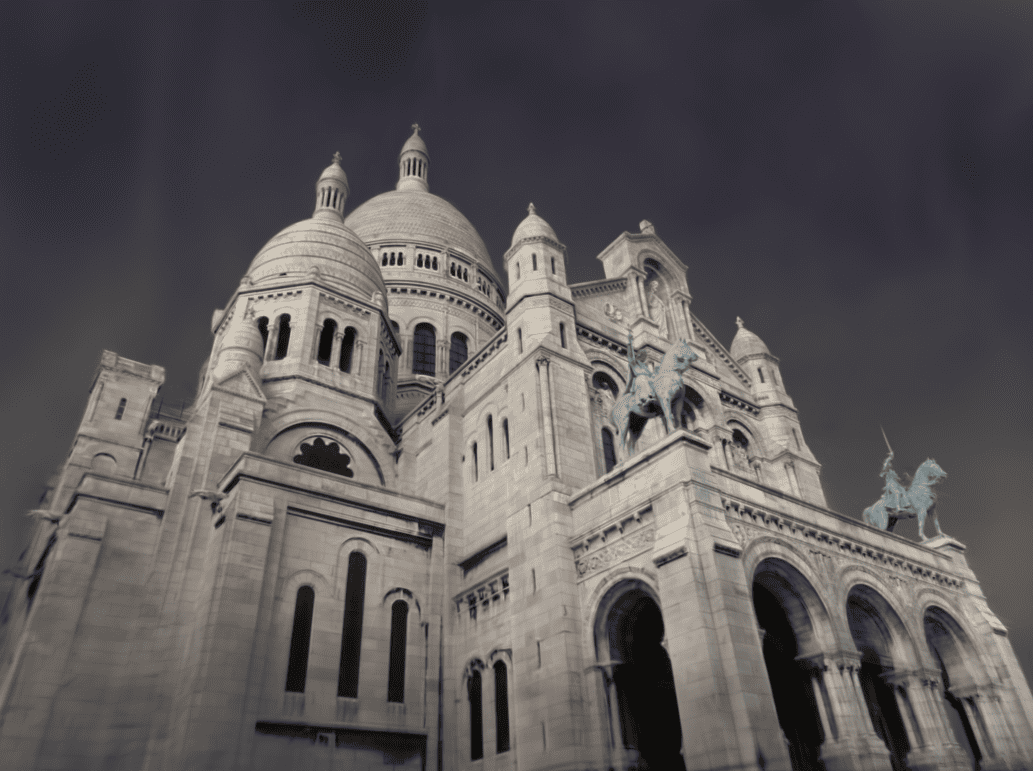} & ~ &
    \includegraphics[width=0.16\textwidth]{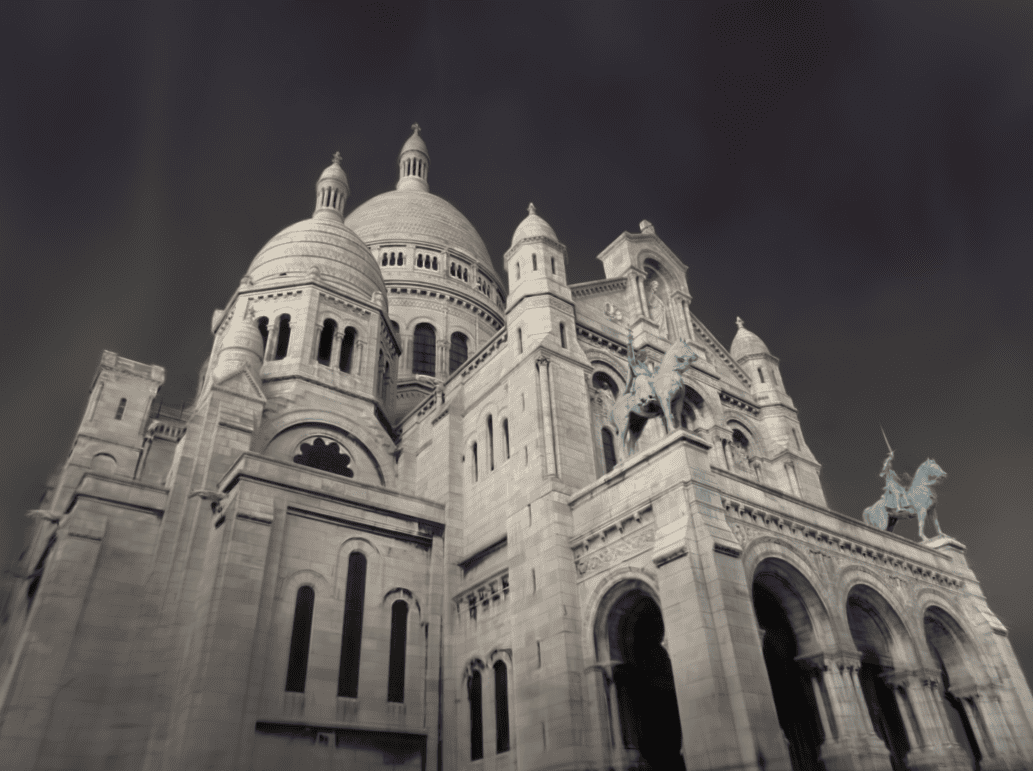} \\
    \multicolumn{8}{c}{ \includegraphics[width=0.8\textwidth]{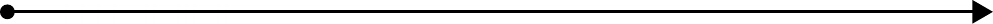}}
    
	\end{tabular}
	\caption{\textbf{Appearance interpolation --} Interpolations between the appearance embeddings of two training images (left, right).
    }
	\label{fig:interpolation}
\end{figure*}

Modeling the image appearance with an embedding vector enabled us to learn a latent space that we can utilize to change the lighting and appearance of any viewpoint at inference time. On top of the training image embeddings shown in~\Cref{fig:learnt}, we can sample vectors from the appearance space using interpolation between the learned embeddings. These interpolations are smooth and natural as shown in~\Cref{fig:interpolation}. We encourage the readers to explore the supplementary videos and appreciate more the naturalness of these interpolations.

\subsection{Transient Objects Removal}
As explained in~\Cref{sec:transient}, our method is able to localize 3D Gaussians that represent transient objects and, consequently, is able to manipulate the scenes omitting these objects.
It should be noted that \method~ is thoughtfully designed to model transient objects without allowing the Gaussians to vary their opacity to compensate for appearance variation. Indeed, according to our experiments, it was observed that less than 20\% (average on trained scenes) of the Gaussians capture transient objects. Consequently, \method~achieves a better reconstruction of the static scene parts without having blurry elongated Gaussians that arise in view-dependent appearances with 3DGS. The capabilities of our model of disentangling static and dynamic Gaussians are clearly highlighted in \Cref{fig:decom}. 

\begin{table}[tb]
    
    \caption{\textbf{Quantitative results --} Results of two ablations of SWAG : SWAG-A and SWAG-T on three real-world scenes from Phototourism~\cite{Jin_2020}.}
    \label{tab:ablation}
    \centering
    \resizebox{0.85\columnwidth}{!}{%
    \begin{tabular}{lcccccccccc}
        
        \cmidrule[\heavyrulewidth]{2-10}
         & \multicolumn{3}{c}{Bradenburg Gate}  & \multicolumn{3}{c}{Sacre Coeur} & \multicolumn{3}{c}{Trevi Fountain}   \\
        \cmidrule(rl){2-4}
        \cmidrule(rl){5-7}
        \cmidrule(rl){8-10}
        
         & PSNR $\uparrow$  & SSIM$\uparrow$  & LPIPS $\downarrow$ & PSNR $\uparrow$  & SSIM$\uparrow$  & LPIPS $\downarrow$ & PSNR $\uparrow$  & SSIM$\uparrow$  & LPIPS $\downarrow$ \\
        \midrule

        SWAG         
        &\cellcolor{orange!25}{26.33}   & \cellcolor{red!25}{0.929} & \cellcolor{red!25}{0.139}
           
            & \cellcolor{red!25}{21.16} & \cellcolor{red!25}{0.860}&\cellcolor{red!25}{0.185}
           
            & \cellcolor{red!25}{23.10}  & \cellcolor{red!25}{0.815} & \cellcolor{red!25}{0.208}\\ 
        
        SWAG-A 
        & \cellcolor{red!25}26.44 &\cellcolor{orange!25}0.928  & \cellcolor{orange!25}0.141          
        & \cellcolor{orange!25}{21.02} &\cellcolor{orange!25}{0.855} &\cellcolor{orange!25}{0.183}
        & \cellcolor{orange!25}22.35 &\cellcolor{orange!25}0.801&\cellcolor{orange!25}0.223\\  
        
        SWAG-T 
            &\cellcolor{yellow!25}24.46  & \cellcolor{yellow!25}0.914 & \cellcolor{yellow!25}0.160
        & \cellcolor{yellow!25}19.06 &\cellcolor{yellow!25}0.836&\cellcolor{yellow!25}0.205
        & \cellcolor{yellow!25}20.13 &\cellcolor{yellow!25}0.769&\cellcolor{yellow!25}0.244\\

        \bottomrule
    \end{tabular}
    }
\end{table}

\subsection{Ablations}
\begin{figure}[tb] \centering
   
    \includegraphics[width=0.19\textwidth]{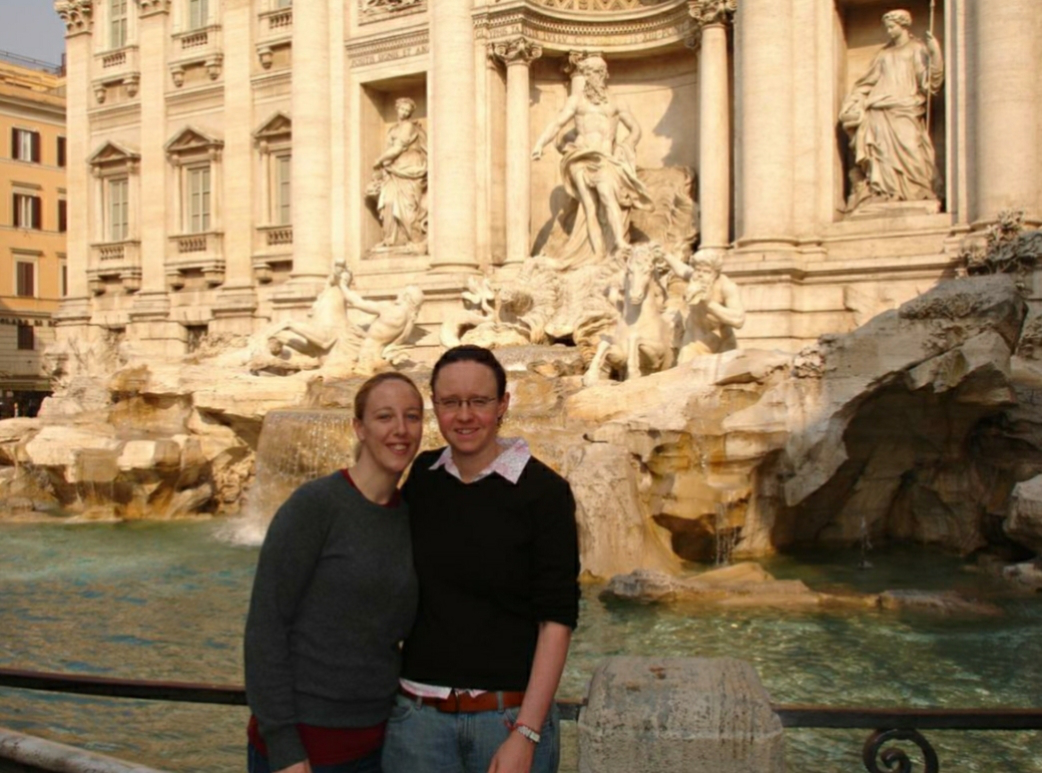}
    \includegraphics[width=0.19\textwidth]{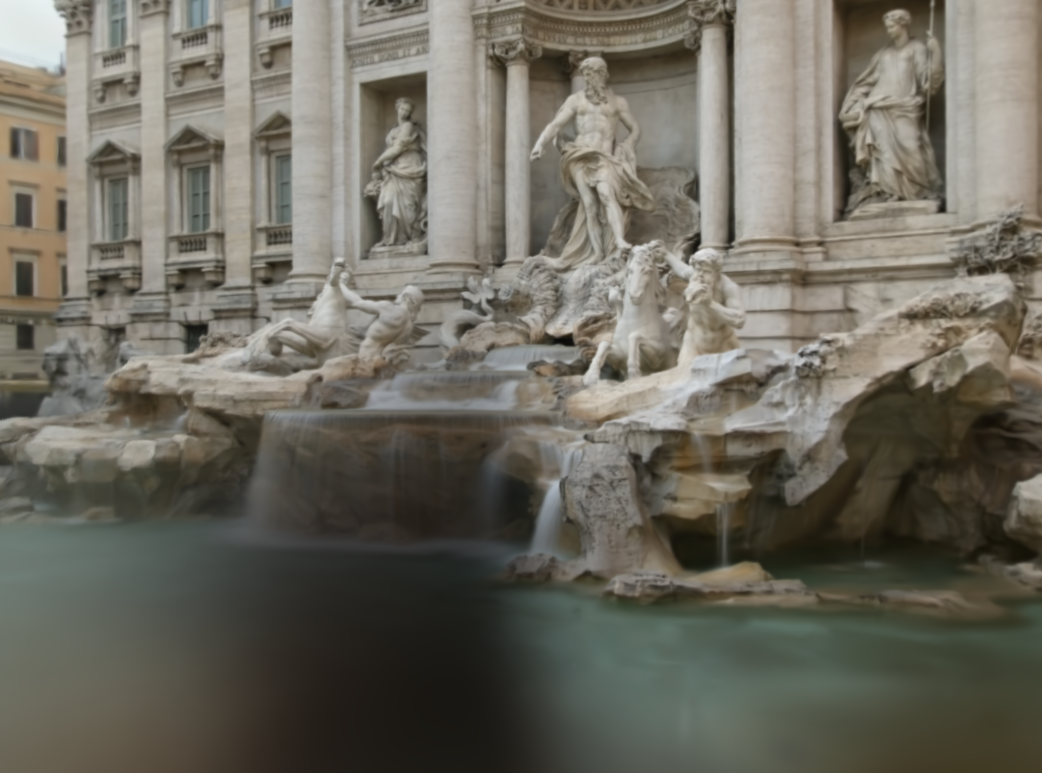}
    \includegraphics[width=0.19\textwidth]{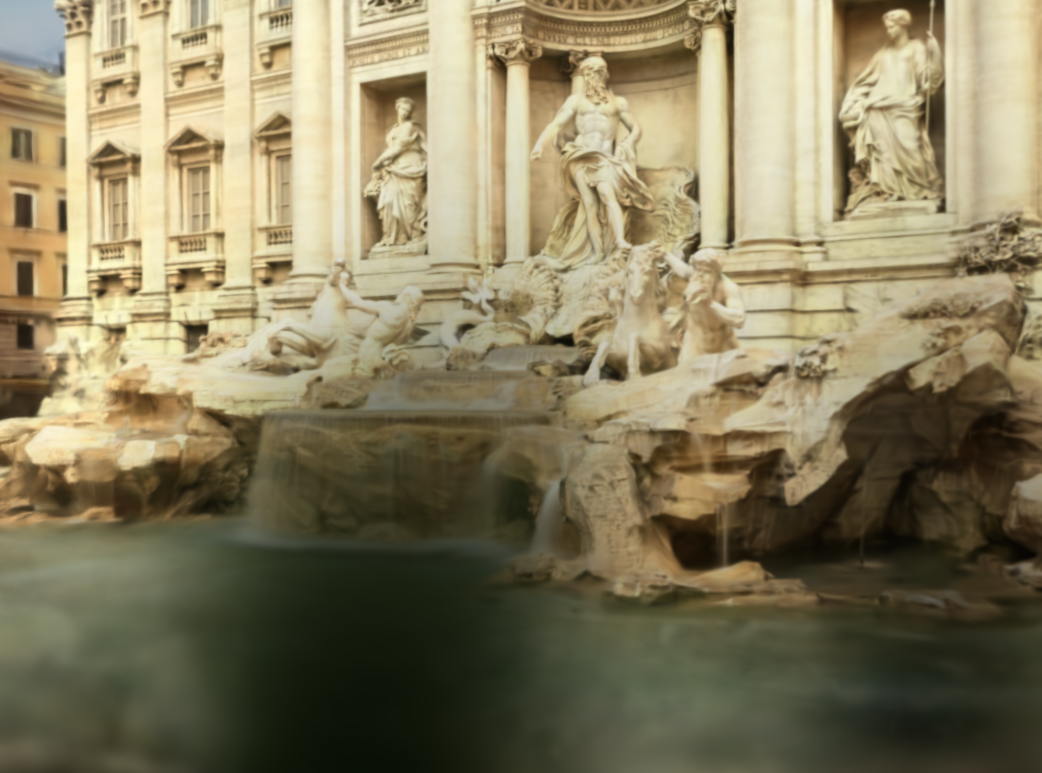}
    \includegraphics[width=0.19\textwidth]{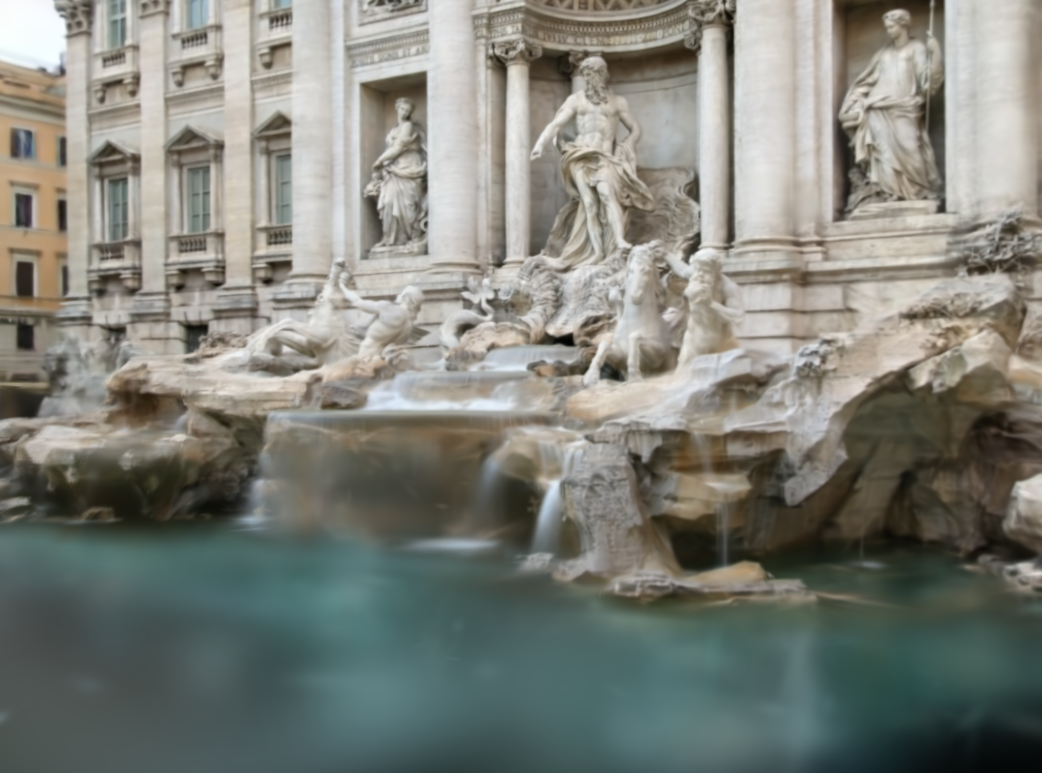}
    \includegraphics[width=0.19\textwidth]{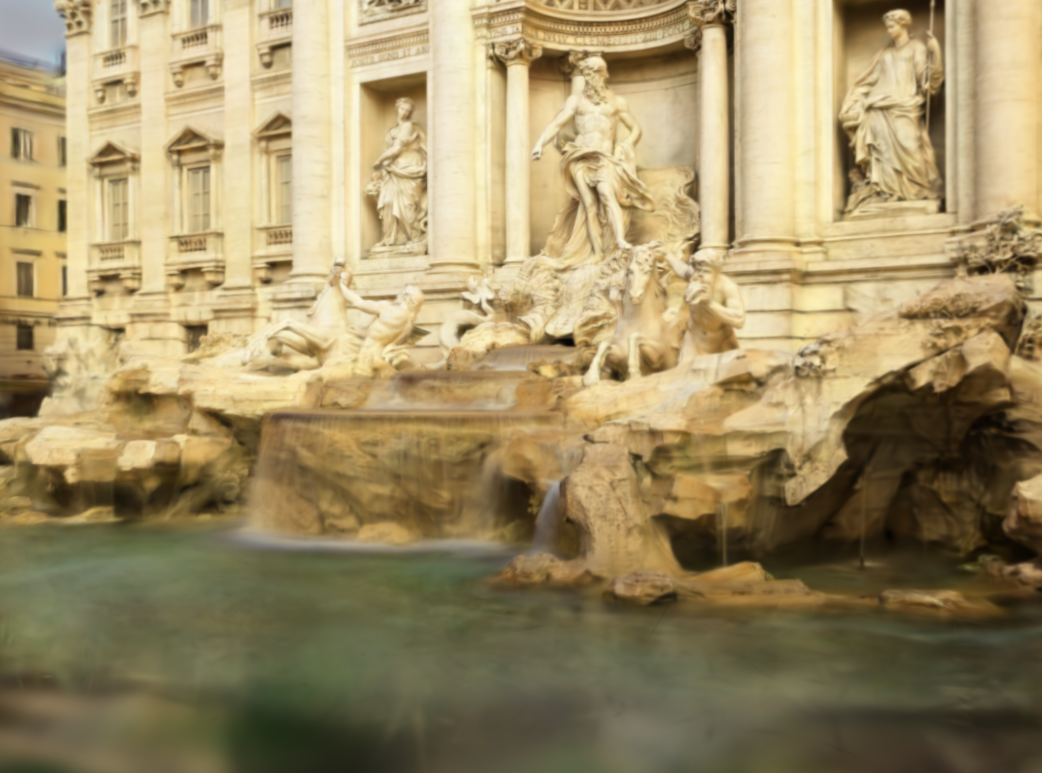}
    \\
     \makebox[0.19\textwidth]{\scriptsize Ground-truth}
     \makebox[0.19\textwidth]{\scriptsize 3DGS}
    \makebox[0.19\textwidth]{\scriptsize SWAG-A}
    \makebox[0.19\textwidth]{\scriptsize SWAG-T}
    \makebox[0.19\textwidth]{\scriptsize SWAG}
    \\
    \caption{\textbf{Ablation study --} Visual comparison of 3DGS, SWAG, and the two variants used in our ablation study: SWAG-T (no appearance) and SWAG-A (no transient).} 
    \label{fig:ablation}
\end{figure}

In this section, we compare two variants of our method to analyze the contribution of each component of \method:
\begin{itemize}
\item \textbf{SWAG-A}, a variation wherein the transient detection part of our model is removed,
\item \textbf{SWAG-T}, a variation wherein the image appearance color dependency is omitted.
\end{itemize}
We compare the results of our model variation on Phototourism scenes and we report the results in \Cref{tab:ablation}. SWAG-A enabled appearance modeling and closer colors to ground truth compared to 3DGS where colors shift away and converge to an average color for all training images. Transient objects' presence still causes floaters in the scene alike 3DGS as shown in \Cref{fig:ablation}. In contrast, SWAG-T enables the elimination of these floaters and produces an occluders-free scene while suffering from color alterations compared to ground truth like 3DGS. Combining these two variations enables SWAG to model varying appearances with the ability to reduce scene floaters.

\subsection{Discussion}
\method~leverages an MLP, a hash grid encoder, and image embedding to adapt 3DGS to \textit{in-the-wild} scenarios. While this empowered 3DGS's reconstruction capabilities in these conditions as it was demonstrated in our experiments, it almost doubles the training time and has 10 times longer inference time per frame as shown~\Cref{tab:quantitative1} compared to 3DGS. Nonetheless, our method still requires significantly less training time than all previous \textit{in-the-wild} baselines and achieves real-time rendering. 

Although SWAG succeeded in removing occluders using transient Gaussians, some parts of the scene where transient objects' presence is more frequent (eg. clouds in the sky and tourists on the roads) can suffer from being detected as transient. Eliminating all the transient Gaussians can leave some holes in these parts of the scene as seen in \Cref{fig:decom} for the Trevi Fountain scene where some parts of the fountain contain black spots. One way to solve this is to remove Gaussians based on an adaptive threshold $\lambda$ rather than considering as static only Gaussian with zero variance: $\mathbf{Var} \left [ \Delta\mathbf{\Tilde{\alpha}^{I}} \right ] < \lambda$, but it requires manual tuning of this threshold $\lambda$.

\section{Conclusion}

In this paper, we presented \method, a method designed for tailoring 3DGS representations to \textit{in-the-wild} scenarios.
\method~incorporates appearance modeling in the Gaussians' colors and employs an adaptive opacity modulation to handle the presence of transient objects.
Extensive experiments demonstrate that SWAG achieves state-of-the-art results on two challenging benchmarks while exhibiting training times orders of magnitude faster than \textit{in-the-wild} NVS baselines while enabling real-time rendering.
As a first step in conditioning 3DGS for \textit{in-the-wild} scene representations, this work suggests potential future research direction, such as extending SWAG to dynamic scenes.

\bibliographystyle{splncs04}
\bibliography{egbib}

In this supplementary material, we provide additional details about our hyperparameters in (\Cref{sec:hyperparameters}), and the datasets we used to evaluate our model (\Cref{sec:datasets}). We also provide an ablation study of our appearance modeling (\Cref{sec:app}) and an analysis of the 3D Gaussians distribution (\Cref{sec:gau}). 
\section{SWAG's Training hyperparameters}
\label{sec:hyperparameters}

A summary of the hyperparameters used during the training of our model for both synthetic and \textit{in-the-wild} scenes is provided in \Cref{tab:suppHyper}. We adjust these parameters across all scenes rather than on a per-scene basis. However, we believe that a per-scene tuning of these parameters can potentially improve the results of our model. 
\

\begin{table}[h!]
    
    \caption{\textbf{Hyperparameters --} Detailed description of parameters used in SWAG.
    }
    \label{tab:suppHyper}
    \centering
    \resizebox{0.6\columnwidth}{!}{%
    \begin{tabular}{llc}
        \toprule
           &Hyperparameter &Value\\
           \midrule    
    
      \multirow{3}{*}{3DGS Optimisation} &Iterations& 60000 \\
       &Densification interval &2000\\
       &Densification end & 30000\\
          
    \midrule
    \multirow{2}{*}{Image embeddings} &latent vector dimension & 24 \\
    &Learning rate  &0.0025\\

    \midrule
    
        \multirow{5}{*}{HashGrid Encoding} &Hash table size T & $2^{19}$\\
         &Finest resolution $N_{max}$ & 2048\\
        &Coarsest resolution $N_{min}$ & 16\\
        &Number of levels L & 12 \\
        &Learning rate &0.0025\\

        \midrule

        \multirow{5}{*}{MLP $\mathcal{F}_{\mathbf{\theta}}$} &Number of layer &3\\
        &Input dimension &51\\
        &Hidden layers size& 64\\
        &Output dimension & 4\\
        &Learning rate & 0.0025\\
        
        \bottomrule
       
    \end{tabular}
    }

\end{table}

\section{Datasets}
\label{sec:datasets}
\subsection{Phototourism Dataset~\cite{Jin_2020}}
This dataset comprises numerous images of touristic landmarks obtained from the internet. These images have visual challenges for reconstruction algorithms such as illumination variation and transient occluders. We initialize the 3D Gaussians using the initial set of sparse points from SfM. We evaluate SWAG on three specific scenes: Brandenburg Gate (950 images), Sacr\'e C\oe ur (859 images), and Trevi Fountain (1407 images). These landmarks are chosen due to their popularity and availability of a large number of internet images. Testing is performed on a standard set that is free of transient occluders.

\subsection{NeRF-OSR Dataset~\cite{rudnev2022nerfosr}}

\begin{table}[tbh]
    
    \caption{NeRF-OSR\cite{rudnev2022nerfosr} Dataset train/test split.}
    \centering
        \label{tab:suppnerfosrsplit}
     \begin{tabular}{lcc}
        \toprule
        Site & Train & Test \\
        \midrule
        lk2& 160 & 95 \\ 
        st& 301 & 96 \\
       lwp&258 &96\\
        \bottomrule
    \end{tabular}

\end{table}

This dataset contains eight sites, each captured from various viewpoints using a DSLR camera. In total, there are 3240 viewpoints captured across 110 different recording sessions. These sessions took place at different times of the day covering different weather conditions including sunny and cloudy days. Train/Test splits are detailed in \Cref{tab:suppnerfosrsplit}.
\begin{table}[bth]
    
    \caption{NeRF-MS\cite{Li_2023_ICCV} Dataset train/test split.}
    \centering
        \label{tab:suppnerfmssplit}
     \begin{tabular}{lcc}
        \toprule
        Site & Train & Test \\
        \midrule
        
       stjohann& 63 & 9 \\ 
        lwp& 58 & 9 \\
       st& 59&9\\
       europa& 51 & 8 \\
       
        \bottomrule
    \end{tabular}
\end{table}

Following NeRF-MS~\cite{Li_2023_ICCV}, we also evaluate SWAG on three sequences chosen for each of the four scenes which are captured under different lighting conditions. From each selected sequence, the first frame of every $8^{\text{th}}$ frame is selected as part of the testing set. Similarly, we use semantic segmentation to eliminate transient occlusions, which suggests a focus on evaluating the model's ability to handle static components. The \textit{train/test} splits used for the evaluation are reported in \Cref{tab:suppnerfmssplit}.

\subsection{Lego Dataset}

\begin{table}[tb]
    
    \caption{\textbf{Quantitative results --} Synthetic Lego Dataset.
    }
    \label{tab:suppquantitativeLego}
    \centering
    \resizebox{1.0\columnwidth}{!}{%
    \begin{tabular}{lccccccccccccc}
        \toprule
         & \multicolumn{3}{c}{Original}  & \multicolumn{3}{c}{Color Perturbation} & \multicolumn{3}{c}{Occlusion} & \multicolumn{3}{c}{Color \& Occlusion} \\
        \cmidrule(rl){2-4}
        \cmidrule(rl){5-7}
        \cmidrule(rl){8-10}
        \cmidrule(rl){11-13}
        
             & PSNR~$\uparrow$  & SSIM$~\uparrow$  & LPIPS~$\downarrow$ & PSNR~$\uparrow$  & SSIM~$\uparrow$  & LPIPS~$\downarrow$ & PSNR~$\uparrow$  & SSIM~$\uparrow$  & LPIPS~$\downarrow$  & PSNR~$\uparrow$  & SSIM~$\uparrow$  & LPIPS~$\downarrow$  \\

         \midrule
        NeRF-W \cite{kerbl3Dgaussians}   
        & \cellcolor{red!25}32.89  &\cellcolor{red!25}0.989 &\cellcolor{red!25}0.019

        &\cellcolor{red!25}31.51 &\cellcolor{red!25}0.987 &\cellcolor{red!25}0.022

        & 25.03& \cellcolor{yellow!25}0.946& \cellcolor{orange!25}0.063
        & \cellcolor{yellow!25}22.19& \cellcolor{yellow!25}0.927& \cellcolor{yellow!25}0.087\\
         \midrule
        Ha-NeRF~\cite{chen2022hallucinated} 
        & -- & -- & --
        & \cellcolor{orange!25}29.60 &\cellcolor{yellow!25}0.937 &0.294  
        & \cellcolor{red!25}30.93& \cellcolor{orange!25}0.949& \cellcolor{red!25}0.023
        & \cellcolor{orange!25}28.41&\cellcolor{orange!25} 0.934& \cellcolor{red!25}0.036\\
        
        \midrule
        3DGS \cite{kerbl3Dgaussians}   
        & \cellcolor{orange!25}32.34  & \cellcolor{orange!25}0.972 & \cellcolor{orange!25}0.042

        & 16.47 &0.553 &\cellcolor{yellow!25}0.267  
        & \cellcolor{yellow!25}26.18& 0.943& 0.076
        & 15.18& 0.532& 0.359\\
        \midrule
        SWAG(ours) 
        & \cellcolor{yellow!25}31.34  &\cellcolor{yellow!25}0.960&\cellcolor{yellow!25}0.058

        & \cellcolor{yellow!25}26.54  &\cellcolor{orange!25}0.951 &\cellcolor{orange!25}0.061 
        & \cellcolor{orange!25}30.02  & \cellcolor{red!25}0.952&\cellcolor{yellow!25}0.065 
        & \cellcolor{red!25}29.22& \cellcolor{red!25}0.945& \cellcolor{orange!25}0.072\\
        \bottomrule
    \end{tabular}
    }

\end{table}

Similarly to NeRF-W~\cite{martinbrualla2020nerfw}, we evaluate our model on two variants of the Lego dataset~\cite{mildenhall2020nerf}: one with color perturbations and another one with artificial occluders added on images. \Cref{tab:suppquantitativeLego} shows a quantitative comparison of various baselines. As shown with \textit{in-the-wild} scenes, the presence of appearance variations and occluders in the training images affects 3DGS effectiveness. Using SWAG, we improved 3DGS performance for these introduced color and occluder perturbations. We also have competitive results with the previous baselines over the three perturbations and outperform these baselines when combining color perturbations and occlusions.

\section{Ablation study on appearance modeling}

\begin{table}[tb]
    
    \caption{Ablation on SWAG appearance modeling effectiveness}
    \label{tab:suppablation}
    \centering
    \resizebox{0.76\columnwidth}{!}{%
    \begin{tabular}{lcccccc}
        
        \cmidrule[\heavyrulewidth]{2-7}
         & \multicolumn{3}{c}{Bradenburg Gate} & \multicolumn{3}{c}{Trevi Fountain}   \\
        \cmidrule(rl){2-4}
        \cmidrule(rl){5-7}

         & PSNR $\uparrow$  & SSIM$\uparrow$  & LPIPS $\downarrow$ & PSNR $\uparrow$  & SSIM$\uparrow$  & LPIPS $\downarrow$  \\
        \midrule
        Affine color transformation 
        &18.34  & 0.777&0.313
     
&19.01&0.761&0.246  
       \\
       \midrule
        Fourier series for pos. emb.
        &18.66  & 0.763&0.347

&19.91&0.741&0.305
        
       \\
    
        \midrule
        SWAG
            &\textbf{26.33}  & \textbf{0.929} & \textbf{0.139}

            & \textbf{23.10}& \textbf{0.815} & \textbf{0.208}
            \\
        \bottomrule
    \end{tabular}
    }
\end{table}

\begin{figure} [tb]
\centering
\begin{tabular}{cccc}
\scriptsize
\includegraphics[width=0.25\textwidth]{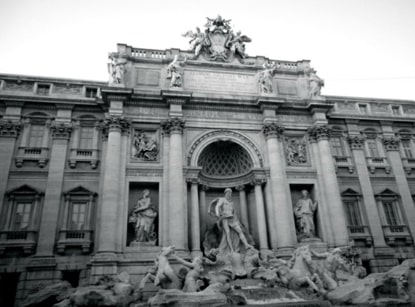} &
\includegraphics[width=0.25\textwidth]{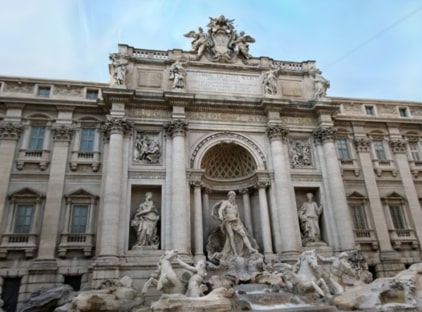} &
\includegraphics[width=0.25\textwidth]{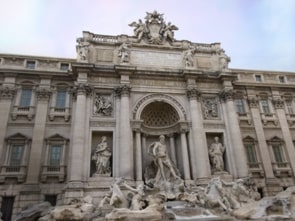} \\
\textbf{(a)}  & \textbf{(b)} & \textbf{(c)}  \\[6pt]
\end{tabular}
\begin{tabular}{cccc}
\includegraphics[width=0.25\textwidth]{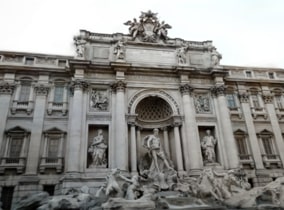} &
\includegraphics[width=0.25\textwidth]{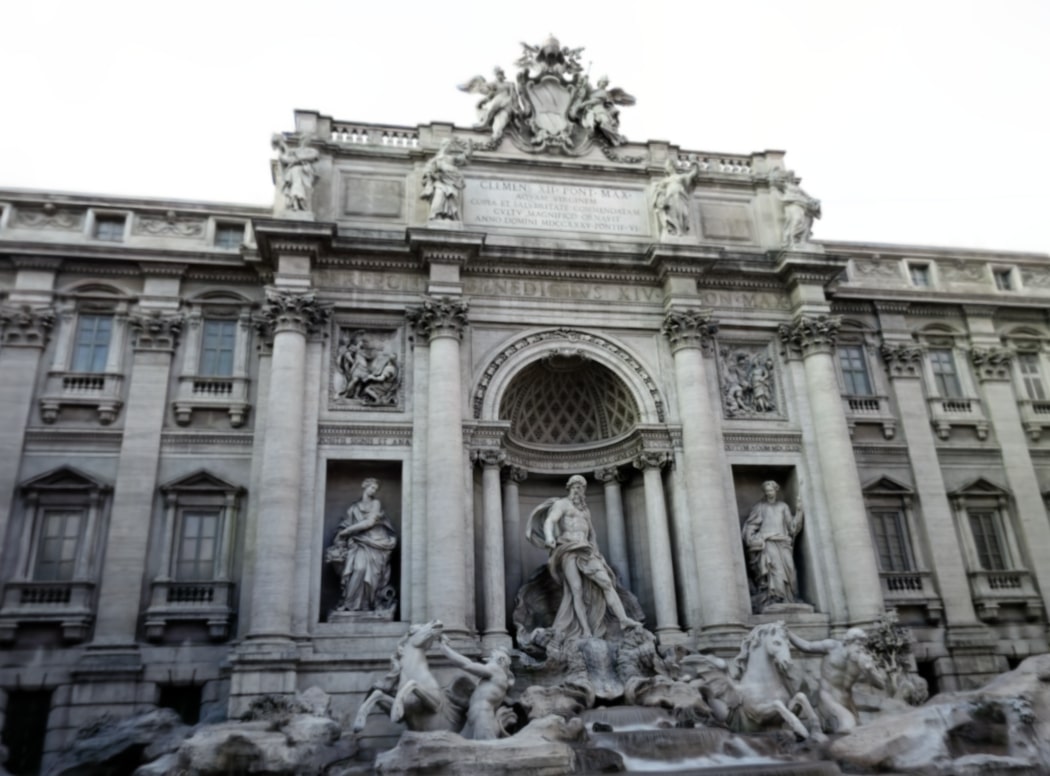} \\
\textbf{(d)}  & \textbf{(e)}  \\[6pt]
\end{tabular}
\caption{ \textbf{(a)} Ground-Truth
\textbf{(b)} 3DGS~\cite{kerbl3Dgaussians}
\textbf{(c)} Affine color transformation prediction
\textbf{(d)} Positional Encoding of 3D Gaussian centers
\textbf{(e)} SWAG
}
\label{fig:suppName}
\end{figure}

\label{sec:app}
In this section, we study the effectiveness of our appearance modeling choices. 
We compare our proposal to these two variants:
\begin{enumerate}
 \item \label{subsec:color}\textbf{An image-dependent affine color transformation for the Gaussian colors:}
we first employed an MLP to predict an affine color transformation (shift + scale) to each Gaussian. 
 \item \textbf{3D Gaussians centers encoding:} fixed sine/cosine positional encoding~\cite{transformers}.
\end{enumerate}
We present quantitative evaluation of these two approaches in \cref{tab:suppablation}. Although affine color transformations enabled injecting more appearance variance to 3DGS renderings (exposure,..), this transformation cannot model all possible appearances in \textit{in-the-wild} scenarios. Using a learned hash encoding (SWAG) for the 3D Gaussians centers enables a better appearance modeling compared to the fixed sine/cosine positional encodings as shown in (d) and (e) in \cref{fig:suppName}.

\section{3D Gaussians distribution analysis}
To model transient objects, we introduce an opacity variation that enables the separation of the final scene Gaussians into static and transient ones, and discarding these Gaussians keeps the static part of the scene. \Cref{tab:supptableAnalys} shows that even though SWAG uses less frequent densification steps than original 3DGS~\cite{kerbl3Dgaussians}, injecting image-dependent information ensured a finer representation with more Gaussians than 3DGS. We also present the number of transient Gaussians that represent transient objects and do not over-construct static parts of the scene except for some failure cases that we previously discussed (black spots on roads, \textit{Gate scene}, and fountain water, \textit{Trevi Fountain scene}).
\label{sec:gau}
\begin{table}[thb]\centering
    \caption{Transient/Static Gaussians distribution}
    \label{tab:supptableAnalys}
    \resizebox{0.48\textwidth}{!}{
    \scriptsize
    \begin{tabular}{*{10}{c}}
        \toprule
       Scene & SWAG &  transient&  3DGS \\
        \midrule
       Brandenburg Gate & 202082& 25273 &192748\\
       Sacre Coeur & 275315 &66754 & 246131  \\
        Trevi Fountain &573644  & 109672 & 517066  \\
        \bottomrule
    \end{tabular}
    }
\end{table}

\end{document}